%% file: main.tex
\documentclass[lettersize, journal]{IEEEtran}

\usepackage{comment}
\usepackage{booktabs}
\usepackage{multirow}
\usepackage{graphicx}
\usepackage{arydshln}
\usepackage{tikz}
\usepackage{subcaption}
\usepackage{amsmath}
\usepackage{hyperref}
\usepackage{cleveref}
\usepackage{enumitem}
\usepackage{graphicx}
\usepackage{xcolor}
\usepackage{cite}
\usepackage{pdfpages}
\makeatletter
\@ifundefined{@setmarks}{\let\@setmarks\relax}{}
\makeatother

\ifCLASSINFOpdf
\else

\fi

\begin{document}
\setcounter{page}{1}

\title{Exploring Emotion Expression Recognition in\\Older Adults Interacting with a Virtual Coach}

\author{Cristina~Palmero, Mikel~deVelasco, Mohamed~Amine~Hmani, Aymen~Mtibaa, Leila~Ben~Letaifa, Pau~Buch-Cardona, Raquel~Justo, Terry~Amorese, Eduardo~González-Fraile, Begoña~Fernández-Ruanova, Jofre~Tenorio-Laranga, Anna~Torp~Johansen, Micaela~Rodrigues~da~Silva, Liva~Jenny~Martinussen, Maria~Stylianou~Korsnes, Gennaro~Cordasco, Anna~Esposito, Mounim~A.~El-Yacoubi, Dijana~Petrovska‐Delacrétaz, M.~Inés~Torres, and~Sergio~Escalera%

\thanks{C. Palmero, P. Buch-Cardona, and S. Escalera are with Universitat de Barcelona and Computer Vision Center, Spain. E-mail: crpalmec7@alumnes.ub.edu.}%
\thanks{M. deVelasco, L. Ben Letaifa, R. Justo, and M.I. Torres are with Universidad del País Vasco UPV-EHU, Spain.}
\thanks{M.A. Hmani, A. Mtibaa, D. Petrovska‐Delacrétaz, and M.A. El-Yacoubi are with SAMOVAR, Telecom SudParis, Institut Polytechnique de Paris, France.}
\thanks{T. Amorese, G. Cordasco, and A. Esposito are with Università degli Studi della Campania L. Vanvitelli, Italy.}
\thanks{E. González-Fraile is with Universidad Internacional de La Rioja, Spain.}
\thanks{B. Fernández-Ruanova and J. Tenorio-Laranga are with Osatek, Spain.}
\thanks{A. Torp Johansen, M. Rodrigues da Silva, L. J. Martinussen, and M. Stylianou Kornes are with Oslo University Hospital, Norway.}

}

\markboth{Journal of \LaTeX\ Class Files,~Vol.~14, No.~8, August~2015}%
{Shell \MakeLowercase{\textit{et al.}}: Bare Demo of IEEEtran.cls for IEEE Journals}

\maketitle

\begin{abstract}
The EMPATHIC project aimed to design an emotionally expressive virtual coach capable of engaging healthy seniors to improve well-being and promote independent aging. One of the core aspects of the system is its human sensing capabilities, allowing for the perception of emotional states to provide a personalized experience. This paper outlines the development of the emotion expression recognition module of the virtual coach, encompassing data collection, annotation design, and a first methodological approach, all tailored to the project requirements. With the latter, we investigate the role of various modalities, individually and combined, for discrete emotion expression recognition in this context: speech from audio, and facial expressions, gaze, and head dynamics from video. The collected corpus includes users from Spain, France, and Norway, and was annotated separately for the audio and video channels with distinct emotional labels, allowing for a performance comparison across cultures and label types. Results confirm the informative power of the modalities studied for the emotional categories considered, with multimodal methods generally outperforming others (around 68\% accuracy with audio labels and 72-74\% with video labels). The findings are expected to contribute to the limited literature on emotion recognition applied to older adults in conversational human-machine interaction.
\end{abstract}

\IEEEpeerreviewmaketitle

\section{Introduction}
\label{sec:intro}
\input{Sections/1_intro.tex}

\section{Related work}
\label{sec:related}

\input{Sections/2_related.tex}

\section{EMPATHIC WoZ Corpus}
\label{sec:data}
\input{Sections/3_data.tex}

\input{Sections/4_labeling.tex}

\section{Methodology}
\label{sec:method}
\input{Sections/5_methodology.tex}

\section{Experimental evaluation}
\label{sec:evaluation}
\input{Sections/6_automatic_experiments.tex}

\section{Discussion}
\label{sec:discussion}
\input{Sections/7_discussion.tex}

\section{Conclusion}
\label{sec:conclusion}
\input{Sections/8_conclusion.tex}

\section*{Acknowledgments}
This work has been partially supported by the European Union’s Horizon 2020 R\&I program under grant agreement No 769872 (EMPATHIC project), the Spanish project PID2022-136436NB-I00, and ICREA under ICREA Academia program.

\ifCLASSOPTIONcaptionsoff
  \newpage
\fi

\bibliographystyle{IEEEtran}

\bibliography{main}

\renewcommand\thesection{\Alph{section}} 
\renewcommand\thesubsectiondis{\thesection.\Alph{subsection}.}
\setcounter{section}{0}
\setcounter{table}{0}

\section*{\textbf{- Supplementary Material -}}

In this supplementary material, we include additional methodological and evaluation details to those provided in Sec.~IV and Sec.~V of the main paper. First, we complete the description of the gaze and head trajectory filtering, and the computation of the \textit{Looking-at-VC} gaze feature (Sec.~\ref{ssec:filter} and Sec.~\ref{ssec:lookingatVC}, respectively, complementing Sec.~IV-C of the main paper). Second, we provide the sample size used to train and test the final models with the different audio- and video-based labels (Sec.~\ref{ssec:sample_size}, complementing Sec.~IV-D). Third, we also provide the complexity of the best models on validation for each evaluated final model (Sec.~\ref{ssec:complexity}, complementing Sec.~V-B). Finally, we report the per-country emotion expression recognition results for the audio-based (Sec.~\ref{sec:audio}, complementing Sec.~V-C), video-based under speech (Sec.~\ref{sec:video_speech}, complementing Sec.~V-D), and video-based under silence (Sec.~\ref{sec:video_silence}, complementing Sec.~V-E) evaluations.

\section{Methodology and evaluation details}

\subsection{Filtering invalid data from estimated head and eye gaze}
\label{ssec:filter}

When filtering invalid head and eye gaze trajectories, anatomically implausible movements refer to: eye rotation larger than 40$^{\circ}$~\cite{shin2016normal}, or faster than 860$^{\circ}$/s between consecutive frames, which is the highest peak angular speed that saccades have been reported to reach~\cite{bahill1975main}; and head movements faster than 700$^{\circ}$/s between consecutive frames, based on existing research on maximum rotation speed for voluntary motion~\cite{grossman1988frequency}.

\subsection{Looking-at-Virtual-Coach computation}
\label{ssec:lookingatVC}

We adapt the work of~\cite{amorese2022using} to find the zone with the highest density of gaze points. More specifically, we find gaze point clusters near the center of the plane using the Mean Shift clustering algorithm~\cite{comaniciu2002mean} with a bandwidth value estimated per video, and select the cluster with the highest number of points. To account for possible noisy estimates of the line of gaze, head pose, and VC's position, we assign weights to each gaze point based on its Mahalanobis distance to the cluster's distribution weighted by the cluster's inverse covariance. The weights are assigned to each gaze point $p_i$ such that:

\[
w(p_i) =
\begin{cases}
       1, & \hspace{0.1em} \text{if } d(p_i, c) \le thr_1\\
       (1 - d(p_i, c))/thr_2, & \hspace{0.1em} \text{if } thr_1 < d(p_i, c) \le thr_2\\
       0, & \hspace{0.1em} \text{otherwise,} \\ 
\end{cases}
\]
\\
where $d(p_i, c)$ is the Mahalanobis distance between the gaze point $p_i$ and the cluster $c$, and the thresholds are set to $thr_1 = 1$ and $thr_2 = 4$ standard deviations. For the second case of the piecewise function, points that belong to the cluster are transformed to be in the range $[0.7, 1)$, whereas points that do not belong to it are transformed to be in the range $(0, 0.7)$. These values were found empirically.
Per-point weights are further binned and converted into a 6D one-hot encoding vector denoting the likelihood of looking at VC from lower to higher. Per-valid-frame vectors are averaged over a time window, producing a 6D feature vector per window.

\subsection{Sample size used for evaluated models}
\label{ssec:sample_size}

Table~\ref{tab:audio_samples} includes the number of data samples per class and country used for the final evaluated models with audio-based labels. Similarly, Table~\ref{tab:video_samples} includes the number of data samples per class and country used for the final evaluated models with video-based labels when training with speech instances (left), and silence instances (right). As can be seen, the class ratios with respect to the original sample size are generally maintained.

\begin{table}[t!]
\centering
\caption{Inter-annotator agreement results for the audio annotations using pairwise Cohen's Kappa.}
\label{tab:audio_annots_kappa}
\begin{tabular}{@{}cccc@{}}
\toprule
\textbf{Annotators} & \textbf{Spain} & \textbf{France} & \textbf{Norway} \\ \midrule
\textbf{1 \& 2} & 0.766 & 0.507 & 0.692 \\
\textbf{1 \& 3} & 0.892 & 0.594 & - \\
\textbf{2 \& 3} & 0.719 & 0.562 & - \\
\midrule
\textbf{Mean} & 0.792 & 0.554 & 0.692 \\ \bottomrule
\end{tabular}%
\end{table}

\begin{table}[t!]
\centering
\caption{Number of audio segments from the EMPATHIC WoZ Corpus used for the evaluation of the final models, per label and per country.}
\label{tab:audio_samples}
\begin{tabular}{@{}lccc@{}}
\toprule
 & \multicolumn{1}{l}{\textbf{Calm}} & \multicolumn{1}{l}{\textbf{Pleased}} & \multicolumn{1}{l}{\textbf{Puzzled}} \\ \midrule
\textbf{Spain} & 33089 & 669 & 898 \\
\textbf{France} & 16534 & 316 & 369 \\
\textbf{Norway} & 12392 & 353 & 29 \\ \bottomrule
\end{tabular}%
\end{table}

\begin{table}[t!]
\centering
\caption{Number of frames from the EMPATHIC WoZ Corpus used for the evaluation of the final models, per label and per country, corresponding to spoken (left) and silent (right) instances.}
\label{tab:video_samples}
\begin{tabular}{@{}lcccccc@{}}
\toprule
 & \multicolumn{2}{c}{\textbf{Neutral}} & \multicolumn{2}{c}{\textbf{Happy}} & \multicolumn{2}{c}{\textbf{Pensive}} \\ \midrule
\textbf{Spain} & 858976 & 815634 & 8163 & 3028 &  184693 & 13233 \\
\textbf{France} & 471343 & 410344 & 27709 & 14213 &  93378 & 12225 \\
\textbf{Norway} & 321622 & 299338  & 10930 & 5031 &  67313 & 17291 \\ \bottomrule
\end{tabular}%
\end{table}

\subsection{Complexity of best evaluated models}
\label{ssec:complexity}

As explained in Sec.~IV-E of the main paper, we evaluate three low-complexity MLP configurations, from lowest to highest: 1) 100 hidden units for the first MLP layer, and 20 for the second, referred to as \textit{low} (L); 2) 200 and 40, referred to as \textit{mid} (M); and 3) 500 and 100, referred to as \textit{high} (H). For each evaluated model, with specific modalities, country and speaking status used for training, and label type, we select the best configuration in a validation subset and apply it to the test splits of each fold. Tables~\ref{tab:complex_audio},~\ref{tab:complex_video_speech} and~\ref{tab:complex_video_silence} report the best configuration for each model of audio, video-under-speech, and video-under-silence evaluations, respectively. 

\begin{table}[t!]
\centering
\caption{Complexity of the best MLP configuration for each evaluated audio-based model. L: 100 and 20 hidden units per layer. M: 200 and 40. H: 500 and 100.}
\label{tab:complex_audio}
\begin{tabular}{@{}lcccc@{}}
\toprule
\textbf{Modality} & \multicolumn{1}{l}{\textbf{Spain}} & \multicolumn{1}{l}{\textbf{France}} & \multicolumn{1}{l}{\textbf{Norway}} & \multicolumn{1}{l}{\textbf{Whole}} \\ \midrule
\textbf{A} & L & L & H & H \\ \midrule
\textbf{F} & M & L & H & L \\
\textbf{G} & H & H & H & L \\ \midrule
\textbf{A+F} & M & L & H & M \\
\textbf{A+G} & M & L & H & L \\
\textbf{A+F+G} & H & L & H & M \\ \bottomrule
\end{tabular}%
\end{table}

\begin{table}[t!]
\centering
\caption{Complexity of the best MLP configuration for each model evaluated on the video-based under speech scenario. L: 100 and 20 hidden units per layer. M: 200 and 40. H: 500 and 100.}
\label{tab:complex_video_speech}
\begin{tabular}{@{}lcccc@{}}
\toprule
\textbf{Modality} & \multicolumn{1}{l}{\textbf{Spain}} & \multicolumn{1}{l}{\textbf{France}} & \multicolumn{1}{l}{\textbf{Norway}} & \multicolumn{1}{l}{\textbf{Whole}} \\ \midrule
\multicolumn{5}{l}{\textit{Training on speech data:}} \\ \midrule
\textbf{F} & H & L & L & H \\ \midrule
\textbf{A} & H & H & M & H \\
\textbf{G} & H & H & M & H \\ \midrule
\textbf{F+A} & L & L & L & L \\
\textbf{F+G} & H & L & M & H \\
\textbf{F+A+G} & M & L & L & L \\ \midrule
\multicolumn{5}{l}{\textit{Training on all data (speech+silence):}} \\ \midrule
\textbf{F} & H & L & L & H \\ \midrule
\textbf{G} & H & M & M & H \\ \midrule
\textbf{F+G} & H & L & M & H \\ \bottomrule
\end{tabular}%
\end{table}

\begin{table}[t!]
\centering
\caption{Complexity of the best MLP configuration for each model evaluated on the video-based under silence scenario. L: 100 and 20 hidden units per layer. M: 200 and 40. H: 500 and 100.}
\label{tab:complex_video_silence}
\begin{tabular}{@{}lcccc@{}}
\toprule
\textbf{Modality} & \multicolumn{1}{l}{\textbf{Spain}} & \multicolumn{1}{l}{\textbf{France}} & \multicolumn{1}{l}{\textbf{Norway}} & \multicolumn{1}{l}{\textbf{Whole}} \\ \midrule
\multicolumn{5}{l}{\textit{Training on silence data:}} \\ \midrule
\textbf{F} & H & M & L & H \\ \midrule
\textbf{G} & M & L & L & H \\ \midrule
\textbf{F+G} & L & L & L & L \\ \midrule
\multicolumn{5}{l}{\textit{Training on all data (speech+silence):}} \\ \midrule
\textbf{F} & H & L & M & L \\ \midrule
\textbf{G} & L & L & L & H \\ \midrule
\textbf{F+G} & H & L & L & H \\ \bottomrule
\end{tabular}%
\end{table}

\section{Audio-based emotion recognition results}
\label{sec:audio}

\begingroup
\setlength{\tabcolsep}{5pt} 
\begin{table*}[t!]
\centering
\caption{Emotion recognition results for audio-based labels trained on the SPAIN (left) and WHOLE (right) training subsets and evaluated on the SPAIN test subset (train$ \rightarrow $test), reported as unweighted average accuracy $\pm$ SEM over 10 folds and 3 runs per fold. Bold: best accuracy per group. Underlined: best accuracy overall.}
\resizebox{\textwidth}{!}{%
\begin{tabular}{@{}lcccccccccc@{}}
\toprule
\multicolumn{2}{l}{}  & 
\multicolumn{4}{c}{\textbf{SPAIN $ \rightarrow $ SPAIN}} & \multicolumn{1}{l}{\textbf{}} & \multicolumn{4}{c}{\textbf{WHOLE $ \rightarrow $ SPAIN}} \\ \midrule
\multicolumn{1}{c}{\textbf{Modality}} & & \multicolumn{1}{c}{\textbf{\begin{tabular}[c]{@{}c@{}}Calm\\ Accuracy\end{tabular}}} & \multicolumn{1}{c}{\textbf{\begin{tabular}[c]{@{}c@{}}Pleased\\ Accuracy\end{tabular}}} & \multicolumn{1}{c}{\textbf{\begin{tabular}[c]{@{}c@{}}Puzzled\\ Accuracy\end{tabular}}} & \multicolumn{1}{c}{\textbf{\begin{tabular}[c]{@{}c@{}}Average\\ Accuracy\end{tabular}}} & &  \multicolumn{1}{c}{\textbf{\begin{tabular}[c]{@{}c@{}}Calm\\ Accuracy\end{tabular}}} & \multicolumn{1}{c}{\textbf{\begin{tabular}[c]{@{}c@{}}Pleased\\ Accuracy\end{tabular}}} & \multicolumn{1}{c}{\textbf{\begin{tabular}[c]{@{}c@{}}Puzzled\\ Accuracy\end{tabular}}} & \multicolumn{1}{c}{\textbf{\begin{tabular}[c]{@{}c@{}}Average\\ Accuracy\end{tabular}}} \\ \midrule
A & & 73.57 $\pm$ 1.8 & 58.17 $\pm$ 1.9 & 60.74 $\pm$ 2.4 &  64.16 $\pm$ 0.6 & &  74.06 $\pm$ 1.3 & 60.48 $\pm$ 2.4 & 63.91 $\pm$ 3.0 & 66.15 $\pm$ 0.9 \\ \midrule  
F & & \textbf{52.61 $\pm$ 3.3} & \textbf{59.61 $\pm$ 4.7} & 23.88 $\pm$ 3.6 & \textbf{45.37 $\pm$ 1.7} & & 19.42 $\pm$ 2.0 & \textbf{57.42 $\pm$ 5.2} & \underline{\textbf{66.76 $\pm$ 3.4}} & \textbf{47.87 $\pm$ 1.3} \\
G & & 43.45 $\pm$ 2.5 & 29.95 $\pm$ 3.6 & \textbf{35.67 $\pm$ 2.7} &  36.36 $\pm$ 1.0 & & \textbf{27.02 $\pm$ 1.9} & 34.36 $\pm$ 3.8 & 48.15 $\pm$ 2.5 & 36.51 $\pm$ 1.2 \\  \midrule 
A+F & & 72.12 $\pm$ 2.1 & \underline{\textbf{61.27 $\pm$ 2.7}} & \underline{\textbf{64.56 $\pm$ 2.4}} & 65.98 $\pm$ 0.7 & & 73.25 $\pm$ 1.1 & \underline{\textbf{63.16 $\pm$ 2.6}} & \textbf{65.35 $\pm$ 2.6} & 67.25 $\pm$ 0.8 \\
A+G & & \underline{\textbf{74.98 $\pm$ 1.7}} &  53.4 $\pm$ 2.0 & 61.03 $\pm$ 2.1 & 63.14 $\pm$ 0.6 & & 74.63 $\pm$ 1.2 & 57.43 $\pm$ 2.2 & 62.18 $\pm$ 3.0 & 64.75 $\pm$ 0.9 \\
A+F+G & & 73.49 $\pm$ 1.3 & 60.57 $\pm$ 3.0 & 64.33 $\pm$ 2.2 &  \underline{\textbf{66.13 $\pm$ 0.7}} & & \underline{\textbf{75.44 $\pm$ 1.2}} & 62.45 $\pm$ 2.8 & 64.04 $\pm$ 3.2 & \underline{\textbf{67.31 $\pm$ 1.0}} \\  
\bottomrule
\end{tabular}%
}
\label{tab:audio_whole_spain_results}
\end{table*}
\endgroup

In this section, we report the results for audio-based emotion recognition with respect to each country separately.

\subsection{SPAIN evaluation}

Table~\ref{tab:audio_whole_spain_results} summarizes the Spanish results. We discuss them below.

\subsubsection{Main modality} Contrary to WHOLE, \textit{pleased} obtains lower accuracy than \textit{puzzled}. Confusion matrices reveal that these two classes get confused with \textit{calm} at a similar proportion.

\subsubsection{Auxiliary modalities}  Average trends are similar to those for WHOLE, although the standard error of the mean (SEM) is higher for F than for G on average and per class. \textit{Pleased} obtains higher accuracy with F than with A for SP\textrightarrow SP. Confusion matrices reveal that, for G, \textit{puzzled} and \textit{pleased} are mostly confused with calm, while for F, \textit{puzzled} is more confused with \textit{calm}. Accuracy remarkably increases for \textit{puzzled} when training on WHOLE while decreasing at a similar degree for \textit{calm}. As a matter of fact, \textit{puzzled} obtains higher accuracy with F than A for WH\textrightarrow SP, while \textit{pleased} decreases with respect to SP\textrightarrow SP. Statistical tests confirm  significant differences (p$<$.0001) for the following pairwise comparisons: for SP\textrightarrow SP, A vs F, A vs G, F vs A+F, and G vs A+G; for WH\textrightarrow SP, all unimodal and unimodal vs bimodal comparisons.

\subsubsection{Multimodality} We observe that adding F to A moderately improves performance overall, and adding G to F+A increases it further, being the top performer. Statistical tests confirm significant differences for A+F vs A+G (p=.008). Class-wise, adding G is beneficial for \textit{calm}, with F+G obtaining the highest accuracy overall. Additionally, adding F is beneficial for \textit{pleased}, and both modalities are beneficial for \textit{puzzled}.

\subsubsection{Expanding training data} Training on WHOLE consistently improves performance for SP, with the highest accuracy overall obtained with A+F+G for WH\textrightarrow SP, although the SEM also increases for all models except for F. Most A-based models improve their performance for all classes. Auxiliary modalities see a substantial decrease in performance for \textit{calm}, while \textit{puzzled} recognition improves. Confusion matrices reveal that \textit{calm} is highly confused with the two minority classes for both auxiliary modalities, although \textit{pleased} is better recognized for F. For all models, the SEM decreases for \textit{calm}, but increases for the other classes, except for the auxiliary modalities, which see a decrease in SEM for \textit{puzzled}. There are no statistically significant differences.

\subsection{FRANCE evaluation}

\begingroup
\setlength{\tabcolsep}{5pt} 
\begin{table*}[t!]
\centering
\caption{Emotion recognition results for audio-based labels trained on the FRANCE (left) and WHOLE (right) training subsets and evaluated on the FRANCE test subset (train$ \rightarrow $test), reported as unweighted average accuracy $\pm$ SEM over 10 folds and 3 runs per fold. Bold: best accuracy per group. Underlined: best accuracy overall.}
\resizebox{\textwidth}{!}{%
\begin{tabular}{@{}lcccccccccc@{}}
\toprule
\multicolumn{2}{l}{}  & 
\multicolumn{4}{c}{\textbf{FRANCE $ \rightarrow $ FRANCE}} & \multicolumn{1}{l}{\textbf{}} & \multicolumn{4}{c}{\textbf{WHOLE $ \rightarrow $ FRANCE}} \\ \midrule
\multicolumn{1}{c}{\textbf{Modality}} & & \multicolumn{1}{c}{\textbf{\begin{tabular}[c]{@{}c@{}}Calm\\ Accuracy\end{tabular}}} & \multicolumn{1}{c}{\textbf{\begin{tabular}[c]{@{}c@{}}Pleased\\ Accuracy\end{tabular}}} & \multicolumn{1}{c}{\textbf{\begin{tabular}[c]{@{}c@{}}Puzzled\\ Accuracy\end{tabular}}} & \multicolumn{1}{c}{\textbf{\begin{tabular}[c]{@{}c@{}}Average\\ Accuracy\end{tabular}}} & &  \multicolumn{1}{c}{\textbf{\begin{tabular}[c]{@{}c@{}}Calm\\ Accuracy\end{tabular}}} & \multicolumn{1}{c}{\textbf{\begin{tabular}[c]{@{}c@{}}Pleased\\ Accuracy\end{tabular}}} & \multicolumn{1}{c}{\textbf{\begin{tabular}[c]{@{}c@{}}Puzzled\\ Accuracy\end{tabular}}} & \multicolumn{1}{c}{\textbf{\begin{tabular}[c]{@{}c@{}}Average\\ Accuracy\end{tabular}}} \\ 
\midrule 
A & & 72.34 $\pm$ 1.8 & \underline{64.04 $\pm$ 3.9} & 62.45 $\pm$ 2.1 & \underline{66.28 $\pm$ 1.2} & & \underline{76.62 $\pm$ 1.7} & 53.76 $\pm$ 4.7 & 60.61 $\pm$ 2.4 & \underline{63.66 $\pm$ 1.5} \\ \midrule
F & & 29.13 $\pm$ 2.1 &  \textbf{44.7 $\pm$ 4.1} & \underline{\textbf{63.49 $\pm$ 2.8}} & \textbf{45.77 $\pm$ 1.0} & & 12.06 $\pm$ 0.9 & \underline{\textbf{64.96 $\pm$ 4.5}} & \textbf{58.84 $\pm$ 3.1} & \textbf{45.29 $\pm$ 1.3}  \\
G & & \textbf{51.85 $\pm$ 2.7} & 31.56 $\pm$ 4.9 & 43.64 $\pm$ 3.4 & 42.35 $\pm$ 1.3 & & \textbf{22.87 $\pm$ 2.0} & 51.97 $\pm$ 4.8 & 44.16 $\pm$ 2.4 & 39.67 $\pm$ 1.5 \\  \midrule
A+F & & 73.5 $\pm$ 2.3 & 59.92 $\pm$ 4.6 & \textbf{61.97 $\pm$ 2.5} & 65.13 $\pm$ 1.5 & & 76.37 $\pm$ 1.6 & 50.62 $\pm$ 5.0 & \underline{\textbf{63.11 $\pm$ 2.4}} & \textbf{63.37 $\pm$ 1.5} \\
A+G & & 73.38 $\pm$ 2.4 & \textbf{62.77 $\pm$ 4.1} &  59.3 $\pm$ 2.1 & \textbf{65.15 $\pm$ 1.2} & & 76.37 $\pm$ 1.8 & \textbf{53.58 $\pm$ 5.0} & 59.64 $\pm$ 2.4 &  63.2 $\pm$ 1.7 \\
 A+F+G & & \underline{\textbf{74.08 $\pm$ 2.4}} & 55.34 $\pm$ 4.5 & 61.06 $\pm$ 2.5 & 63.49 $\pm$ 1.5 & & \textbf{76.4 $\pm$ 1.6} &  51.0 $\pm$ 5.1 &  62.5 $\pm$ 2.5 &  63.3 $\pm$ 1.5 \\ 
\bottomrule
\end{tabular}%
}
\label{tab:audio_whole_france_results}
\end{table*}
\endgroup

Table~\ref{tab:audio_whole_france_results} summarizes the French results. We discuss them below.

\subsubsection{Main modality} Trends are similar to those for WHOLE, although the SEM for \textit{pleased} is higher than that of \textit{puzzled} for FR. There is less confusion among classes than for SP.

\subsubsection{Auxiliary modalities} Trends are equivalent to those for WHOLE, with the SEM of F consistently lower than that of G. With F, \textit{puzzled} obtains the highest accuracy among all models with FR\textrightarrow FR, contrary to SP and NO. \textit{Pleased} accuracy notably increases when training on WHOLE. The confusion patterns of G are similar to those for SP; however, for F, \textit{calm} is greatly confused with \textit{puzzled}. Statistical tests confirm  significant differences (p$<$.0001) for the following pairwise comparisons: for FR\textrightarrow FR, A vs F, A vs G, F vs A+F, and G vs A+G; for WH\textrightarrow FR, all unimodal (F vs G obtaining p=.037) and unimodal vs bimodal comparisons.

\subsubsection{Multimodality} Contrary to SP and NO, for FR adding more modalities is detrimental to accuracy, although there are no statistically significant differences. Class-wise, for FR\textrightarrow FR, \textit{calm} obtaining the highest accuracy with A+F+G, while for WH\textrightarrow FR, \textit{puzzled} obtains the highest accuracy with A+F. However, these class-wise performance increases seem to come from pure accuracy redistribution, not increased discriminative power. See the discussion on the main paper (Sec.~V-C.3) for more details.

\subsubsection{Expanding training data} Training with WHOLE is not beneficial for FR, with all models consistently decreasing in accuracy on average and increasing their SEM. Class-wise, we observe that, for A-based models, \textit{calm} does benefit largely from this training regime, usually at the expense of a decrease in accuracy for the other classes. \textit{Puzzled} does improve for some multimodal models. By contrast, \textit{pleased} accuracy decreases substantially. We hypothesize that the distribution of audio features for \textit{pleased} instances is different for FR than for the other two countries, thus the additional variability is particularly detrimental. Auxiliary modalities follow an inverse trend compared to WHOLE, but the same as for SP, decreasing performance for \textit{calm} while increasing it greatly for \textit{pleased}. Following SP, the SEM for \textit{calm} decreases for all models, while for the other classes, it tends to increase for all models except for G.

\subsection{NORWAY evaluation}

\begingroup
\setlength{\tabcolsep}{5pt} 
\begin{table*}[t!]
\centering
\caption{Emotion recognition results for audio-based labels trained on the NORWAY (left) and WHOLE (right) training subsets and evaluated on the NORWAY test subset (train$ \rightarrow $test), reported as unweighted average accuracy $\pm$ SEM over 10 folds and 3 runs per fold. Bold: best accuracy per group. Underlined: best accuracy overall.}
\resizebox{\textwidth}{!}{%
\begin{tabular}{@{}lcccccccccc@{}}
\toprule
\multicolumn{2}{l}{} & 
\multicolumn{4}{c}{\textbf{NORWAY $ \rightarrow $ NORWAY}} & \multicolumn{1}{l}{\textbf{}} & \multicolumn{4}{c}{\textbf{WHOLE $ \rightarrow $ NORWAY}} \\ \midrule
\multicolumn{1}{c}{\textbf{Modality}} & & \multicolumn{1}{c}{\textbf{\begin{tabular}[c]{@{}c@{}}Calm\\ Accuracy\end{tabular}}} & \multicolumn{1}{c}{\textbf{\begin{tabular}[c]{@{}c@{}}Pleased\\ Accuracy\end{tabular}}} & \multicolumn{1}{c}{\textbf{\begin{tabular}[c]{@{}c@{}}Puzzled\\ Accuracy\end{tabular}}} & \multicolumn{1}{c}{\textbf{\begin{tabular}[c]{@{}c@{}}Average\\ Accuracy\end{tabular}}} & &  \multicolumn{1}{c}{\textbf{\begin{tabular}[c]{@{}c@{}}Calm\\ Accuracy\end{tabular}}} & \multicolumn{1}{c}{\textbf{\begin{tabular}[c]{@{}c@{}}Pleased\\ Accuracy\end{tabular}}} & \multicolumn{1}{c}{\textbf{\begin{tabular}[c]{@{}c@{}}Puzzled\\ Accuracy\end{tabular}}} & \multicolumn{1}{c}{\textbf{\begin{tabular}[c]{@{}c@{}}Average\\ Accuracy\end{tabular}}} \\  \midrule
A & & 87.79 $\pm$ 0.8 & 53.69 $\pm$ 4.1 & 20.83 $\pm$ 11.0 & 63.12 $\pm$ 1.9 & & \underline{79.85 $\pm$ 1.1} & 66.52 $\pm$ 4.1 & 30.09 $\pm$ 11.2 & 66.59 $\pm$ 2.2 \\ \midrule
F & & 43.79 $\pm$ 3.7 & \underline{\textbf{66.34 $\pm$ 3.7}} & \textbf{29.17 $\pm$ 12.6} & \textbf{49.89 $\pm$ 1.8} & & 12.5 $\pm$ 1.5 &  \underline{\textbf{81.6 $\pm$ 3.9}} & \textbf{23.15 $\pm$ 12.1} & \textbf{43.33 $\pm$ 2.6}  \\
G & & \textbf{63.09 $\pm$ 2.2} & 29.79 $\pm$ 4.0 &   1.39 $\pm$ 0.7 & 40.02 $\pm$ 1.8 & & \textbf{21.69 $\pm$ 1.8} &  60.5 $\pm$ 4.3 &   12.5 $\pm$ 6.5 & 36.83 $\pm$ 2.4 \\  \midrule
A+F & & 88.12 $\pm$ 0.9 & 60.96 $\pm$ 3.9 & 27.78 $\pm$ 12.7 &  67.3 $\pm$ 2.0 & & 79.43 $\pm$ 1.3 & 75.16 $\pm$ 4.3 & \underline{\textbf{30.56 $\pm$ 11.1}} & \underline{\textbf{70.13 $\pm$ 2.2}} \\
A+G & & 88.04 $\pm$ 0.6 & 53.17 $\pm$ 4.0 & 26.39 $\pm$ 12.9 & 63.88 $\pm$ 2.0 & & \textbf{79.77 $\pm$ 1.2} & 68.06 $\pm$ 3.9 &  16.67 $\pm$ 8.9 & 65.59 $\pm$ 2.3  \\
A+F+G & & \underline{\textbf{88.24 $\pm$ 0.8}} & \textbf{61.02 $\pm$ 3.6} & \underline{\textbf{29.17 $\pm$ 12.5}} & \underline{\textbf{67.7 $\pm$ 2.1}} & & 79.61 $\pm$ 1.2 & \textbf{76.7 $\pm$ 3.9} &  20.83 $\pm$ 9.6 & 69.75 $\pm$ 2.2 \\ 
\bottomrule
\end{tabular}%
}
\label{tab:audio_whole_norway_results}
\end{table*}
\endgroup

Table~\ref{tab:audio_whole_norway_results} summarizes the Norwegian results. We discuss them below.

\subsubsection{Main modality} For NO, there is a substantial difference across class performance due to data imbalance. More concretely, only 0.23\% of the NO dataset belongs to \textit{puzzled} instances, and some folds do not contain any test instance of this class, which makes this subset harder to evaluate. For this country, \textit{pleased} is more often confused with \textit{calm} than for the other countries.

\subsubsection{Auxiliary modalities} Accuracy trends are similar to those for WHOLE. For NO\textrightarrow NO, \textit{pleased} gets greatly confused with \textit{neutral} with G. By contrast, with F, \textit{pleased} obtains the highest accuracy among all models both when training with NO and with WHOLE. \textit{Pleased} accuracy also increases notably with G when adding more training data. Statistical tests confirm statistically significant differences (p$<$.0001) for the following pairwise comparisons: for NO\textrightarrow NO, A vs G, G vs A+G; for WH\textrightarrow NO, A vs F, A vs G, F vs A+F, and G vs A+G.

\subsubsection{Multimodality} NO shows similar trends with respect to WHOLE and SP, improving performance with multimodal models but with no statistically significant differences. More concretely, A+F+G obtains the highest performance. This is the highest performance increase caused by multimodal fusion across all countries.

\subsubsection{Expanding training data} Training with WHOLE is beneficial for A-based models, showing a higher improvement than for SP, although again, the SEM also increases. Class-wise, however, the accuracy for \textit{calm} decreases and for \textit{pleased} it increases, which may be an effect of the significant increase in the number of training instances for the latter (around 279\% increase). For \textit{puzzled}, the results are mixed, highly likely due to the extremely low number of test instances, despite also having the number of training instances increased (around 4369\%). Auxiliary modalities, by contrast, do not benefit from training with WHOLE on average. Class-wise, both mean and SEM greatly decrease for \textit{calm}, while for \textit{pleased} they increase. Similarly to the other countries, for G, we see the \textit{calm}-\textit{pleased} confusion reversed, and for F, \textit{calm} gets confused with the minority classes.

\section{Video-based recognition results under speech}
\label{sec:video_speech}

In this section, we report the results for emotion expression recognition with video-based labels, for those instances in which the user is speaking.

\subsection{SPAIN evaluation}

\begingroup
\setlength{\tabcolsep}{5pt} 
\begin{table*}[t!]
\centering
\caption{Emotion recognition results for video-based labels trained on the SPAIN (left) and WHOLE (right) training subsets under speech only or speech and silence instances, and evaluated on the SPAIN test subset under speech (train$ \rightarrow $test), reported as unweighted average accuracy $\pm$ SEM over 10 folds and 3 runs per fold. Bold: best accuracy per group. Underlined: best accuracy per training type. Italics: best accuracy overall.}
\resizebox{\textwidth}{!}{%
\begin{tabular}{@{}lcccccccccc@{}}
\toprule
\multicolumn{2}{l}{} & \multicolumn{4}{c}{\textbf{SPAIN $ \rightarrow $ SPAIN}} & \multicolumn{1}{l}{\textbf{}} & \multicolumn{4}{c}{\textbf{WHOLE $ \rightarrow $ SPAIN}} \\ \midrule
\multicolumn{1}{c}{\textbf{Modality}} & & \multicolumn{1}{c}{\textbf{\begin{tabular}[c]{@{}c@{}}Neutral\\ Accuracy\end{tabular}}} & \multicolumn{1}{c}{\textbf{\begin{tabular}[c]{@{}c@{}}Happy\\ Accuracy\end{tabular}}} & \multicolumn{1}{c}{\textbf{\begin{tabular}[c]{@{}c@{}}Pensive\\ Accuracy\end{tabular}}} & \multicolumn{1}{c}{\textbf{\begin{tabular}[c]{@{}c@{}}Average\\ Accuracy\end{tabular}}} & &  \multicolumn{1}{c}{\textbf{\begin{tabular}[c]{@{}c@{}}Neutral\\ Accuracy\end{tabular}}} & \multicolumn{1}{c}{\textbf{\begin{tabular}[c]{@{}c@{}}Happy\\ Accuracy\end{tabular}}} & \multicolumn{1}{c}{\textbf{\begin{tabular}[c]{@{}c@{}}Pensive\\ Accuracy\end{tabular}}} & \multicolumn{1}{c}{\textbf{\begin{tabular}[c]{@{}c@{}}Average\\ Accuracy\end{tabular}}} \\  \midrule
\multicolumn{11}{l}{\textit{Training on speech data:}}  \\ \midrule
F & & 72.56 $\pm$ 2.6 & 59.06 $\pm$ 3.8 & 60.45 $\pm$ 3.2 & 64.09 $\pm$ 1.3 & & 76.75 $\pm$ 1.8 & 57.39 $\pm$ 4.9 & 60.36 $\pm$ 3.4 & 65.14 $\pm$ 1.5 \\ \midrule
A & & 52.91 $\pm$ 1.5 & \textbf{56.33 $\pm$ 5.2} & 67.11 $\pm$ 1.7 & \textbf{58.95 $\pm$ 1.5} & & 51.27 $\pm$ 1.6 & \textbf{50.99 $\pm$ 5.1} &  71.2 $\pm$ 2.0 & 58.35 $\pm$ 1.5  \\
G & &  \textbf{58.95 $\pm$ 1.6} & 26.54 $\pm$ 3.6 & \underline{\textbf{76.51 $\pm$ 2.6}} & 55.08 $\pm$ 1.3 & & \textbf{58.37 $\pm$ 2.5} & 37.71 $\pm$ 3.5 & \underline{\textbf{79.12 $\pm$ 3.1}} &  \textbf{59.3 $\pm$ 1.4} \\  \midrule
F+A & & 75.25 $\pm$ 2.0 & \textit{\underline{\textbf{60.06 $\pm$ 5.3}}} & 63.45 $\pm$ 2.8 & 66.41 $\pm$ 1.5 & & 76.17 $\pm$ 1.9 & \textit{\underline{\textbf{61.92 $\pm$ 5.5}}} & 65.26 $\pm$ 3.3 & 68.04 $\pm$ 1.7 \\
F+G & & 76.27 $\pm$ 2.0 & 57.96 $\pm$ 3.5 & 70.43 $\pm$ 2.6 & 68.55 $\pm$ 1.1 & & \textit{\underline{\textbf{79.22 $\pm$ 1.8}}} & 56.68 $\pm$ 5.0 & 70.11 $\pm$ 3.0 & 69.14 $\pm$ 1.5 \\
F+A+G & & \textit{\underline{\textbf{77.4 $\pm$ 2.1}}} & 58.28 $\pm$ 5.1 & \textbf{71.34 $\pm$ 2.8} & \textit{\underline{\textbf{69.34 $\pm$ 1.4}}} & & 79.12 $\pm$ 1.7 &  61.6 $\pm$ 5.5 &  \textbf{70.7 $\pm$ 3.0} & \textit{\underline{\textbf{70.84 $\pm$ 1.7}}} \\ \midrule
\multicolumn{5}{l}{\textit{Training on all data (speech + silence):}}  \\ \midrule
F & & 69.33 $\pm$ 2.7 & \underline{59.12 $\pm$ 3.6} & 64.06 $\pm$ 3.2 & 64.19 $\pm$ 1.4 & & 73.14 $\pm$ 2.0 & \underline{58.42 $\pm$ 4.9} & 64.06 $\pm$ 3.2 & 65.44 $\pm$ 1.5 \\ \midrule 
G & & 47.92 $\pm$ 1.8 &  40.8 $\pm$ 4.5 & \textit{\underline{81.83 $\pm$ 2.4}} & 57.46 $\pm$ 1.6 & & 50.65 $\pm$ 2.8 &  45.1 $\pm$ 3.6 & \textit{\underline{82.04 $\pm$ 2.8}} & 59.91 $\pm$ 1.2 \\  \midrule 
F+G & & \underline{73.79 $\pm$ 2.2} & 57.89 $\pm$ 3.5 & 75.25 $\pm$ 2.4 & \underline{69.33 $\pm$ 1.0} & & \underline{77.09 $\pm$ 1.8} & 57.59 $\pm$ 5.0 & 73.21 $\pm$ 2.8 & \underline{69.73 $\pm$ 1.5} \\ 
\bottomrule

\end{tabular}%
}
\label{tab:video_speech_whole_spain_results}
\end{table*}
\endgroup

Table~\ref{tab:video_speech_whole_spain_results} summarizes the Spanish results. We detail them below.

\subsubsection{Main modality}
Contrary to WHOLE, for SP (and for the other countries), \textit{neutral} obtains the highest accuracy, while the accuracy for \textit{happy} and \textit{pensive} is reduced and similar. This time, the accuracy distribution does follow the per-class sample size. The two minority classes are confused with \textit{neutral} at a similar rate.

\subsubsection{Auxiliary modalities} For SP\textrightarrow SP, A outperforms F on average, while class-wise trends are the same as for WHOLE. For WH\textrightarrow SP, trends are the same as for WHOLE. By contrast, A does not outperform F for \textit{happy}. The SEM is reduced with both modalities on average and for all classes except for \textit{neutral} with G for WH\textrightarrow SP, and for \textit{happy} with A. Statistical results confirm significant differences for all cases of G vs F+G (p$<$.01),  and for A vs F+A (p$<$.01) only for WH\textrightarrow SP training with speech data.

\subsubsection{Multimodality}
Similar trends to WHOLE, with F+A+G obtaining the highest accuracy on average. Class-wise, SEM increases for \textit{happy} with multimodal models, but it decreases for the other classes. \textit{Pensive} is the class that most benefits from multimodality. Statistical significance results when training on speech data: for SP\textrightarrow SP, F+A vs F+A+G (p=.009); for WH\textrightarrow SP, F vs F+G (p=.004), F vs F+A (p=.015), F vs F+A+G (p$<$.001), and F+A vs F+A+G (p=.003). Significance results when training on speech and silence data: for WH\textrightarrow SP, F vs F+G (p=.001).

\subsubsection{Expanding training data including other countries} On average, we observe that training with the WHOLE dataset marginally increases accuracy, with the largest increase obtained for G when trained on speech data and no statistically significant differences. Class-wise, and in contrast to the other countries, we observe that, for F-based models, training with WHOLE causes the accuracy for \textit{neutral} to increase, especially for F alone, while for the auxiliary modalities, it only increases when training with speech and silence data. The accuracy for \textit{happy} and \textit{pensive} depends on the modality used: for F-based models, their accuracy decreases for all except for F+A, for which it increases, and F alone, for which \textit{pensive} stays the same; for A alone, \textit{happy} decreases and \textit{pensive} increases; and for G alone, they both increase. Furthermore, for F-based models, the SEM for \textit{neutral} is decreased, while for the other classes it is increased. By contrast, for auxiliary modalities, the SEM for \textit{neutral} and \textit{pensive} is increased, while for \textit{happy}, it is maintained. It is important to highlight that SP is the most unbalanced dataset over the three countries for both training regimes and, while its minority class, \textit{happy}, is increased the most with respect to the other countries (473\% when training with speech and 635\% when training with speech and silence), this does not translate into a higher accuracy for such class, suggesting that the higher number of samples does not provide the necessary variability to increase discriminative power. What is more, based on the findings from the silence-based scenario, for which the behavior for \textit{happy} and \textit{pensive} is reversed, we also hypothesize that the facial deformations caused by speaking are different across countries, hence adding variability that might not be necessary for a possibly narrow distribution of the minority classes for SP. However, this effect is compensated for when adding A, which helps increase discriminative power.

\subsubsection{Expanding training data including silence instances} Adding silence instances during training marginally but consistently improves performance on average for both SP\textrightarrow SP and WH\textrightarrow SP, with no statistically significant differences. Class-wise, we observe that training with silence instances decreases accuracy for \textit{neutral} while increasing it for \textit{pensive} compared to when training only with speech data. More concretely, G obtains the highest accuracy overall for \textit{pensive} for both SP\textrightarrow SP and WH\textrightarrow SP when training with speech and silence data, and the accuracy for \textit{happy} also increases. This translates into a redistribution of confusion from \textit{neutral} to the minority classes for all models. For G, class accuracies are the most balanced for the WH\textrightarrow SP scenario with this training regime, while for F-based models, it is the SP\textrightarrow SP scenario that obtains the best balance. Contrary to the effect of adding more training data with cross-country samples, incorporating silence instances for SP\textrightarrow SP mainly causes an increase in variability for \textit{neutral} instances, since the increase in \textit{happy} and \textit{pensive} instances is relatively low (95\% increase for \textit{neutral}, 37\% \textit{happy}, 7\% \textit{pensive}), and due to the subsampling technique only 15\% of the \textit{neutral} instances are used during training. By contrast, for WH\textrightarrow SP, the number of \textit{neutral} instances is doubled (with similar relative increase percentages as with SP\textrightarrow SP), but only 7.8\% of such instances are used. 

\subsection{FRANCE evaluation}

\begingroup
\setlength{\tabcolsep}{5pt} 
\begin{table*}[t!]
\centering
\caption{Emotion recognition results for video-based labels trained on the FRANCE (left) and WHOLE (right) training subsets under speech only or speech and silence instances, and evaluated on the FRANCE test subset under speech (train$ \rightarrow $test), reported as unweighted average accuracy $\pm$ SEM over 10 folds and 3 runs per fold. Bold: best accuracy per group. Underlined: best accuracy per training type. Italics: best accuracy overall.}
\resizebox{\textwidth}{!}{%
\begin{tabular}{@{}lcccccccccc@{}}
\toprule
\multicolumn{2}{l}{} & \multicolumn{4}{c}{\textbf{FRANCE $ \rightarrow $ FRANCE}} & \multicolumn{1}{l}{\textbf{}} & \multicolumn{4}{c}{\textbf{WHOLE $ \rightarrow $ FRANCE}} \\ \midrule
\multicolumn{1}{c}{\textbf{Modality}} & & \multicolumn{1}{c}{\textbf{\begin{tabular}[c]{@{}c@{}}Neutral\\ Accuracy\end{tabular}}} & \multicolumn{1}{c}{\textbf{\begin{tabular}[c]{@{}c@{}}Happy\\ Accuracy\end{tabular}}} & \multicolumn{1}{c}{\textbf{\begin{tabular}[c]{@{}c@{}}Pensive\\ Accuracy\end{tabular}}} & \multicolumn{1}{c}{\textbf{\begin{tabular}[c]{@{}c@{}}Average\\ Accuracy\end{tabular}}} & &  \multicolumn{1}{c}{\textbf{\begin{tabular}[c]{@{}c@{}}Neutral\\ Accuracy\end{tabular}}} & \multicolumn{1}{c}{\textbf{\begin{tabular}[c]{@{}c@{}}Happy\\ Accuracy\end{tabular}}} & \multicolumn{1}{c}{\textbf{\begin{tabular}[c]{@{}c@{}}Pensive\\ Accuracy\end{tabular}}} & \multicolumn{1}{c}{\textbf{\begin{tabular}[c]{@{}c@{}}Average\\ Accuracy\end{tabular}}} \\  \midrule
\multicolumn{5}{l}{\textit{Training on speech data:}}  \\ \midrule
F & & 77.57 $\pm$ 2.0 & 51.59 $\pm$ 4.0 & 50.35 $\pm$ 3.7 & 59.84 $\pm$ 1.5 & & 70.16 $\pm$ 2.6 & 66.28 $\pm$ 4.5 & 57.62 $\pm$ 4.0 & 64.69 $\pm$ 2.0 \\ \midrule
A & & 34.42 $\pm$ 1.2 &  \textbf{52.6 $\pm$ 4.8} & 64.79 $\pm$ 1.9 &  \textbf{50.6 $\pm$ 1.6} & & 18.42 $\pm$ 1.4 & \textit{\underline{\textbf{68.63 $\pm$ 3.6}}} & 60.37 $\pm$ 2.0 & 49.14 $\pm$ 0.8 \\
G & & \textbf{45.16 $\pm$ 2.0} & 34.75 $\pm$ 4.2 & \underline{\textbf{67.29 $\pm$ 3.6}} & 49.07 $\pm$ 0.8 & & \textbf{46.25 $\pm$ 3.0} & 44.46 $\pm$ 4.1 & \underline{\textbf{69.48 $\pm$ 3.9}} & \textbf{53.39 $\pm$ 1.5}  \\  \midrule 
F+A & & 80.27 $\pm$ 2.0 & \textit{\underline{\textbf{55.61 $\pm$ 4.1}}} & 44.12 $\pm$ 4.4 &  60.0 $\pm$ 1.9 & & 70.78 $\pm$ 2.4 & \textbf{67.9 $\pm$ 4.0} & 59.51 $\pm$ 4.1 & 66.07 $\pm$ 1.8 \\
F+G & & 80.24 $\pm$ 1.8 & 50.04 $\pm$ 4.0 & \textbf{57.48 $\pm$ 4.0} & 62.59 $\pm$ 1.5 & & \textit{\underline{\textbf{72.17 $\pm$ 2.4}}} &  65.3 $\pm$ 4.2 & 65.76 $\pm$ 3.6 & 67.74 $\pm$ 1.6 \\
F+A+G & & \textit{\underline{\textbf{81.3 $\pm$ 1.9}}} &  53.7 $\pm$ 4.1 & 53.35 $\pm$ 4.8 & \underline{\textbf{62.78 $\pm$ 1.9}} & & 71.72 $\pm$ 2.3 &  67.4 $\pm$ 3.9 & \textbf{66.98 $\pm$ 3.9} &  \textit{\underline{\textbf{68.7 $\pm$ 1.6}}} \\ 
\midrule
\multicolumn{5}{l}{\textit{Training on all data (speech + silence):}}  \\ \midrule
F & & 75.26 $\pm$ 2.1 & \underline{52.45 $\pm$ 4.0} &  53.7 $\pm$ 3.8 & 60.47 $\pm$ 1.5 & & 67.17 $\pm$ 2.7 &  \underline{67.7 $\pm$ 4.4} &  61.6 $\pm$ 3.8 & 65.49 $\pm$ 1.9  \\ \midrule
G &  & 36.01 $\pm$ 1.6 & 40.98 $\pm$ 3.8 & \textit{\underline{71.09 $\pm$ 3.5}} & 49.36 $\pm$ 0.7 & & 37.85 $\pm$ 3.0 & 51.68 $\pm$ 3.7 & \textit{\underline{71.46 $\pm$ 3.8}} & 53.67 $\pm$ 1.5 \\  \midrule
F+G & & \underline{77.82 $\pm$ 1.9} & 51.07 $\pm$ 3.9 & 62.77 $\pm$ 4.1 & \textit{\underline{63.89 $\pm$ 1.6}} & & \underline{69.2 $\pm$ 2.5} & 66.85 $\pm$ 4.0 & 69.47 $\pm$ 3.6 &  \underline{68.5 $\pm$ 1.6} \\ 
\bottomrule

\end{tabular}%
}
\label{tab:video_speech_whole_france_results}
\end{table*}
\endgroup

Table~\ref{tab:video_speech_whole_france_results} summarizes the French results. We discuss them below.

\subsubsection{Main modality} 
The accuracy gap between the minority classes and \textit{neutral} is higher for FR than for SP due to an increase in confusion. Nonetheless, it is worth mentioning that, when training with WHOLE, \textit{happy} performance almost equals that of \textit{neutral}.

\subsubsection{Auxiliary modalities} FR maintains the SP trends. One interesting difference is that, for FR, A moderately surpasses F when recognizing \textit{happy}, in addition to \textit{pensive} as in previous experiments, which is the behavior observed for WHOLE.  The SEM increases with A on average, while it decreases with G. Statistical significance results: when training on speech only, for FR\textrightarrow FR, all F vs A and F vs G comparisons (p=.024 for FR\textrightarrow FR and p=.005 for WH\textrightarrow FR) and all G/A vs F+G/A comparisons (all p$<$.001 except A vs F+A for FR\textrightarrow FR, p=.024); when training on speech and silence, F vs G (p=.013 for FR\textrightarrow FR, and p=.003 for WH\textrightarrow FR) and G vs F+G (p=.005 for FR\textrightarrow FR and p$<$.0001 for WH\textrightarrow FR).

\subsubsection{Multimodality} Similar trends to WHOLE, with F+A+G obtaining the highest accuracy on average again, and with \textit{pensive} being the most benefited class. However, contrary to the other countries, we find that adding A hinders the recognition of \textit{pensive} for FR\textrightarrow FR, and the accuracy increase on average is also minimal. Nonetheless, the \textit{pensive} anomaly gets corrected when training on WHOLE. The SEM increases on average, and for \textit{pensive} and \textit{happy}, except for F+G for the latter, while decreasing for \textit{neutral}. Statistical significance results when training on speech data: for FR\textrightarrow FR, F vs F+G (p=.02), F+A vs F+A+G (p=.021); for WH\textrightarrow FR, F vs F+G (p=.024), F vs F+A+G (p=.007), F+G vs F+A+G (p=.044), F+A vs F+A+G (p=.005). Statistical significance results when training on speech and silence data: F vs F+G (p=.009 for FR\textrightarrow FR, and p=.018 for WH\textrightarrow FR).

\subsubsection{Expanding training data including other countries} On average, we find that training on WHOLE consistently improves performance, with statistically significant differences for the following cases: when training with speech data, F+G (p=.027), F+A (p=.048), and F+A+G (p=.028); and when training with speech and silence data, F+G (p=.032). Contrary to SP, for FR \textit{neutral} performance greatly decreases while \textit{happy} and \textit{pensive} increases. It is not only an accuracy re-distribution across classes since the average accuracy also increases. Indeed, training with WHOLE increases the effective number of instances and variability for \textit{happy} and \textit{pensive} (69\% and 270\%, respectively), and the variability for \textit{neutral} (250\%), for which only a 15\% of the data is used. By analyzing confusion patterns, we find that training with WHOLE substantially increases the discriminative power of \textit{happy} and \textit{pensive} for F-based models. For G, only \textit{pensive} discriminative power increases, while for A we mainly observe a redistribution in confusion patterns.

\subsubsection{Expanding training data including silence instances} This scenario also consistently improves performance with respect to training on speech data only, but to a lesser extent than when including other countries. This performance improvement is statistically significant in the following cases at various p-value levels (p$<$.05): for FR\textrightarrow FR, F, and F+G; for WH\textrightarrow FR, F, and F+G. Class-wise, we observe a similar behavior to including other countries but to a lower extent, with the accuracy for \textit{neutral} decreasing while the accuracy for the other classes increases. Additionally, training with speech and silence tends to maintain overall confusion patterns with respect to training with speech only for each training regime separately (i.e., FR\textrightarrow FR and WH\textrightarrow FR). For F-based models, we obtain the best class accuracy balance on the WH\textrightarrow FR scenario. As for SP, G trained with speech and silence instances achieves the highest performance for \textit{pensive} across all evaluated models for both FR\textrightarrow FR and WH\textrightarrow FR, at the expense of an increase of \textit{neutral} samples being misclassified as \textit{pensive}. Similar to SP as well, and continuing with G, \textit{happy} accuracy increases due to a redistribution of confusion with \textit{neutral}, despite WH\textrightarrow FR obtaining the highest accuracy for \textit{happy} among the four scenarios. Thus, the most balanced results is obtained with WH\textrightarrow FR trained with speech only.

\subsection{NORWAY evaluation}

\begingroup
\setlength{\tabcolsep}{5pt} 
\begin{table*}[t!]
\centering
\caption{Emotion recognition results for video-based labels trained on the NORWAY (left) and WHOLE (right) training subsets under speech only or speech and silence instances, and evaluated on the NORWAY test subset under speech (train$ \rightarrow $test), reported as unweighted average accuracy $\pm$ SEM over 10 folds and 3 runs per fold. Bold: best accuracy per group. Underlined: best accuracy per training type. Italics: best accuracy overall.}
\resizebox{\textwidth}{!}{%
\begin{tabular}{@{}lcccccccccc@{}}
\toprule
\multicolumn{2}{l}{} & \multicolumn{4}{c}{\textbf{NORWAY $ \rightarrow $ NORWAY}} & \multicolumn{1}{l}{\textbf{}} & \multicolumn{4}{c}{\textbf{WHOLE $ \rightarrow $ NORWAY}} \\ \midrule
\multicolumn{1}{c}{\textbf{Modality}} & & \multicolumn{1}{c}{\textbf{\begin{tabular}[c]{@{}c@{}}Neutral\\ Accuracy\end{tabular}}} & \multicolumn{1}{c}{\textbf{\begin{tabular}[c]{@{}c@{}}Happy\\ Accuracy\end{tabular}}} & \multicolumn{1}{c}{\textbf{\begin{tabular}[c]{@{}c@{}}Pensive\\ Accuracy\end{tabular}}} & \multicolumn{1}{c}{\textbf{\begin{tabular}[c]{@{}c@{}}Average\\ Accuracy\end{tabular}}} & &  \multicolumn{1}{c}{\textbf{\begin{tabular}[c]{@{}c@{}}Neutral\\ Accuracy\end{tabular}}} & \multicolumn{1}{c}{\textbf{\begin{tabular}[c]{@{}c@{}}Happy\\ Accuracy\end{tabular}}} & \multicolumn{1}{c}{\textbf{\begin{tabular}[c]{@{}c@{}}Pensive\\ Accuracy\end{tabular}}} & \multicolumn{1}{c}{\textbf{\begin{tabular}[c]{@{}c@{}}Average\\ Accuracy\end{tabular}}} \\ \midrule
\multicolumn{5}{l}{\textit{Training on speech data:}}  \\ \midrule
F & & 84.03 $\pm$ 1.6 & 49.73 $\pm$ 5.4 &  47.0 $\pm$ 3.8 & 60.26 $\pm$ 2.7 & & 70.8 $\pm$ 1.8 & 68.08 $\pm$ 4.9 &  59.4 $\pm$ 4.0 & 66.09 $\pm$ 2.2 \\ \midrule
A & & 46.58 $\pm$ 1.2 & \textit{\underline{\textbf{63.82 $\pm$ 4.0}}} & \textit{\underline{\textbf{68.32 $\pm$ 2.0}}} & \textbf{59.57 $\pm$ 1.4} & & 42.24 $\pm$ 1.6 & \textbf{75.74 $\pm$ 2.8} & 65.31 $\pm$ 1.6 &  \textbf{61.1 $\pm$ 1.1}   \\
G & & \textbf{69.65 $\pm$ 3.0} & 21.41 $\pm$ 3.1 & 66.99 $\pm$ 4.2 & 52.68 $\pm$ 1.8 & & \textbf{42.86 $\pm$ 3.2} & 53.86 $\pm$ 4.5 & \textbf{68.46 $\pm$ 4.2} & 55.06 $\pm$ 1.6  \\  \midrule
F+A & & 85.13 $\pm$ 1.3 & \textbf{56.82 $\pm$ 4.9} & 52.69 $\pm$ 3.8 & 64.88 $\pm$ 2.6 & & 74.64 $\pm$ 1.5 & \textit{\underline{\textbf{75.89 $\pm$ 3.8}}} & 60.21 $\pm$ 4.0 & 70.25 $\pm$ 2.0  \\
F+G & & 86.69 $\pm$ 1.1 & 47.62 $\pm$ 5.4 & 59.45 $\pm$ 4.1 & 64.59 $\pm$ 2.8 & & 75.15 $\pm$ 1.5 & 67.25 $\pm$ 5.2 & 67.62 $\pm$ 4.4 & 70.01 $\pm$ 2.5 \\
F+A+G & & \textit{\underline{\textbf{87.77 $\pm$ 1.0}}} & 55.13 $\pm$ 4.9 & \textbf{62.63 $\pm$ 3.9} & \textit{\underline{\textbf{68.51 $\pm$ 2.7}}} & & \textit{\underline{\textbf{76.7 $\pm$ 1.2}}} & 75.34 $\pm$ 3.9 & \underline{\textbf{70.69 $\pm$ 4.0}} & \textit{\underline{\textbf{74.24 $\pm$ 2.2}}} \\ 
\midrule
\multicolumn{5}{l}{\textit{Training on all data (speech + silence):}}  \\ \midrule
F & & 81.57 $\pm$ 1.8 & \underline{53.71 $\pm$ 5.1} & 48.37 $\pm$ 3.8 & 61.22 $\pm$ 2.5 & & 66.5 $\pm$ 1.9 & 69.13 $\pm$ 4.9 & 62.85 $\pm$ 4.0 & 66.16 $\pm$ 2.3 \\ \midrule
G & & 60.23 $\pm$ 3.4 & 37.93 $\pm$ 5.5 & \underline{66.65 $\pm$ 4.2} & 54.93 $\pm$ 2.0  & & 31.69 $\pm$ 3.0 & 62.77 $\pm$ 5.1 & 70.27 $\pm$ 4.1 & 54.91 $\pm$ 1.6  \\  \midrule
F+G & & \underline{85.05 $\pm$ 1.2} & 52.31 $\pm$ 5.1 & 59.77 $\pm$ 4.0 & \underline{65.71 $\pm$ 2.8} & & \underline{70.95 $\pm$ 1.8} & \underline{69.25 $\pm$ 5.2} & \textit{\underline{71.45 $\pm$ 4.1}} & \underline{70.55 $\pm$ 2.4} \\ 
\bottomrule

\end{tabular}%
}
\label{tab:video_speech_whole_norway_results}
\end{table*}
\endgroup

Table~\ref{tab:video_speech_whole_norway_results} summarizes the Norwegian results, which we discuss below.

\subsubsection{Main modality} Trends are similar to FR, although the gap between \textit{neutral} and the minority classes is even higher, highly likely due to the decrease in sample size.

\subsubsection{Auxiliary modalities} We observe similar trends as for other countries; however, for NO\textrightarrow NO, A obtains the highest accuracy among all models for \textit{pensive}, instead of G. Actually, for the latter, \textit{happy} instances are mostly detected as \textit{neutral}, showing the highest confusion among countries. Statistically significant results: for NO\textrightarrow NO, F vs G/A (p=.046 and p=.048, respectively) and G vs F+G (p=.008) when training with speech data, and G vs F+G (p=.011) when training with speech and silence; for WH\textrightarrow NO, F vs G and G vs F+G for both training types (p$<$.0001), F vs G (p$<$.0001), A vs G (p=.007) and A vs F+A (p=.001) when training with speech data. 

\subsubsection{Multimodality} Contrary to the other countries, adding G and A to F obtain similar performance on average, with F+A slightly outperforming F+G. However, it is with the combination of the three modalities that the highest performance is achieved. Also, on average, the SEM decreases for F+A while it increases for F+G. Class-wise trends are similar, with \textit{pensive} again benefiting the most from multimodality. For F+A, the SEM decreases for \textit{neutral} and \textit{happy}, and increases for \textit{pensive}, with the latter especially for F+G. Statistical significance results when training on speech data: for NO\textrightarrow NO, all pairwise comparisons (all p$<$.001 except F+G vs F+A+G, with p=.003); for WH\textrightarrow NO, all pairwise comparisons (p$<$.001 except F vs F+A and F vs F+G vs F+A+G with p $<$.01). Statistical significance results when training on speech and silence data: for all cases, F vs F+G (p$<$.0001).

\subsubsection{Expanding training data including other countries} As for FR, training with WHOLE substantially improves performance, with statistically significant differences for all models at different p-value levels (p$<$.01) except for A and G. The largest performance increase is obtained by F+A+G. Class-wise, we also observe the same trend as for FR for F-based models, with \textit{neutral} accuracy decreasing while \textit{happy} and \textit{pensive} accuracy increases. For G, \textit{neutral}-\textit{happy} confusion is redistributed, while \textit{pensive} maintains its discriminative power. Finally, for A, \textit{neutral} gets more predicted as \textit{happy} than when training on NO, although the overall change is positive for the latter.

\subsubsection{Expanding training data including silence instances} This training scenario also improves performance consistently but to a lesser extent than when adding instances from other countries. Statistical tests show significant differences when training with speech only compared to speech and silence with F+G (p=.035) for NO\textrightarrow NO. We also see the same trends class-wise as for FR. Unlike other countries, NO obtains the highest accuracy for \textit{pensive} with F+G instead of with G for WH\textrightarrow NO. For F-based models, WH\textrightarrow NO training on speech and silence achieves the highest balance among class accuracies, while for G alone it is better when training on speech only, despite \textit{happy} achieving lower accuracy. Nonetheless, if we are interested in being better at detecting the minority classes even with a slight increase in false positives, then training on speech and silence would be recommended.

\section{Video-based recognition results under silence}
\label{sec:video_silence}

This section presents the emotion recognition results with video-based labels for instances where the user is not speaking.

\subsection{SPAIN evaluation}

\begingroup
\setlength{\tabcolsep}{5pt} 
\begin{table*}[t!]
\centering
\caption{Emotion recognition results for video-based labels trained on the SPAIN (left) and WHOLE (right) training subsets under silence only or speech and silence instances, and evaluated on the SPAIN test subset under silence (train$ \rightarrow $test), reported as unweighted average accuracy $\pm$ SEM over 10 folds and 3 runs per fold. Underlined: best accuracy per training type. Italics: best accuracy overall.}
\resizebox{\textwidth}{!}{%
\begin{tabular}{@{}lcccccccccc@{}}
\toprule
\multicolumn{2}{l}{} & \multicolumn{4}{c}{\textbf{SPAIN $ \rightarrow $ SPAIN}} & \multicolumn{1}{l}{\textbf{}} & \multicolumn{4}{c}{\textbf{WHOLE $ \rightarrow $ SPAIN}} \\ \midrule
\multicolumn{1}{c}{\textbf{Modality}} & & \multicolumn{1}{c}{\textbf{\begin{tabular}[c]{@{}c@{}}Neutral\\ Accuracy\end{tabular}}} & \multicolumn{1}{c}{\textbf{\begin{tabular}[c]{@{}c@{}}Happy\\ Accuracy\end{tabular}}} & \multicolumn{1}{c}{\textbf{\begin{tabular}[c]{@{}c@{}}Pensive\\ Accuracy\end{tabular}}} & \multicolumn{1}{c}{\textbf{\begin{tabular}[c]{@{}c@{}}Average\\ Accuracy\end{tabular}}} & &  \multicolumn{1}{c}{\textbf{\begin{tabular}[c]{@{}c@{}}Neutral\\ Accuracy\end{tabular}}} & \multicolumn{1}{c}{\textbf{\begin{tabular}[c]{@{}c@{}}Happy\\ Accuracy\end{tabular}}} & \multicolumn{1}{c}{\textbf{\begin{tabular}[c]{@{}c@{}}Pensive\\ Accuracy\end{tabular}}} & \multicolumn{1}{c}{\textbf{\begin{tabular}[c]{@{}c@{}}Average\\ Accuracy\end{tabular}}} \\ \midrule
\multicolumn{5}{l}{\textit{Training on silence data:}}  \\ \midrule
F & & 69.42 $\pm$ 3.5 & \underline{60.91 $\pm$ 4.9} &  73.3 $\pm$ 1.9 &  67.82 $\pm$ 2.1 & & 77.2 $\pm$ 2.8 &  \textit{\underline{67.55 $\pm$ 4.8}} &  65.57 $\pm$ 3.0 &  70.07 $\pm$ 1.9 \\ \midrule
G & & 66.46 $\pm$ 1.9 &  33.21 $\pm$ 5.7 &  80.01 $\pm$ 3.0 &  60.67 $\pm$ 1.8 & & 70.72 $\pm$ 2.0 &  27.23 $\pm$ 4.1 & \textit{\underline{78.35 $\pm$ 3.2}} &  59.87 $\pm$ 1.2 \\  \midrule 
F+G & & \underline{80.61 $\pm$ 2.0} &  56.19 $\pm$ 5.7 &  \textit{\underline{84.57 $\pm$ 1.1}} &  \textit{\underline{74.35 $\pm$ 1.9}} & & \underline{84.92 $\pm$ 1.4} & 64.35 $\pm$ 5.2 &  75.85 $\pm$ 1.6 & \textit{\underline{75.4 $\pm$ 1.8}} \\ \midrule
\multicolumn{5}{l}{\textit{Training on all data (speech + silence):}}  \\ \midrule
F & & 77.84 $\pm$ 3.0 & \textit{\underline{62.22 $\pm$ 5.0}} & 63.18 $\pm$ 2.0 &  67.71 $\pm$ 2.0 & & 83.54 $\pm$ 2.2 &  \underline{65.17 $\pm$ 5.2} &  59.59 $\pm$ 3.1 &  69.63 $\pm$ 1.9 \\ \midrule
G & & 67.14 $\pm$ 1.6 &  40.78 $\pm$ 6.1 &  \underline{75.08 $\pm$ 3.1} &  61.42 $\pm$ 2.2 & &  76.29 $\pm$ 1.7 & 35.47 $\pm$ 5.6 &  \underline{75.01 $\pm$ 3.2} & 63.1 $\pm$ 1.7 \\  \midrule
F+G & & \textit{\underline{86.71 $\pm$ 1.5}} &  58.06 $\pm$ 5.6 &  73.55 $\pm$ 1.6 &  \underline{73.33 $\pm$ 2.0} & & \textit{\underline{89.6 $\pm$ 1.2}} &  61.39 $\pm$ 6.1 &  66.72 $\pm$ 2.0 &  \underline{72.92 $\pm$ 2.0} \\ 
\bottomrule

\end{tabular}%
}
\label{tab:video_silence_whole_spain_results}
\end{table*}
\endgroup

Table~\ref{tab:video_silence_whole_spain_results} summarizes the Spanish results, which we discuss below.

\subsubsection{Main modality} Contrary to WHOLE, \textit{pensive} obtains the highest accuracy followed by \textit{neutral} for SP\textrightarrow SP, and \textit{pensive} is the most stable. Confusion patterns reveal a \textit{neutral}-\textit{pensive} confusion with the same proportion both ways, whereas \textit{happy} is mostly confused with \textit{neutral} in one direction.

\subsubsection{Auxiliary modality} We observe the same trends as for WHOLE, with G outperforming F for \textit{pensive} and being the top performer for such class for all training configurations except for SP\textrightarrow SP when training with silence data, for which F+G outperforms G for this class. By contrast, \textit{happy} performance is quite low. In addition, the SEM decreases on average and for \textit{neutral}, but increases for \textit{happy} and \textit{pensive}. We find significant differences for F vs F+G for SP\textrightarrow SP (p=.041 and p=.046 when training on silence, and speech and silence, respectively).

\subsubsection{Multimodality} Similar trends on average and class-wise, with F vs F+G statistically different when WH\textrightarrow SP (p=.041). In addition, the SEM for \textit{happy} increases with multimodality.

\subsubsection{Expanding training data including other countries} Training with WHOLE increases accuracy mostly when training on silence data only, except for G. By contrast, when training on speech and silence data, only F and G obtain an accuracy increase. Class-wise, \textit{neutral} and \textit{happy} increase performance, while \textit{pensive} decreases. G has a different behavior, decreasing accuracy for \textit{happy} when training with WHOLE. This class-wise behavior is partly opposed to what is found for SP on the video-under-speech evaluation scenario, for which \textit{neutral} increased performance, \textit{happy} decreased for F-based models and increased for G alone, and \textit{pensive} was maintained for F-based models and increased for G. It also contrasts with the behavior found for FR and NO across speech and silence scenarios, for which \textit{neutral} accuracy decreases while for \textit{happy} and \textit{pensive} accuracy increases (except for FR with G for the latter, which follows the same behavior as SP) when training with WHOLE. Compared to the speech-based scenario, \textit{happy} is easier to distinguish from \textit{neutral} with F alone when there are no facial deformations caused by speaking, while for G we notice an increase in confusion with \textit{neutral}. For \textit{pensive}, F-based models experience an increase in confusion with \textit{neutral} when training with WHOLE, which might be explained by a difference in the annotation procedure or user behavior for \textit{pensive} between SP and the other countries. The SEM increases on average and for \textit{pensive} when training with WHOLE, while for \textit{neutral} it decreases for F-based models and increases for G. Finally, for \textit{happy}, SEM decreases for G, while for F-based models it decreases when training on silence and increases when training on silence and speech.

\subsubsection{Expanding training data including speech instances} Contrary to the speech evaluation scenario, training on speech and silence data is detrimental when evaluating on silence only. On average, we only observe an increase in accuracy for G alone. Class-wise, for SP\textrightarrow SP we observe an increase in accuracy for \textit{neutral} and \textit{happy}, and a decrease for \textit{pensive}. By contrast, for WH\textrightarrow SP the accuracy for \textit{happy} decreases for F-based models instead, which concurs with our previous hypotheses, and increases for G. The SEM slightly increase on average and decrease for \textit{happy} for SP\textrightarrow SP, while for both training regimes the SEM of \textit{neutral} decreases. 

\subsection{FRANCE evaluation}

\begingroup
\setlength{\tabcolsep}{5pt} 
\begin{table*}[t!]
\centering
\caption{Emotion recognition results for video-based labels trained on the FRANCE (left) and WHOLE (right) training subsets under silence only or speech and silence instances, and evaluated on the FRANCE test subset under silence (train$ \rightarrow $test), reported as unweighted average accuracy $\pm$ SEM over 10 folds and 3 runs per fold. Underlined: best accuracy per training type. Italics: best accuracy overall.}
\resizebox{\textwidth}{!}{%
\begin{tabular}{@{}lcccccccccc@{}}
\toprule
\multicolumn{2}{l}{} & \multicolumn{4}{c}{\textbf{FRANCE $ \rightarrow $ FRANCE}} & \multicolumn{1}{l}{\textbf{}} & \multicolumn{4}{c}{\textbf{WHOLE $ \rightarrow $ FRANCE}} \\ \midrule
\multicolumn{1}{c}{\textbf{Modality}} & & \multicolumn{1}{c}{\textbf{\begin{tabular}[c]{@{}c@{}}Neutral\\ Accuracy\end{tabular}}} & \multicolumn{1}{c}{\textbf{\begin{tabular}[c]{@{}c@{}}Happy\\ Accuracy\end{tabular}}} & \multicolumn{1}{c}{\textbf{\begin{tabular}[c]{@{}c@{}}Pensive\\ Accuracy\end{tabular}}} & \multicolumn{1}{c}{\textbf{\begin{tabular}[c]{@{}c@{}}Average\\ Accuracy\end{tabular}}} & &  \multicolumn{1}{c}{\textbf{\begin{tabular}[c]{@{}c@{}}Neutral\\ Accuracy\end{tabular}}} & \multicolumn{1}{c}{\textbf{\begin{tabular}[c]{@{}c@{}}Happy\\ Accuracy\end{tabular}}} & \multicolumn{1}{c}{\textbf{\begin{tabular}[c]{@{}c@{}}Pensive\\ Accuracy\end{tabular}}} & \multicolumn{1}{c}{\textbf{\begin{tabular}[c]{@{}c@{}}Average\\ Accuracy\end{tabular}}} \\ \midrule
\multicolumn{5}{l}{\textit{Training on silence data:}}  \\ \midrule
F & & 78.16 $\pm$ 2.4 &  \textit{\underline{48.13 $\pm$ 5.0}} &  50.35 $\pm$ 3.2 &  58.88 $\pm$ 1.6 & & 73.36 $\pm$ 2.4 &  \textit{\underline{72.74 $\pm$ 4.6}} &  59.29 $\pm$ 2.4 &  68.46 $\pm$ 2.0 \\ \midrule
G & & 52.24 $\pm$ 2.7 &  36.49 $\pm$ 4.3 & \textit{\underline{79.54 $\pm$ 3.2}} & 56.09 $\pm$ 1.8 & & 56.21 $\pm$ 3.0 &  40.36 $\pm$ 4.3 & \textit{\underline{76.58 $\pm$ 3.1}} &  57.72 $\pm$ 1.4 \\  \midrule 
F+G & & \underline{84.16 $\pm$ 1.8} &  47.39 $\pm$ 5.1 &  72.58 $\pm$ 2.2 & \textit{\underline{68.05 $\pm$ 1.7}} & & \underline{78.07 $\pm$ 2.0} & 70.85 $\pm$ 4.4 &  74.41 $\pm$ 2.8 &  \textit{\underline{74.44 $\pm$ 2.1}} \\ \midrule
\multicolumn{5}{l}{\textit{Training on all data (speech + silence):}}  \\ \midrule
F & & 84.31 $\pm$ 1.9 &  \underline{47.34 $\pm$ 5.0} &  41.41 $\pm$ 3.3 &  57.69 $\pm$ 1.5 & & 78.68 $\pm$ 2.2 &  \underline{70.71 $\pm$ 4.7} &  51.84 $\pm$ 2.3 &  67.08 $\pm$ 1.9 \\ \midrule
G & &  59.74 $\pm$ 2.6 &  36.14 $\pm$ 5.0 &  \underline{75.58 $\pm$ 3.0} &  57.15 $\pm$ 1.6 & & 63.11 $\pm$ 3.1 &  42.56 $\pm$ 5.8 &  \underline{75.53 $\pm$ 2.8} & 60.4 $\pm$ 0.7 \\  \midrule
F+G & & \textit{\underline{89.57 $\pm$ 1.7}} &  45.74 $\pm$ 5.0 &  53.66 $\pm$ 3.3 &  \underline{62.99 $\pm$ 1.6} & & \textit{\underline{82.88 $\pm$ 2.0}} &  69.87 $\pm$ 4.5 &  64.24 $\pm$ 2.8 &  \underline{72.33 $\pm$ 2.1} \\ 
\bottomrule

\end{tabular}%
}
\label{tab:video_silence_whole_france_results}
\end{table*}
\endgroup

Table~\ref{tab:video_silence_whole_france_results} summarizes the French results. We report them below.

\subsubsection{Main modality} For FR, trends are similar with respect to the speech evaluation, with a decrease in accuracy and stability for \textit{happy}. 

\subsubsection{Auxiliary modality} In line with previous per-country findings, G alone obtains the highest accuracy overall for \textit{pensive}. The average performance difference between F and G is even smaller than for SP. The following statistically significant differences are found: for FR\textrightarrow FR, G vs F+G when training on silence (p=.009); for WH\textrightarrow FR, F vs G (p=.012) and G vs F+G (p$<$.001) when training with silence data, and G vs F+G (p=.008) when training on silence and speech.

\subsubsection{Multimodality} F+G surpasses F on average significantly (p=.001 for FR\textrightarrow FR, and p$<$.001 for WH\textrightarrow FR). We observe the same trends class-wise, with F+G obtaining the highest performance overall for \textit{neutral}, while for \textit{happy} adding G is not beneficial. With respect to SEM changes, however, we see an slight increase on average, for \textit{happy} it tends to decrease, while for \textit{pensive} it increases for all cases except for FR\textrightarrow FR when training with silence, when it decreases instead.

\subsubsection{Expanding training data including other countries} Training with WHOLE substantially increases performance on average. Statistical tests confirm significant differences for all models (p=[.01,.03] when training with silence and p$<$.01 when training with silence and speech) except G. Class-wise trends for F-based models are the same as for the evaluation with speech data, with \textit{neutral} accuracy decreasing, while the accuracy and discriminative power for \textit{happy} and \textit{pensive} increases. By contrast, for G, \textit{neutral} and \textit{happy} accuracy increase while \textit{pensive} accuracy decreases when training with silence, and it is maintained when trained on speech and silence. Standard errors for F increase on average and for \textit{neutral}, but decrease for \textit{happy} and \textit{pensive}, whereas for G, they decrease on average and for \textit{pensive}, but increase for \textit{neutral} and \textit{happy}.

\subsubsection{Expanding training data including speech instances} This scenario causes the accuracy to decrease on average for F-based models, while the opposite is observed for G. We obtain significant differences for all comparisons at different p-value levels (p$<$.05) except for G. Class-wise, the accuracy for \textit{neutral} increases while for \textit{happy} and \textit{pensive} decreases, which is the opposite than when adding data from other countries, and which follows WHOLE trends. Note that, for G, the accuracy of \textit{happy} is maintained for FR\textrightarrow FR and decreased for WH\textrightarrow FR. Confusion patterns are similar to those for SP. Standard errors also follow the same trends in general.

\subsection{NORWAY evaluation}

\begingroup
\setlength{\tabcolsep}{5pt} 
\begin{table*}[t!]
\centering
\caption{Emotion recognition results for video-based labels trained on the NORWAY (left) and NORWAY (right) training subsets under silence only or speech and silence instances, and evaluated on the NORWAY test subset under silence (train$ \rightarrow $test), reported as unweighted average accuracy $\pm$ SEM over 10 folds and 3 runs per fold. Underlined: best accuracy per training type. Italics: best accuracy overall.}
\resizebox{\textwidth}{!}{%
\begin{tabular}{@{}lcccccccccc@{}}
\toprule
\multicolumn{2}{l}{} & \multicolumn{4}{c}{\textbf{NORWAY $ \rightarrow $ NORWAY}} & \multicolumn{1}{l}{\textbf{}} & \multicolumn{4}{c}{\textbf{WHOLE $ \rightarrow $ NORWAY}} \\ \midrule
\multicolumn{1}{c}{\textbf{Modality}} & & \multicolumn{1}{c}{\textbf{\begin{tabular}[c]{@{}c@{}}Neutral\\ Accuracy\end{tabular}}} & \multicolumn{1}{c}{\textbf{\begin{tabular}[c]{@{}c@{}}Happy\\ Accuracy\end{tabular}}} & \multicolumn{1}{c}{\textbf{\begin{tabular}[c]{@{}c@{}}Pensive\\ Accuracy\end{tabular}}} & \multicolumn{1}{c}{\textbf{\begin{tabular}[c]{@{}c@{}}Average\\ Accuracy\end{tabular}}} & &  \multicolumn{1}{c}{\textbf{\begin{tabular}[c]{@{}c@{}}Neutral\\ Accuracy\end{tabular}}} & \multicolumn{1}{c}{\textbf{\begin{tabular}[c]{@{}c@{}}Happy\\ Accuracy\end{tabular}}} & \multicolumn{1}{c}{\textbf{\begin{tabular}[c]{@{}c@{}}Pensive\\ Accuracy\end{tabular}}} & \multicolumn{1}{c}{\textbf{\begin{tabular}[c]{@{}c@{}}Average\\ Accuracy\end{tabular}}} \\ \midrule
\multicolumn{5}{l}{\textit{Training on silence data:}}  \\ \midrule
F & & 80.91 $\pm$ 2.6 & \textit{\underline{55.41 $\pm$ 4.9}} & 38.3 $\pm$ 5.0 &  58.21 $\pm$ 2.8 & & 64.02 $\pm$ 3.6 & \textit{\underline{65.08 $\pm$ 4.4}} &  58.04 $\pm$ 5.9 &  62.38 $\pm$ 2.5 \\ \midrule
G & & 65.2 $\pm$ 3.6 &  37.64 $\pm$ 5.3 & \underline{62.78 $\pm$ 5.5} &  55.21 $\pm$ 2.5 & &  47.0 $\pm$ 2.2 &  58.26 $\pm$ 6.0 &   69.0 $\pm$ 5.7 &  58.09 $\pm$ 3.2 \\  \midrule
F+G & & \underline{84.78 $\pm$ 1.9} &  53.46 $\pm$ 4.7 &  46.75 $\pm$ 5.1 &  \underline{61.66 $\pm$ 2.9} & & \underline{69.08 $\pm$ 2.7} &  63.54 $\pm$ 5.2 &  \textit{\underline{71.04 $\pm$ 5.4}} & \textit{\underline{67.89 $\pm$ 2.6}} \\ \midrule
\multicolumn{5}{l}{\textit{Training on all data (speech + silence):}}  \\ \midrule
F & & 85.62 $\pm$ 1.9 &  51.65 $\pm$ 4.9 &  35.61 $\pm$ 4.9 &  57.63 $\pm$ 2.8 & & 71.36 $\pm$ 3.0 &  \underline{62.84 $\pm$ 4.7} &  52.04 $\pm$ 5.5 &  62.08 $\pm$ 2.4 \\ \midrule 
G & & 71.26 $\pm$ 3.3 &  20.84 $\pm$ 5.1 &  \textit{\underline{66.47 $\pm$ 6.2}} &  52.85 $\pm$ 2.4 & & 53.96 $\pm$ 3.2 &  58.23 $\pm$ 4.8 &  \underline{67.88 $\pm$ 5.2} &  60.02 $\pm$ 3.1 \\  \midrule
F+G & & \textit{\underline{88.33 $\pm$ 1.4}} & 50.9 $\pm$ 5.0 &  48.67 $\pm$ 4.8 &  \textit{\underline{62.63 $\pm$ 2.8}} & & \textit{\underline{77.71 $\pm$ 2.0}} &  62.69 $\pm$ 4.9 &  62.77 $\pm$ 5.5 &  \underline{67.72 $\pm$ 2.5} \\ 
\bottomrule

\end{tabular}%
}
\label{tab:video_silence_whole_norway_results}
\end{table*}
\endgroup

Table~\ref{tab:video_silence_whole_norway_results} summarizes Norwegian results. We detail them below.

\subsubsection{Main modality} Again, trends do not follow those of WHOLE, with \textit{pensive} scoring the lowest followed by \textit{happy}. Compared to the speech scenario, evaluating with silence instances substantially decreases performance for \textit{pensive}. The confusion between \textit{neutral} and the minority classes is higher than for SP but lower than for FR. In the unique case of NO, however, a proportion of \textit{pensive} instances are misclasified as \textit{happy}.

\subsubsection{Auxiliary modality} G alone obtains the highest accuracy for \textit{pensive} for all scenarios except for WH\textrightarrow NO training on silence, for which F+G slightly outperforms it. The difference between F and G on average is moderately larger than for FR but smaller than for SP. No statistically significant differences are found. Compared to the other countries, \textit{pensive} gets more misclasified with the other two classes.

\subsubsection{Multimodality} Similar trends, with statistically significant differences for all pairwise comparisons (p$<$.01) except on NO\textrightarrow NO training on silence data. Similar trends class-wise as well. The SEM increases on average and for \textit{happy}, while it decreases for \textit{neutral}.

\subsubsection{Expanding training data including other countries} Training on WHOLE consistently increases performance on average, although only the increase for G when training on speech and silence is statistically significant (p=.03). Class-wise trends are maintained from FR for F-based models, while G follows F trends instead, although the increase for \textit{pensive} is smaller. Despite the fact that training with WHOLE increases \textit{pensive} performance with G, it does not necessarily increase its discriminative power, since confusion with \textit{happy} increases. Standard errors increase for all classes with F-based models while decreasing on average, whereas with G they increase for all except for \textit{neutral}.

\subsubsection{Expanding training data including speech instances} Unlike for SP and FR, the effect of this training regime for NO is unclear, with extremely similar results on average. Only for G we observe larger differences. Class-wise trends are maintained, with \textit{neutral} performance increasing while \textit{happy} and \textit{pensive} decreases. For the latter, we find the exceptions of G and F+G for NO\textrightarrow NO, for which \textit{pensive} accuracy increases. For F-based models, we find that confusion increases between \textit{neutral} and the minority classes at a similar rate, as opposed to SP and FR, for which confusion happens mostly between \textit{neutral} and \textit{pensive}.

\end{document}

%% file: Sections/1_intro.tex
\IEEEPARstart{E}{motion} recognition plays a pivotal role in conversational human-machine interaction (HMI)~\cite{jaimes2007multimodal, mckeown2011semaine}, enabling systems to perceive and respond to users' emotional states~\cite{EMPATHIC_ANALYSIS, Vazquez2023}.
Research in affective computing has long proved the possibility of detecting emotion expressions with data-driven approaches using different input modalities, mainly linguistic, acoustic, and facial expressions~\cite{poria2017review,rouast2019deep}. Multimodal approaches have also shown promising results in enhancing recognition accuracy and robustness~\cite{d2015review}. 
The literature has shifted from recognizing acted, contextless prototypical expressions to more spontaneous reactions and in-the-wild data, which presents numerous challenges for unimodal and multimodal models~\cite{d2015review}. In addition to the increased appearance and behavioral variability, spontaneous emotions are more subtle and difficult to disambiguate, significantly differing from acted emotions in surface representation~\cite{schuller2019affective}. Consequently, the emotional space is smaller and more compact~\cite{chakraborty2017, deVelasco2022a}. In addition, natural contexts suffer from a high imbalance in emotional categories, negatively affecting the learning process of data-driven approaches. Conversational HMI users tend to exhibit less intense emotional responses than when interacting with other humans due to the limited emotional capacity of artificial agents, resulting in more neutral expressions~\cite{deVelasco2022a}.  
Visual-based recognition is further affected by the \textit{speaking effect}, for which facial deformations caused by speaking can be mistaken for emotional expressions~\cite{mariooryad2015facial}. Lastly, establishing a reliable gold standard remains difficult due to subjective perceptual annotation procedures, resulting in low agreement and potential discrepancies between emotions expressed and perceived~\cite{zeng2007survey}.

Emotion recognition research has traditionally focused on young adults. However, the global aging of the population is generating new socioeconomic challenges. Thus, older adults have been positioned as a clear beneficiary of technological development in HMI~\cite{EMPATHIC_ANALYSIS, CITA_GO_ON_PETRA, McTear_VITA2023, demaeght2022multimodal}. Understanding emotions in older adults is of special importance, as it can lead to the development of customized virtual coaching applications and companions, and healthcare technologies that foster active aging and independent living. This presents additional computational challenges, since aging changes facial features, voices, and speaking styles, among others~\cite{magai2006emotion, Folster2014agevary}. For example, a higher intensity in facial expressions and speech is associated with a higher emotion recognition accuracy; however, older subjects display less intense vocal and facial expressions compared to younger subjects~\cite{levenson1991emotion, ma2019elderreact}. Therefore, models trained in other age groups do not perform optimally for recognition in older adults~\cite{ma2019elderreact}, and models trained on data from this age group tend to show lower performance than other groups~\cite{wang2015speech, lopes2018facial}.
The lack of public databases including older adults hinders progress on this front. To our knowledge, only two datasets that focus on older adults provide non-acted data: the ComParE Elderly Emotion Sub-Challenge speech dataset~\cite{schuller2020interspeech}, which includes personal narratives; and ElderReact~\cite{ma2019elderreact}, containing monologue videos of older adults reacting to specific items spontaneously, but possibly exaggerating their responses due to the nature of the videos.
Very few non-acted, interaction-oriented datasets include older adults~\cite{kossaifi2019sewa}, and we are unaware of any HMI dataset featuring them.

This paper presents a comprehensive study on computational, non-verbal discrete emotion expression recognition in interactions between older adults and a simulated Virtual Coach (VC), as a specific case of HMI scenario. 
The work was developed as part of the European EMPATHIC project~\cite{Empathic_Midterm,Empathic_Demo}, which aimed to explore and validate new interaction paradigms for empathic, expressive, and advanced VCs to improve independent, healthy-life-years of this age group. As part of the project, 157 participants over 65 years old from three countries (Spain, France, and Norway) were recorded interacting with an initial version of the EMPATHIC-VC in a Wizard of Oz (WoZ) paradigm. This data aided in system development and in the study of the interaction between older adults and VCs\footnote{The EMPATHIC data corpus can be found in the European Language Resources Association (ELRA) catalogue (\url{http://catalogue.elra.info/en-us/}): Corpus ISLRN: 631-345-309-445-9 and ELRA ID: ELRA-S0414.}. Under this framework, we first describe the emotion annotation procedure and methodological choices tailored to the project requirements. Then, using a deep learning-based approach, we investigate the contribution of different modalities for emotion recognition, including speech, facial expressions, eye gaze, and head dynamics, individually and combined, in various evaluation scenarios.

This framework provides two unique features that we exploit in our work. First, it allows us to perform a comparative analysis across cultures (where culture influences not only emotional expression but also the annotation procedure) and languages as well as multi-country versus country-specific training, two aspects that have received limited attention~\cite{kossaifi2019sewa}. Second, the project involved a channel-specific annotation, providing two distinct sets of emotional labels, audio- and video-based (explained in Sec.~\ref{sec:data}). Thus, we consider speech from audio and facial expressions from video as main modalities, while eye gaze and head movements act as additional modalities that can also be extracted from video. We assess the effectiveness of the main modalities in recognizing their associated labels, and the possible performance improvement when being combined with the remaining (\textit{auxiliary}) modalities. Additionally, we conduct a cross-channel evaluation, wherein the main modality of one label type and the auxiliary modalities are individually employed to recognize the labels derived from the other channel. This offers insight into the transferability and adaptability of modalities across label types, possibly revealing shared emotional cues. Lastly, we analyze performance differences between training and evaluating on spoken and silent instances, to understand how the presence or absence of speech affects performance for this age group.

This paper is organized as follows. Sec.~\ref{sec:related} reviews current trends in affective computing for emotion recognition with the modalities considered in this work. Sec.~\ref{sec:data} describes the corpus and the annotation protocol. Sec.~\ref{sec:method} details our computational approach. Sec.~\ref{sec:evaluation} describes the evaluation protocol and results of all evaluation scenarios, which are discussed in Sec.~\ref{sec:discussion}. Finally, Sec.~\ref{sec:conclusion} concludes the paper.

%% file: Sections/2_related.tex
In this section, we first summarize the two main models of emotion used in affective computing. Second, we review related computational approaches for the automatic recognition of emotional states from speech, facial expressions, gaze, and head cues. Finally, we discuss multimodal approaches using such cues, with an emphasis on works featuring older adults.

\subsection{Models of emotion}

Expressions of emotion are generally represented by two main different models: a categorical or discrete model, and a dimensional or continuous model. The categorical model identifies a set of discrete emotional categories, ranging from the basic Ekman emotions~\cite{ekman1999basic} (\textit{happy}, \textit{surprised}, \textit{contempt}, \textit{sad}, \textit{fearful}, \textit{disgusted}, and \textit{angry}), to a larger set with more specific and realistic affective states. Indeed, ordinary communication involves a variety of complex feelings that cannot be characterized by a reduced, fixed set of categories~\cite{gunes2010automatic}. 
Therefore, such categories are usually selected considering the task at hand. For instance, categories such as \textit{bored}, \textit{frustrated}, \textit{delighted}, \textit{calm}, \textit{satisfied}, or \textit{excited} are more applicable to HMI scenarios than most basic states~\cite{calvo2010affect}. 

Given the complexity of the emotional semantic space, a number of researchers~\cite{gunes2010automatic,schuller2011recognising} are more in favour of adopting a dimensional model such as the circumplex model of affect~\cite{russell1980circumplex}. In the dimensional model, each affective state is represented by a point in a two-dimensional space, where the \textit{valence} dimension represents the polarity of the emotion, i.e., a positive or negative value along a continuum, and the \textit{arousal} dimension represents the degree of emotional activation. i.e. values vary from low to high along a continuum. Other versions include a third dimension, \textit{dominance}, which represents a sense of control over the situation while experiencing the emotion. The Valence, Arousal, and Dominance (VAD) model has been widely exploited for audio/video-based emotion recognition~\cite{Valstar:2014,gunes2010automatic, deVelasco2022a}, allowing for the encoding of slight emotional changes over time~\cite{Valstar:2014}.

Both models have their own advantages and drawbacks. For instance, emotional categories may not account for intensity and exhibit fuzzy boundaries. Conversely, dimensional models introduce more subjectivity in emotion scaling across raters. Ultimately, the choice depends on the objectives of the task. In our case, we are interested in detecting prespecified events of interest that are expected to occur during the interaction, for which the EMPATHIC-VC system can react and adapt to, in a practical and interpretable way. Thus, the categorical model better fits the needs of the system.

\subsection{Emotions from speech}
\input{Sections/2_related_audio.tex}

\subsection{Emotions from facial expressions}
\input{Sections/2_related_faces.tex}

\subsection{Emotions from eye gaze and head pose}
\input{Sections/2_related_gaze.tex}

\subsection{Multimodal emotion recognition}
\input{Sections/2_related_multimodal.tex}

%% file: Sections/2_related_audio.tex
The speech signal captures the speaker's communicative intention, encompassing not only the words spoken but also the intonation, prosody, pauses, and other paralinguistic elements that contribute to the message. In the same way, speech provides a lot of information about the speaker, their accent, profile, speaking style, and current emotional state~\cite{huang2019speech, deVelasco_tesis}. 

The most commonly used features for speech emotion recognition (SER) are based on Low-Level Descriptors (LLDs), such as zero crossing rates, pitch, formants, energy, jitter, shimmer, spectral centroids, Mel-Frequency Cepstral Coefficients (MFCC), flux, etc., as well as on their descriptive statistics or functionals (e.g., mean, SD, quartiles)~\cite{huang2019speech, PandaAudioFeatures, deVelasco2022a}. Some works have proposed the standardization of the feature sets. However, only GeMAPS, which contains a combination of the previously mentioned LLDs and functionals, and the feature sets proposed in the ComPaRE challenge series, which are variations of GeMAPS, have become a reference~\cite{schuller2013interspeech, eyben2015}. Spectrograms have also been used as a sequence of features represented as an image, which has been demonstrated to be specifically useful to feed Convolutional Neural Networks (CNNs)~\cite{deVelasco_tesis}.
More recently, the first framework for self-learning rich representations of speech was published, called Wav2Vec~\cite{Baevski2020}, which was initially used for SER in English by~\cite{Luna-Jimenez2021} and in Spanish by~\cite{deVelasco2022a}. Shortly afterward, new frameworks were proposed, including Hubert~\cite{Hsu2021}, UniSpeech~\cite{wang2021unispeech}, and WavLM~\cite{chen2022wavlm}. A comparison of self-supervised representations for SER can be found in~\cite{deVelasco_tesis}.

Similarly to other domains, the rise of deep learning also caused a gradual transition from traditional classifiers to deep neural networks (DNNs) for SER~\cite{singh2022, deLope2023review}.  
Current approaches feature Multilayer Perceptrons (MLP)~\cite{deVelasco_tesis}, CNNs~\cite{deVelasco2022a}, Recurrent Neural Networks~\cite{Wang2020, deVelasco_tesis}, Transformers~\cite{Morais2022, wagner2023dawn}, and combinations of different DNNs~\cite{atmaja2022survey,deVelasco2022a}.

%% file: Sections/2_related_faces.tex
Facial expressions are considered one of the most significant means for humans to express their emotions and intentions in their daily communication~\cite{ekman1999basic}. 
Facial expression recognition (FER) systems can be divided into two main categories according to the type of facial input they rely on: static-image FER and dynamic-sequence FER ~\cite{Li_Deng2018}. In static-based methods, the feature representation is based only on the spatial information associated with a single image, whereas dynamic-based methods consider the temporal relation among contiguous frames as well as the facial deformation dynamics. 

In turn, and similarly to SER, FER approaches can also be divided into conventional and deep learning based approaches.
The former is usually composed of three major steps: face and landmarks detection, feature extraction, and emotion classification. These conventional algorithms usually extract face-based handcrafted features such as pixel intensities~\cite{Mohammadi2014}, Gabor filters~\cite{ChengjunLiu2002}, local binary patterns~\cite{Shan2009}, and histograms of oriented gradients~\cite{Mavadati2013}. However, these often lack enough generalizability in in-the-wild settings.
By contrast, deep learning-based approaches are used as a conjoint feature extraction tool and facial expression classifier, reducing the dependency on preprocessing techniques and human expertise-based feature extraction. The same neural network approaches discussed for SER have been applied to FER with similar results; hence, we avoid repeating them here. We refer the reader to the surveys of~\cite{corneanu2016survey, Li_Deng2018, ko2018brief} for a comprehensive review of the state of the art. As an example of approach related to the one used in this paper, we highlight the work of
~\cite{mollahosseini2016going}, one of the first to demonstrate the capability of CNNs to recognize the Ekman emotions by outperforming traditional methods on popular posed and spontaneous expression datasets.

%% file: Sections/2_related_gaze.tex
Extensive behavioral and neuroscience literature has confirmed a relationship between eye state, gaze direction, and facial expressions on the perception of emotions and mental states~\cite{graham2012neurocognitive}. 
Eye-related features that have been studied or used the most in affective computing are: pupil size, blinks, gaze direction, direct/averted gaze, extracted patterns of gaze events, and eye aperture/closure~\cite{lim2020emotion, o2018affective}.
Dedicated eye trackers are generally required to extract these features with high accuracy. However, this is impractical for many everyday scenarios or HMI settings, such as the EMPATHIC-VC, where a non-obtrusive or lower-cost approach is preferred. For such scenarios, regular cameras can now be used to estimate eye gaze and approximate the location of pupil and eye landmarks by means of appearance- or model-based methods~\cite{zhang2019evaluation}. In particular, appearance-based gaze estimation has improved significantly during the past decade, boosted by deep learning advances~\cite{ghosh2021automatic}. Therefore, a number of works compute statistical features, or functionals, from the raw or smoothed estimated gaze trajectories over a time window, or compute features (e.g., eye closure, pupil size) based on specific eye landmarks instead~\cite{alghowinem2016multimodal, o2018affective, abdou2022gaze}. Blinks can usually be detected via dedicated appearance-based methods~\cite{cortacero2019rt}, or by detecting the action unit (AU) \#45.

On a related note, head rotation plays an important role in stabilizing gaze to fixate on objects of interest~\cite{guitton1987gaze}. 
There is evidence of the relationship between head pose dynamics and expression and perception of different emotional and mental states~\cite{el2005real, hess2007looking, busso2007rigid}, being particularly related to emotional intensity~\cite{karg2013body}. Affect recognition works have relied on head pose categorizations such as head tilts, nods, and shakes~\cite{el2005real, gunes2010dimensional}, which usually require specific action detectors. More recent approaches directly use temporal 3D rotational angles (yaw, pitch, and roll) to describe head motion trajectories, as well as angular displacement, velocity, acceleration, and window-based functionals computed from such trajectories~\cite{adams2015decoupling, alghowinem2016multimodal, samanta2017role}, dynamic features based on the discrete Fourier transform~\cite{ding2018low}, or clustered sequences of kinemes~\cite{samanta2020emotion}. Head orientation is generally extracted with appearance-based methods and model-based 3D head registration~\cite{khan2021head}.

Due to their relationship, a handful of works have combined head and gaze features together for emotion recognition~\cite{xue2021investigating}. Although these features have been proven to be sufficient for specific affective states in some scenarios~\cite{o2019eye, samanta2020emotion}, they are usually added to facial or speech modalities to provide complementary rather than redundant information.

%% file: Sections/2_related_multimodal.tex
With significant advancements in multimodal machine learning, multimodal emotion recognition has gained considerable momentum lately (see~\cite{ramachandram2017deep, baltruvsaitis2018multimodal, guo2019deep} for exhaustive surveys on multimodal machine learning, and~\cite{poria2017review, zeng2007survey, d2015review, rouast2019deep} for multimodal emotion recognition). By leveraging the complementary information of multiple modalities, multimodal systems can achieve higher accuracy and reliability compared to unimodal systems.
Multimodal fusion methods are broadly classified into feature-based, decision-based, and hybrid approaches. The former consists in combining the features extracted from different modalities, with methods that range from naive feature concatenation to attention-based approaches. Feature-based fusion allows learning from cross-modal correlations; however, an alignment among modalities is required since they may have different sampling rates or representations (e.g., video frames vs audio segments), and not all modalities may be available at all times. Instead, decision-based fusion combines the scores or predictions of unimodal models for a final multimodal prediction, thus alleviating the alignment and incomplete data problems but disregarding cross-modal correlations. Lastly, hybrid approaches combine feature- and decision-based fusion. Generally, the best fusion type is task- and dataset-dependent.

Most multimodal emotion recognition works combine at least paralinguistic and facial expression features~\cite{wu2014survey}, or acoustic and linguistic~\cite{atmaja2022survey}.
Gaze and head cues are usually combined with other features like facial information~\cite{cohn2004multimodal, wu2019continuous} and/or speech cues~\cite{alghowinem2016multimodal, alhargan2017multimodal, o2019speech}. Nonetheless, their use is less explored compared to audiovisual fusion. One of the few works that combine speech, facial expressions, and gaze features is that of~\cite{abdou2022gaze}, which uses audio GeMAPS features, and a subset of the gaze functionals proposed by~\cite{o2019eye} and facial features extracted from a pre-trained CNN from video.

The ComParE challenge recently drew attention to emotion recognition for older adults, in which challenge participants could use acoustic and linguistic features~\cite{schuller2020interspeech}. However, the work of~\cite{ma2019elderreact} is one of the few addressing discrete emotion recognition using the modalities considered in this work for such age group, based on ElderReact. More specifically, they extract gaze and head features, facial AUs (including blink), and facial landmarks from video, and voice quality, MFCCs, and prosody features from audio.

%% file: Sections/3_data.tex
Here, we describe the subset of the EMPATHIC WoZ Corpus considered for this work, and the protocol followed for the annotation of audio and facial expressions from video.

\subsection{Data collection}

The target population of the EMPATHIC project was defined as healthy older adults based on the following inclusion criteria: 1) above the age of 65 or turning 65 in 2019; 2) good hearing and sight (with or without glasses/hearing aid); 3) living independently at home; and 4) read, write, and speak the testing language fluently. Recruitment\footnote{This protocol was approved by appropriate Institutional Review Boards of Spain (Ethical and Scientific Research Committee of University of the Basque Country -UPV/EHU. Cod: PI2018152), France (Commission Nationale de l'Informatique et des Libertés-CNIL- Cod: 2182146), and Norway (Ethical and Scientific Research Committee of the Oslo University Hospital).} involved participants from Spain, France and Norway. A total of 157 participants (105 female) were recruited and participated in the first recording sessions of the project, of which 153 are included in the corpus. Participants are distributed as follows: 78 Spanish (54 female, mean age 69.5), 44 French (28 female, mean age 73.5), 31 Norwegian (21 female, mean age 74.8). The overall mean age was 71.8 years (SD=$\pm$6.8). All participants were properly informed and signed an informed consent prior to enrolling on the study. Hereinafter, we refer to the Spanish subset as SP, the French as FR, the Norwegian as NO, and the complete data as WHOLE (or WH).

\begin{figure}[t!]
\centering
\includegraphics[width=0.8\linewidth]{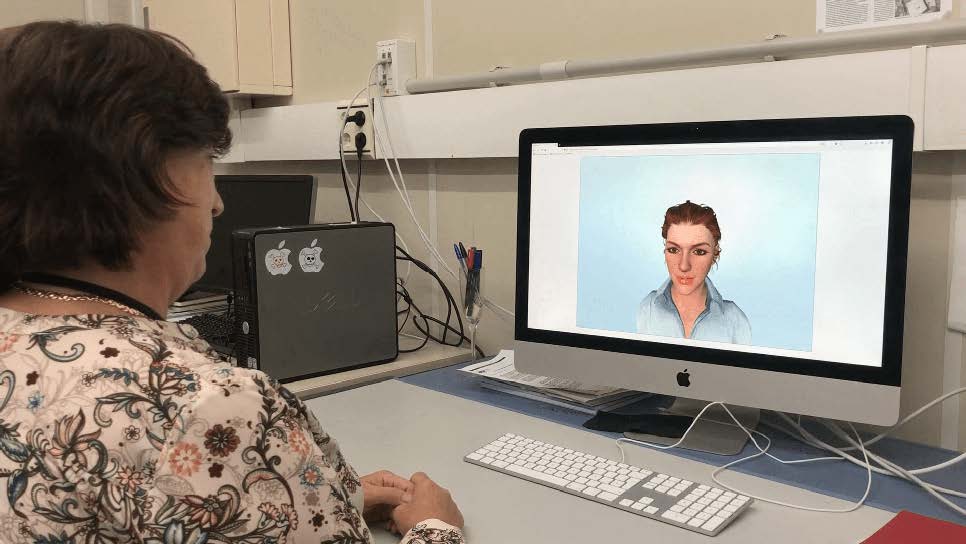}
\caption{Setup with a participant during an interaction session.}
\label{fig:empathic_example}
\end{figure}

We used the WoZ paradigm for data acquisition, commonly used when building systems based on natural language and other artificial intelligence-driven applications~\cite{Schogl2015}. Its key principle is that study participants believe they are interacting with an autonomous system, while actually the actions of the system are controlled by a human (i.e., the \textit{wizard}). This wizard is usually in a different room and connected to the study setting through a network connection.  
The interaction sessions combined different questionnaires and interaction with the EMPATHIC-VC, detailed in~\cite{EMPATHIC_ANALYSIS}. The setup consisted of a computer equipped with a webcam, a microphone, and an Internet connection (see Fig.~\ref{fig:empathic_example}). At the start of the session, participants chose one of five available visual representations of agents for their VC session. 
During the interaction, participants were alone with the VC to avoid bias or undesired interactions with the supervisor. Two dialogues of 5–10 min each were completed. The first served as an introduction to the system and thus did not focus on any specific issues. The second focused on the user's nutrition habits and goals. 

%% file: Sections/4_labeling.tex
\subsection{Definition of labels}

Our emotional labels correspond to the users' perceived expression of emotion. The procedure for selecting such emotion categories followed a three-step data-driven approach.

First, we considered the 27 categories defined in~\cite{cowen2017self}, which are based on the self-reported emotional states elicited by around 200 short videos over a population of nearly 1000 people. The list defines a rich semantic space of emotions, which includes categories such as \textit{amusement} that were found to capture well the subjective emotional experience.

As a second step, we removed the categories that were highly unlikely to be encountered during the interaction between the user and the EMPATHIC-VC, and added some labels that might potentially be perceived in these interactions. We worked with the target languages simultaneously, i.e., Spanish, French, and Norwegian, to provide accurate terms to express the same feelings in different languages, considering that cultural context can be accounted for by the translation that the native speakers of each language can provide relative to the \textit{Lingua Franca} (which in our case was English). The selected 18 labels were: \textit{relieved}, \textit{bored}, \textit{excited}, \textit{calm}, \textit{sad}, \textit{amused}, \textit{puzzled}, \textit{pleased}, \textit{interested}, \textit{tense}, \textit{surprised}, \textit{concerned}, \textit{enthusiastic}, \textit{skeptical}, \textit{embarrassed}, \textit{tired}, \textit{delighted}, and \textit{annoyed}.

Finally, we ran a set of pilot experiments on SP. Our goal was two-fold: 1) shorten the previous list by only considering the subset of emotions perceived during the interaction; 2) assess to what extent we could match totally or partially this list to the list of basic emotions defined by Ekman, which are typically featured in visual-based discrete emotion recognition datasets. This pilot, as well as the posterior results of the annotation procedures, defined the final labels to be considered for audio and video channels, which are presented next.

\subsection{Annotation protocol}\label{subsec:anota}

Few works are found in the literature aimed at establishing the amount of emotional information provided through the different audio and video channels. In particular, the study of such channels separately and their combination concludes that the latter does not always yield the best perception results, as might be otherwise expected (e.g.,~\cite{kossaifi2019sewa}). Previous studies have established that the emotional information provided by each channel or combination strongly depends on the specific emotion, context, and language~\cite{esposito2009}. For instance, there is vast evidence in the literature that, with respect to dimensional models, arousal can be better detected from the audio channel, while valence is much better estimated from the video channel~\cite{russell2003facial}. It has also been suggested that humans, when posed with the task of decoding emotional states, selectively attend to emotional cues that align closely with their personal and cultural experiences, thus minimizing the cognitive effort required for emotional processing during the task. Consequently, when annotating perceived emotional expressions from audio and video simultaneously, raters tend to explore the most familiar channel: if rater and rated person are culturally and/or language akin, the rater tends to exploit the auditory signal, whereas when they are culturally distant, they tend to rely more on visual cues~\cite{riviello2012cross}. 

One of the salient attributes considered in the EMPATHIC project is culture and cultural differences, so it was important that the annotation be carried out separately per country by native speakers to be able to capture subtle culture-specific emotional cues. Therefore, in order to avoid the annotators' reliance towards a single channel, we decided to separate channels at the annotation level, having different annotators for each channel. This, in turn, results in a richer variety of emotional information from different perceptual channels, which can be later leveraged by the EMPATHIC-VC system. 
We employed instructed annotators to be able to control the whole procedure and update it if necessary. Preliminary trials showed that annotators preferred having access to the entire video or speech file instead of annotating isolated snippets due to the presence of context, which helped them make more accurate estimations of the users' emotional state.

The annotation process consisted in determining the start and end times of all events associated to given emotions categories throughout a WoZ interaction. To ensure a high inter-rater agreement, we employed a sequential annotation process. Initially, each annotator received a set of files to annotate independently. Subsequently, the within-country inter-rater agreement was calculated with an ad-hoc measure based on event overlap. If the agreement score fell below a predefined threshold, annotators engaged in discussions and re-annotated the files. When the threshold was met, the annotators received the remaining files and continued the process of discussing and re-annotating until the desired level of agreement was attained.

\subsubsection{Audio annotations} \label{sec:audio_annotations}

These were carried out by listening to the audio signal in Transcriber\footnote{\url{https://transcriber.en.softonic.com/}} with nine native annotators (three per country). For the specific case of audio, the perceived emotions were labeled in terms of both, the categorical and the VAD models. The categorical labels were: \textit{calm/tired/bored}, \textit{pleased/amused}, \textit{puzzled}, \textit{sad}, and \textit{tense}. The first two labels consist of a combination of similar categories, which was decided after the first annotation rounds as they were highly confused among annotators. For simplicity, we henceforth refer to them as \textit{calm} and \textit{pleased}. Parts of the audio signal with no annotated label are  not categorized. The labels assigned to the dimensional VAD model were also discretized for simplicity, and defined as: 1) \textit{positive}, \textit{neutral}, and \textit{negative}, for valence; 2) \textit{excited}, \textit{slightly excited}, and \textit{neutral}, for arousal; and 3) \textit{dominant}, \textit{neither dominant nor dominated}, and \textit{defensive}, for dominance.

\begin{figure}[t!]
\includegraphics[width=0.95\linewidth]{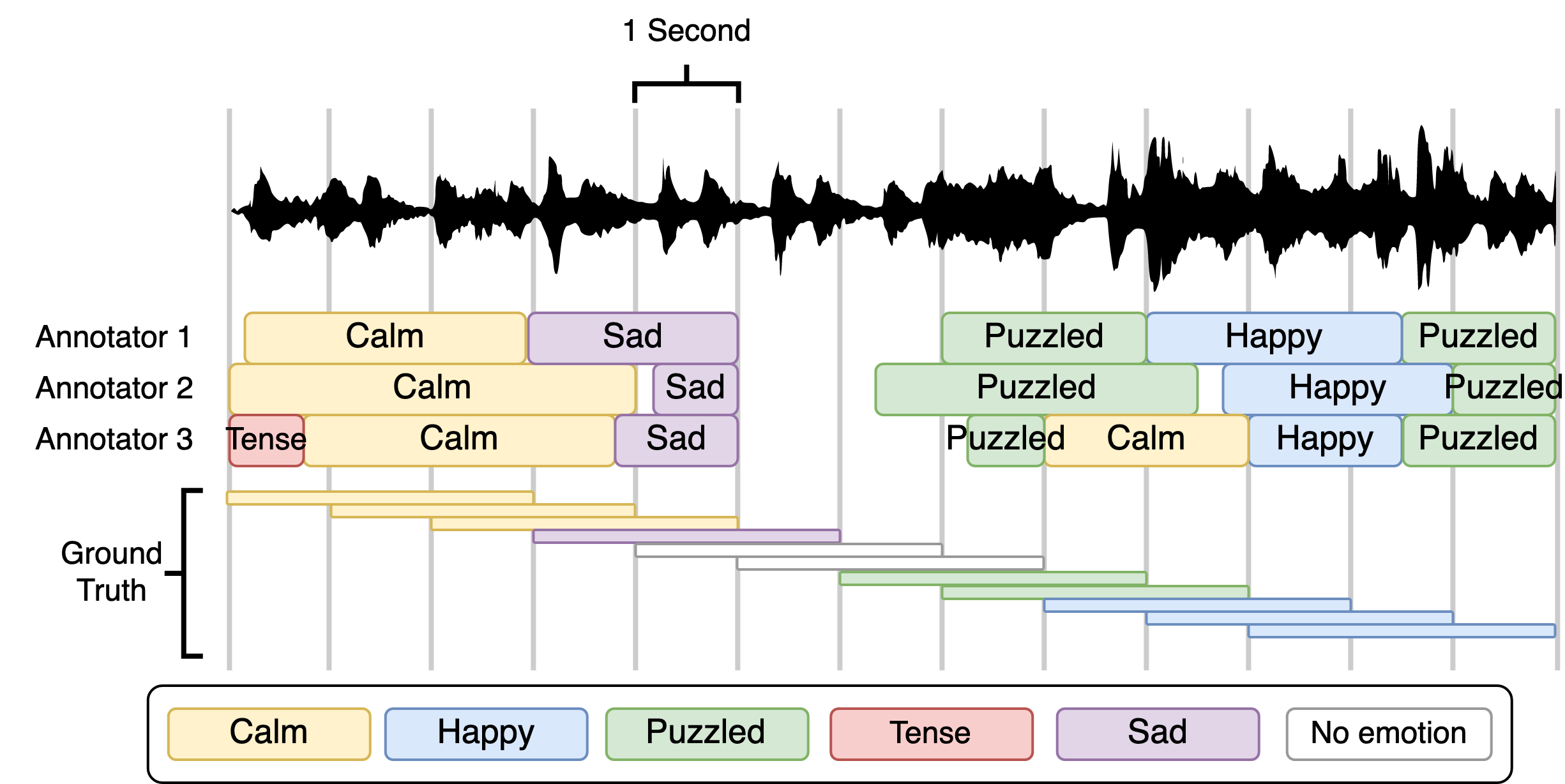}
\caption{Representation of the segmentation of annotated emotion expression categories to create the gold standard for the audio modality. \textit{Happy} corresponds to \textit{pleased/amused}.}
\label{audio_segmentation}
\end{figure}

The inter-annotator agreement for the categorical annotation was computed with Cohen's Kappa for each pair of annotators at the millisecond level. SP and NO scored an average coefficient of 0.792 and 0.692, respectively, which indicate substantial agreement~\cite{mchugh2012interrater}, while FR scored 0.554, indicating moderate agreement. Once the entire corpus was labeled, we combined the raters annotations 
using 3-s segments with a 1 s stride (see Fig.~\ref{audio_segmentation}). To assign an emotion to each segment (i.e., the gold standard, or \textit{ground truth}), the majority emotion was assigned if that emotion spanned a specific percentage of the whole segment. Otherwise, the segment was left without annotation, referred to as \textit{discarded}. Thus, each segment has four different annotations: one categorical and three for VAD.

\subsubsection{Video annotations} \label{sec:video_annotations}

The annotation was carried out with an in-house software by six native annotators, two from each country. Annotators were instructed to watch muted videos, taking into account only facial expressions and head movements, thus disregarding out-of-face information such as body or hand movements. In addition, they were instructed to watch a short snippet of each user's video (up to 1 min) to familiarize with the user's baseline facial expression. For video annotations specifically, a cross-country calibration was performed after the first set of files was annotated for a small, random subset of videos. This was performed to ensure a common understanding of the instructions, and of the minimum intensity an expression should have to be categorized as such, which can be more objectively determined across countries than for audio-based annotations.

The categorical labels considered were: \textit{sad}, \textit{annoyed/angry} (henceforth referred to as \textit{angry}), \textit{surprised}, \textit{happy/amused} (henceforth referred to as \textit{happy}), \textit{pensive}, and \textit{other}. The first four are part of the Ekman’s basic emotions~\cite{ekman1999basic}.
~\textit{Pensive} is a mental state rather than an emotional expression; however, it was included in our model as it was found to be a frequent facial expression during the conversation when users were preparing their response, as in previous HMI-oriented works~\cite{steininger2002development, el2005real}.
Similarly to audio-based annotations, some categories were combined into a single label due to being often confused by annotators. Raters were instructed to annotate as one of the first five categories those segments in which it was clear to them that the expression was present simultaneously. The label \textit{other} was used to denote those segments in which an expression was occurring but was not included in our expression list, or when more than one expression from the list was present. Finally, all non-labeled instances were considered to be a \textit{neutral} expression, denoting the baseline face as well as calmed, quiet, or very subtle emotional expressions which do not exceed the consensual expression thresholds.

Post-hoc inter-rater reliability was computed at frame level by means of Cohen's kappa coefficient, achieving a value of 0.7 for SP and FR, and 0.68 for NO, which indicates substantial agreement~\cite{mchugh2012interrater}.
We used the intersection between the two annotators to create the final gold standard. Frames with no intersection were discarded for automatic processing, representing around 8\% of the total amount of frames. 

\subsection{Analysis of labels}

\begin{table}[t!]
\centering
\caption{Number of audio segments extracted from speech emotional annotations of the EMPATHIC WoZ Corpus.}
\resizebox{\columnwidth}{!}{%
\begin{tabular}{@{}lccccccc@{}}
\toprule
 & \multicolumn{1}{l}{\textbf{Calm}} & \multicolumn{1}{l}{\textbf{Pleased}} & \multicolumn{1}{l}{\textbf{Puzzled}} & \multicolumn{1}{l}{\textbf{Sad}} & \multicolumn{1}{l}{\textbf{Tense}} & \multicolumn{1}{l}{\textbf{Discarded}} & \multicolumn{1}{l}{\textbf{Silence}} \\ \midrule
\textbf{Spain} & 38359 & 833 & 1022 & 151 & 81 & 4607 & 37910 \\
\textbf{France} & 19875 & 445 & 453 & 1 & 11 & 2819 & 15978 \\
\textbf{Norway} & 13960 & 474 & 44 & 0 & 0 & 1775 & 15764 \\ \bottomrule
\label{tab:audio_annots}
\end{tabular}%
}
\end{table}

\begin{table}[t!]
\centering
\caption{Number of frames extracted from the video emotional annotations of the EMPATHIC WoZ Corpus, corresponding to spoken (top) and silence instances (bottom).}
\resizebox{\columnwidth}{!}{%
\begin{tabular}{@{}lccccccc@{}}
\toprule
 & \textbf{Neutral} & \textbf{Happy} & \textbf{Pensive} & \textbf{Surprise} & \textbf{Angry} & \textbf{Sad} & \textbf{Other} \\ \midrule
\multirow{2}{*}{\textbf{Spain}} & 864112 & 8163 & 186735 & 115 & 0 & 0 & 0 \\
 & 824162 & 3028 & 13427 & 56 & 0 & 0 & 0 \\ \midrule
\multirow{2}{*}{\textbf{France}} & 484061 & 28118 & 98646 & 162 & 103 & 0 & 693 \\
 & 421980 & 14385 & 12915 & 107 & 28 & 0 & 278 \\ \midrule
\multirow{2}{*}{\textbf{Norway}} & 345876 & 11945 & 67859 & 72 & 0 & 0 & 239 \\
 & 317166 & 6253 & 17303 & 68 & 0 & 0 & 415 \\ 
\bottomrule
\label{tab:video_annots}
\end{tabular}%
}
\end{table}

A thorough analysis of corpus annotations is reported in~\cite{greco2021emotional}. Here, we summarize the findings, with an emphasis on the categorical labels that will be used in our evaluation.

The number of final audio segments per emotion category is detailed in Table~\ref{tab:audio_annots}. As can be seen, \textit{calm} is the most frequent emotion with around 95\% of the samples, with respect to instances where the user is speaking and disregarding \textit{discarded}, whereas \textit{sad} and \textit{tense} are quasi absent. Specifically for NO, users rarely showed a \textit{puzzled} expression. With regards to the VAD model, we highlight the following differences: 1) around 30\% of FR segments and only 3-4\% of SP and NO segments are marked with \textit{slightly excited} for the arousal dimension, while the rest is \textit{neutral}; 2) SP segments are mostly divided between \textit{positive} and \textit{neutral} valence; 3) about 25\% of FR segments have \textit{positive} valence, while for NO they are mainly \textit{neutral}; and 4) participants in the three datasets are often \textit{neither dominant nor dominated}.

Table~\ref{tab:video_annots} provides the distribution of emotion categories from video corresponding to spoken (top) and silence (bottom) instances separately. The reported quantities do not include the 0.3\% of frames that are not matched to any audio segment, which mainly happened at the end of the video due to audio-video length mismatch. Similarly to audio annotations, video annotations lead to highly imbalanced results. \textit{Pensive} was the most frequent manually labeled expression, appearing 11\% of the time, followed by \textit{happy}, present in 2\% of the total images. Despite these findings, the \textit{neutral} category clearly dominates over all categories, appearing around 87\% of the time.

As observed, the main challenges encountered in the EMPATHIC WoZ corpus are: 1) the imbalance between the different emotion classes; and 2) the imbalanced number of subjects across countries and limited data samples, particularly for audio. The former indicates that the interaction with the VC did not lead the users to experience strong emotions like \textit{sad}, \textit{angry} and \textit{surprise}, and is in line with what it is usually observed in real, spontaneous HMI interactions. In addition, many of the users may have an \textit{a priory} positive attitude since they are volunteers to participate in the experiment. The reduced number of audio samples is partly caused by the amount of time that participants had to wait for the WoZ to respond. The high class imbalance is a problem for data-driven models to properly learn any discriminative information for the minority classes. Hence, for this study, we reduce the number of categories to the three most represented for each label type. That is, for audio, we maintain \textit{calm}, \textit{pleased}, and \textit{puzzled}, whereas for video we keep \textit{neutral}, \textit{happy}, and \textit{pensive}.

\begin{table*}[t!]
\centering
\caption{Contingency table for audio-video labels. Percentage computed over the total of rows and columns per country, using the audio segments as the unit of measure.}
\begin{tabular}{@{}cccccccccccc@{}} 
\toprule
 & \multicolumn{3}{c}{\textbf{Spain}} & & \multicolumn{3}{c}{\textbf{France}} & &  \multicolumn{3}{c}{\textbf{Norway}} \\ \midrule
 & \textbf{Neutral} & \textbf{Happy} & \textbf{Pensive} & & \textbf{Neutral} & \textbf{Happy} & \textbf{Pensive} & &\textbf{Neutral} & \textbf{Happy} & \textbf{Pensive} \\
 \midrule
 \textbf{Calm} & 78.06 & 0.34 & 17.08 & & 76.18 & 3.25 & 16.19 & & 78.60 & 1.27 & 16.63 \\
\textbf{Pleased} & 1.34	& 0.36 & 0.18 & & 1.21 & 0.87 & 0.06 & & 2.25 & 0.84 & 0.12 \\
\textbf{Puzzled} & 2.07 & 0.004 & 0.57 & & 1.65 & 0.02 & 0.58 & &  0.21 & 0 & 0.08 \\ 
\bottomrule
\end{tabular}
\label{tab:audio-video}
\end{table*}

Table~\ref{tab:audio-video} depicts the relationship among audio-video labels per country, using the audio segments as reference and computing the most repeated video category for the valid frames within the start-end times of an audio segment. We find that audio-based \textit{calm} and video-based \textit{neutral} coincide 76-79\% of the time (66-70\% if including all labels). However, there is no evident one-to-one correspondence for the remaining cases. Given that each channel contributes distinct information, we retain the two label types as independent entities, allowing the system to estimate both of them at each time step.

%% file: Sections/5_methodology.tex
In this section, we describe our methodology and training strategy for data-driven recognition of emotional states using different modalities. The methodological choices depend on the EMPATHIC-VC system requirements, which follow those of common multi-agent systems~\cite{jaimes2007multimodal}. The \textit{human sensing} module, which includes emotion recognition, is one of the multiple modules that must communicate timely with a \textit{dialogue manager}. The manager controls the conversation flow by integrating the information from \textit{human sensing} and other modules to transfer the appropriate VC reactions to the \textit{natural language generation} and avatar animation modules~\cite{Empathic_Demo, Vazquez2023}. In the final system, some modules would be located in remote servers and thus data transfers would be done via network. Thus, efficiency in the whole process is crucial to ensure a seamless and natural interaction. Therefore, we prioritize independent, lightweight computational submodules for each channel, which can operate asynchronously and produce estimates at the lowest granularity level for further processing.

\begin{figure}[t!]
    \centering
    \includegraphics[width=\linewidth]{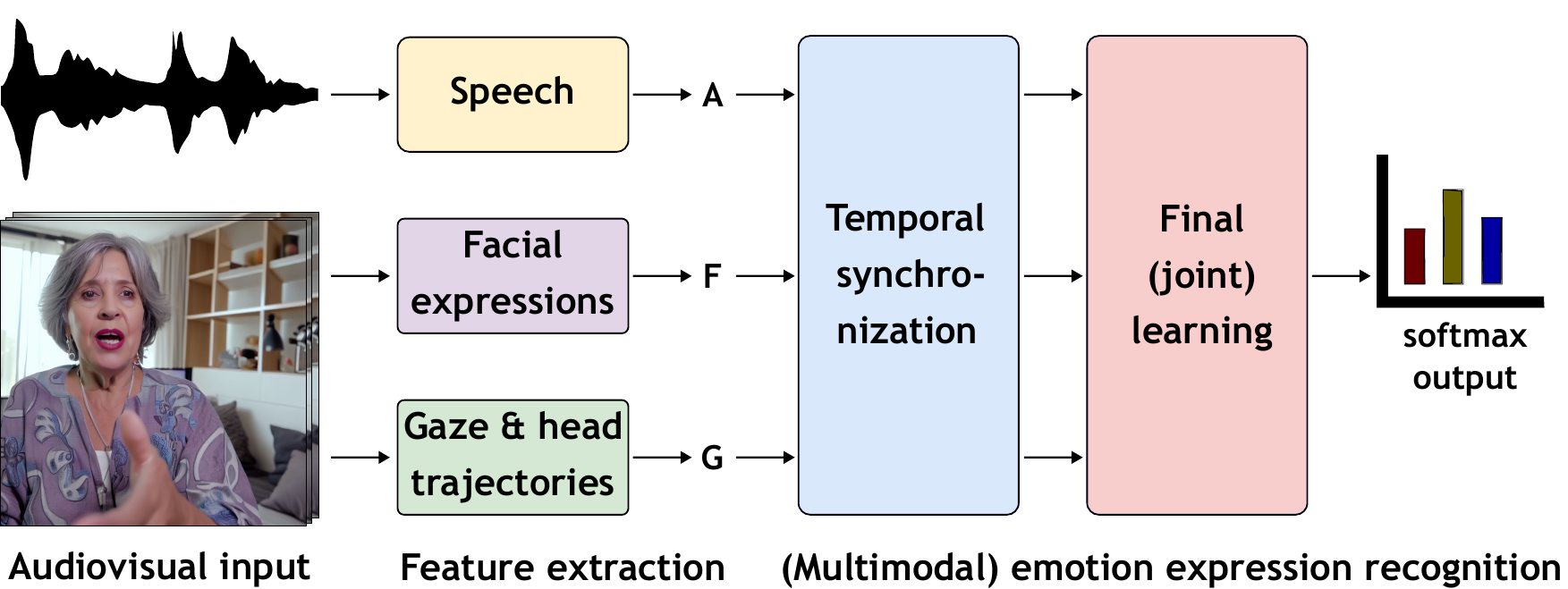}
    \caption{Overview of the methodological pipeline.}
    \label{fig:pipeline}
\end{figure}

Fig.~\ref{fig:pipeline} shows an overview of our methodological pipeline. In summary, we first extract features from the different modalities. More specifically, for the main modalities (i.e., speech from audio and facial expressions from video), we train individual models for their respective labels on all the available data to learn rich emotional features. In parallel, we extract additional features from video, namely \textit{looking-at-VC}, head, 3D gaze, and eye movement information. Since the features of each modality are extracted at different time resolutions (i.e., audio features every 3 s, facial features at every frame, and additional features every 1.5 s), we apply fixed modality synchronization tailored to each label type, which allows us to perform the cross-channel and multimodal evaluation. Finally, the previously extracted features are combined and further evolved with an MLP to recognize the user's emotional state for the audio- and video-based labels separately.

\subsection{Speech features from audio}
\input{Sections/5_method_audio.tex}

\subsection{Facial expression features from video}
\input{Sections/5_method_faces.tex}

\subsection{Additional features from video}
\input{Sections/5_method_gaze.tex}

\subsection{Temporal synchronization of modalities}
\input{Sections/5_method_sync.tex}

\subsection{Final models}
\input{Sections/5_method_multimodal.tex}

%% file: Sections/5_method_audio.tex
In this work, we only consider those audio segments with an associated emotion while the user is speaking and for which the avatar speaks less than one-third of the segment duration.

First, we use the WavLM speech model~\cite{chen2022wavlm} to extract the acoustic information of each segment using the raw signal waveform. WavLM was trained on 94k hours of English-spoken audios extracted from three large-scale speech datasets, and can obtain high performance in SER, among other tasks. We extract the features of its last hidden states, which outputs a 1024D vector every 1/50 s. This results in 150 feature vector instances per segment. We compute the average for the time dimension to reduce the final feature length to 1024.

Then, we feed such features to four two-layer MLPs: one for categorical emotional state recognition and three for the dimensional model. Although the main goal is to perform categorical recognition, we decided to include the dimensional model to leverage the available annotations and enrich the feature representation, given the relationship between the two models~\cite{russell1980circumplex}. We also chose to train them separately since each task may converge at a different rate; thus, a multi-task learning approach would not be optimal for all the outputs. 
The first layer reduces the 1024 extracted features to 64 with ReLU as the activation function, while the second one is in charge of extracting the logits for the prediction of the emotional states via softmax. Cross-entropy is used as a loss function, and the Adam optimizer is used to train the four networks with a learning rate of 0.001 over 5K iterations. To deal with the imbalance of the data, the sampling probability for the samples of the minority classes is four times higher than \textit{neutral}.

Finally, the logits of the four models are concatenated to the computed WavLM features in a hybrid fusion fashion, resulting in a 1031D feature vector. This way, we preserve the generic speech representation and augment it with a reduced set of domain-specific information. We refer to this feature set, and consequently to this modality, as A.

%% file: Sections/5_method_faces.tex
\label{ssec:face_method}

For this work, we adopt a static-based approach, despite dynamic ones usually being better suited for this task~\cite{zeng2007survey}. In our context, issues such as frame loss during data transfer or limitations in network capacity leading to low frame rates may impede the effective utilization of fine-grained dynamics. 

For each video frame, we first detect faces using FaceBoxes~\cite{zhang2017faceboxes} and estimate 68 facial landmarks in the image space by means of 3DDFA\_v2~\cite{guo2020towards}. Less than 1\% of the data is lost on these steps. Using these landmarks, the face is rotated, scaled, and cropped to obtain a normalized RGB image of 224x224 pixels.
Then, we use the Xception CNN model~\cite{Chollet2016} pretrained on ImageNet~\cite{Deng2009} to extract discriminative features from the face images,
and add four fully connected dense layers to the top of the network, each followed by ReLU and dropout (with a rate of 0.5), in addition to a final softmax layer for FER. 
During optimization, we found that the best strategy was to freeze the first 70 layers of Xception and fine tuning the last 10. Consequently, we finetune such layers and train the added ones from scratch on both spoken and silent instances. According to this transfer learning scheme, we get a total of 23.6M parameters, where 16.5M are trainable, and the remaining 7M are non-trainable. Training is based on the Adam optimizer, with a learning rate of 0.001. To tackle the class imbalance issue, we use a weighted cross-entropy loss function where the weight of each emotion class is associated with the inverse frequency in the training set. 

Lastly, we extract the output 256D features from the last hidden layer. We refer to this feature set and modality as F.

%% file: Sections/5_method_gaze.tex
\begin{table}[t!]
\centering
\caption{Functionals computed for each element of the additional modalities (3D gaze vector, eye rotation, and head pose). $x_{\cdot}$ and $y_{\cdot}$ correspond to horizontal and vertical 3D gaze/eye rotation components, respectively. $y$, $p$, and $r$ correspond to head yaw, pitch, and roll, respectively.}
\label{tab:gaze_feats}
\begin{tabular}{@{}cc@{}}
\toprule
\textbf{Element} & \textbf{Functionals} \\ \midrule
\begin{tabular}[c]{@{}c@{}}
$x_{\cdot}$, $y_{\cdot}$, \\ 
abs $\Delta x_{\cdot}$, abs $\Delta y_{\cdot}$,\\ 
abs $\Delta x_{\cdot}/t$, abs $\Delta y_{\cdot}/t$,\\
abs $\Delta \mathbf{g}/t$, abs $\Delta  \mathbf{e}/t$,\\
$y$, $p$, $r$, \\
abs $\Delta y$, abs $\Delta p$, abs $\Delta r$\\ 
\end{tabular} & \begin{tabular}[c]{@{}c@{}}min, max, mean, SD, \\ range (except for abs $\Delta x_{\cdot}$ and abs $\Delta y_{\cdot}$),\\ 25th perc, 50th perc, 75th perc, IQR\end{tabular} \\ \midrule
\begin{tabular}[c]{@{}c@{}}$\Delta x_{\cdot}$, $\Delta y_{\cdot}$, $\Delta  \mathbf{g}$,  $\Delta  \mathbf{e}$,\\
$\Delta y$, $\Delta p$, $\Delta r$, \\
abs $\Delta y/t$, abs $\Delta p/t$, abs $\Delta r/t$\\ 
\end{tabular}& \begin{tabular}[c]{@{}c@{}}mean, SD\end{tabular} \\ \bottomrule
\multicolumn{2}{l}{\footnotesize SD: Standard deviation; perc: Percentile; IQR: interquantile range.}
\end{tabular}%
\end{table}

We also use the video stream to compute a series of additional features based on the per-frame estimated 3D direction vector (represented as horizontal and vertical gaze angles) and head pose (yaw, pitch, and roll). 

To do so, we first fit a 3D face morphable model~\cite{huber2016multiresolution} to the detected 2D landmarks from Sec.~\ref{ssec:face_method} and apply perspective-n-point~\cite{Lepetit:160138} to estimate the 3D head position and orientation, as well as the 3D eye position as the gaze origin. A normalized face image is fed to the 3D gaze estimation model ETH-XGAZE, trained on a dataset of homonymous name~\cite{zhang2020eth}. Although none of the existing gaze estimation datasets include older adults, a qualitative examination of the estimated gaze direction using this model showed that performance was mostly impacted for users wearing colored lens or eyeglasses with substantial reflection caused by the computer screen. We did not detect blinks or pupil size due to their low reliability in our scenario.
To reduce noise, the estimated head pose and gaze trajectories are postprocessed with a 5-frame median filter. Combining head pose $\mathbf{h} = (y, p, r)$ and gaze $\mathbf{g} = (x_g, y_g)$ vectors, we further convert the 3D gaze direction into an eye-in-head gaze direction vector, that is, mimicking eye rotation $\mathbf{e} = (x_e, y_e)$. The three data sources are filtered to discard invalid data (e.g., frames with incorrectly detected faces, or for which head or eye movements are not anatomically plausible). Following other works on eye and head pose processing~\cite{bulling2010eye,samanta2020emotion}, all trajectories are processed using a sliding window of 1.5 s and stride of 1, and centered at every half second throughout the video. 
Windows smaller than 0.5 s or for which more than 50\% of the frames are invalid are discarded, accounting for around 2\% of the windows. 
We compute a 227D feature vector for each window, containing information from the three data sources represented as functionals of the trajectories (see Table~\ref{tab:gaze_feats}), and a complementary attention measure, described below. Due to the effect of glasses on the resulting eye trajectories, we add a manually annotated ternary flag to denote whether the participant is wearing glasses, and if so, whether the eyes are clearly visible.
We refer to the resulting 228D feature set and modality as G.

\subsubsection{Looking at VC} The EMPATHIC-VC system can use this measure to estimate whether the user is engaged with the VC. Here, we largely follow~\cite{amorese2022using} to estimate the location of the VC, wherein it is assumed that the zone (cluster) in the 3D space with the highest density of gaze vectors intersecting the camera plane is where the VC is located. We create a 6D one-hot encoding vector denoting the looking-at-VC likelihood from lower to higher, based on the distance from the gaze point to the cluster. Per-valid-frame vectors are averaged over a time window, producing a 6D vector per window. 

\subsubsection{3D gaze direction}
We compute $\mathbf{g}$ functionals of: per-component (i.e., $x_g, y_g$) gaze angles, per-component angle differences (e.g., $\Delta x_g$) and their magnitude (e.g., abs $\Delta x_g$) between any two consecutive frames, direction ($\Delta \mathbf{g}$) and speed ($\Delta \mathbf{g}/t$) of the gaze vector between any two consecutive frames, and per-component speed (e.g., $\Delta x_g/t$) between any two consecutive frames. This results in a 67D feature vector.

\subsubsection{Eye rotation}
We compute the same functionals for $\mathbf{e}$ as for $\mathbf{g}$, resulting in a 67D feature vector. 

\subsubsection{Head rotation} 
Lastly, we compute $\mathbf{h}$ functionals for the following: per-component (i.e., $y, p, r$) head pose angle, per-component angle differences (e.g., $\Delta y$) and their magnitude (e.g., abs $\Delta y$) between any two consecutive frames, and per-component speed (e.g., $\Delta y/t$) with respect to any two consecutive frames. This results in a 87D feature vector. 

%% file: Sections/5_method_sync.tex
In order to effectively integrate and analyze the multimodal data captured from different sources, we employ a fixed modality synchronization approach per label type. 

For the audio-based evaluation, for which the system would output an estimate every 3 s, we compute the average and SD of the available per-frame F features within an audio segment, resulting in a 512D vector. This provides a robust facial expression descriptor that is less susceptible to accidental fluctuations despite disregarding facial temporal dynamics. Preliminary experiments evaluated a second version, consisting in concatenating the features of the most central frame of each second of the audio segment, hence maintaining such dynamics. However, the former version outperformed the latter for the majority of settings. Regarding G, we use the window aligned to the center of the audio segment, thus discarding those windows at the extremes of the segment. 

Conversely, for video-based evaluation, the temporal resolution is increased to frame level. Thus, each G window and A segment are used multiple times and matched to different frames. In particular, we associate each frame with a specific G window and A segment based on its closest proximity to the central timestamp of the respective window and segment.

This way, all F frames, G windows, and A segments receive audio and video labels. 
For each evaluation case, feature sets that do not have correspondence due to missing data of any of the modalities are omitted, resulting in around 86\% of the original data for audio and 98\% for video.

%% file: Sections/5_method_multimodal.tex
The extracted features from a given modality are normalized according to the training set range and fed to a 2-layer MLP with ReLU activation and dropout of 0.5, followed by a softmax dense layer for classification of a given label type. We evaluate three low-complexity MLP configurations, with number of hidden layers 100-20, 200-40, and 500-100. For multimodal evaluation, the feature sets of the different modalities are concatenated before being fed to the MLP. We evaluated other attention-based fusion approaches in preliminary experiments, such as self- and cross-modal attention~\cite{rajan2022cross}. However, their performance was equivalent to concatenation, so we proceed with the latter for the experimental evaluation.

We tackle data imbalance by randomly sampling instances of each class with the same probability. Additionally, due to the small sample size of the audio-based evaluation, we employ an oversampling strategy such that each sample of the minority class (\textit{pleased} for SP and FR, and \textit{puzzled} for NO and WH) is utilized around three times per epoch. To maintain an approximate balance between classes, the other classes are sampled a similar number of times. The training samples per epoch are thus set to 5418 samples for SP, 2556 for FR, 234 for NO, and 10494 for WH. Conversely, since the sample size for video-based evaluation is considerably higher but also contains higher redundancy, we set the training sample size to 7500 for all countries. Samples are randomly selected; thus, at the end of the training stage, all samples from the minority classes are seen multiple times, while for \textit{neutral}, only a fraction is seen.

All evaluated models are trained with cross-entropy loss, Adam optimizer, learning rate of 0.0001, and batch size of 64. We empirically set the number of training epochs to 100 for all countries and evaluations except for NO with audio-based labels, for which we train for 200 epochs.

%% file: Sections/6_automatic_experiments.tex
Here, we present a comprehensive experimental evaluation to assess the impact of different modalities on the recognition performance of emotional states for audio and video labels. 

\subsection{Research questions}
\label{ssec:questions}

The characteristics of the EMPATHIC WoZ scenario allow us to evaluate the contribution of the different modalities for the considered emotions in various contexts. First, we separately consider the evaluation scenario with audio-based labels and the one with video-based labels. We have a main modality for each label type: A for audio-based and F for video-based labels. We refer to the remaining modalities (e.g., F and G for audio-based evaluation) as auxiliaries for that evaluation scenario. Main and auxiliary modalities can be combined to improve performance. Each evaluation is performed in each country individually (SP, FR, and NO) and on WH. The latter allows us to evaluate trends of the complete set of data and quantify the effect of training with country-specific data in comparison to a larger multi-country set. The audio-based scenario only includes data where the user is speaking. By contrast, for the video-based scenario, we can compare the performance of evaluating spoken content to silent content. What is more, as for the country-oriented evaluation, we can assess the effect of training the final video-based model with speaking-status-specific data in comparison to with all data.

On this basis, we aim to answer the following research questions: \textit{RQ}$_{1}$) Can the main modality for a given label type obtain the same discriminative power for all the classes considered?; \textit{RQ}$_{2}$) Can the auxiliary modalities achieve similar performance to the main modality?; \textit{RQ}$_{3}$) Is multimodality beneficial?; \textit{RQ}$_{4}$) Are there noteworthy differences in performance among countries?; \textit{RQ}$_{5}$) Does training with data from multiple countries prove beneficial with respect to country-specific training?; \textit{RQ}$_{6}$) For video-based evaluation, does training with spoken and silent instances prove beneficial with respect to spoken/silent-specific training?; \textit{RQ}$_{7}$) Are there any performance differences between spoken and silence instances?; and \textit{RQ}$_{8}$) Are there any performance differences between audio- and video-based evaluation?

\subsection{Evaluation protocol}
We build 10-fold subject-independent training and test splits for each country subset (SP, FR, NO) and a fourth one with the data of all the countries (WH) following approximately a 9:1 ratio. The folds for WH contain the same subjects as the per-country folds. Architecture selection (i.e., the number of MLP hidden units) and hyperparameter tuning are carried out independently per experiment based on random validation subpartitions of the training splits. For each experiment, the best configuration over all folds is then retrained on the whole per-fold training split and used for all folds. Best architectures per experiment are reported in Sec.~A-C of the supplementary material. We perform 10-fold cross-validation three times following the same splits for all models to account for the stochasticity of the data sampling and whole learning process.

Performance is measured per fold by means of the unweighted average accuracy, also known as unweighted average recall, which gives the same weight to the accuracy of each class regardless of the number of samples for each class. Per-class accuracy is thus equivalent to per-class recall (i.e., the number of samples predicted correctly out of the total number of samples for a given class). Note that the test splits of some folds do not contain all classes, especially \textit{puzzled} for NO. In such cases, the average accuracy is computed only for the classes that have at least one sample in the test split. We also perform multiple pairwise comparisons with the corrected repeated k-fold cross-validation t-test~\cite{bouckaert2004evaluating} to test for statistically significant differences (p$<$.05) among average accuracy results. We control for the false discovery rate using the BKY correction~\cite{benjamini2006adaptive}, grouped by country subset. 

Due to the large amount of experiments, in the following subsections we report the results for the WH dataset, trends observed across different countries, and noteworthy highlights specific to each country. For additional results on a per-country basis, please refer to Sec.~B,~C and~D of the supplementary material for audio-based, video-based under speech, and video-based under silence results, respectively. Country sets used for training and testing are denoted as \textit{training country}\textrightarrow \textit{testing country} (e.g., WH\textrightarrow SP to denote training with WH and testing with SP). The WH models selected for the country-specific comparison are the ones that worked better for the WH validation sets, so the reported performance would be different if selecting the best models for each country independently. Likewise, models trained on WH with silence and speech data are also the ones that worked better for the WH validation sets.

\input{Sections/6_3_fusion_results.tex}

%% file: Sections/6_3_fusion_results.tex
\subsection{Audio-based emotion expression recognition results}
\label{ssec:audio_results}

\begingroup
\setlength{\tabcolsep}{5pt} 
\begin{table}[t!]
\centering

\caption{\textbf{Audio-based results on WHOLE}, reported as unweighted avg. accuracy $\pm$ SEM over 10 folds and 3 runs per fold. Bold: best acc. per group. Underlined: best acc. overall.}
\begin{tabular}{@{}lcccc@{}}
\toprule
\multicolumn{1}{c}{\textbf{Modality}} & \multicolumn{1}{c}{\textbf{\begin{tabular}[c]{@{}c@{}}Calm\\ Accuracy\end{tabular}}} & \multicolumn{1}{c}{\textbf{\begin{tabular}[c]{@{}c@{}}Pleased\\ Accuracy\end{tabular}}} & \multicolumn{1}{c}{\textbf{\begin{tabular}[c]{@{}c@{}}Puzzled\\ Accuracy\end{tabular}}} & \multicolumn{1}{c}{\textbf{\begin{tabular}[c]{@{}c@{}}Average\\ Accuracy\end{tabular}}} \\ \midrule
A & 76.42 $\pm$ 0.8 & 63.54 $\pm$ 1.3 & 59.89 $\pm$ 2.3 & 66.61 $\pm$ 0.6 \\ 
\midrule 
F &  15.9 $\pm$ 1.3 & \underline{\textbf{69.91 $\pm$ 3.2}} & \underline{\textbf{63.41 $\pm$ 3.0}} & \textbf{49.74 $\pm$ 0.9} \\
G & \textbf{25.26 $\pm$ 1.6} & 49.74 $\pm$ 2.3 & 44.66 $\pm$ 2.0 & 39.89 $\pm$ 0.9 \\  \midrule
A+F & 75.68 $\pm$ 0.5 & \textbf{67.59 $\pm$ 1.6} & \textbf{61.65 $\pm$ 2.2} &  \underline{\textbf{68.31 $\pm$ 0.5}} \\
A+G & 76.62 $\pm$ 0.7 & 62.38 $\pm$ 1.4 & 58.52 $\pm$ 2.3 & 65.84 $\pm$ 0.6 \\
 A+F+G & \underline{\textbf{76.93 $\pm$ 0.7}} &  67.11 $\pm$ 1.7 & 60.41 $\pm$ 2.7 & 68.15 $\pm$ 0.6 \\  
\bottomrule
\end{tabular}%
\label{tab:audio_whole_results}
\end{table}
\endgroup

Table~\ref{tab:audio_whole_results} summarizes the results for the different experiments with audio-based labels on the WH dataset. We report below the results with respect to each research question.

\subsubsection{Main modality} A alone obtains higher accuracy with \textit{calm}, followed by \textit{pleased} and then \textit{puzzled}, correlated with the number of samples per class.
Furthermore, \textit{puzzled} gets more confused with \textit{calm} than \textit{pleased}. Country-wise, we see similar trends, except that, for SP, \textit{puzzled} obtains higher accuracy than \textit{pleased}. For FR, the accuracy for \textit{pleased} is less stable than for \textit{puzzled}. 
In general, the results are more stable across runs than across folds. The mean SD across folds for WH is 3.3\%, while the mean SD across runs is around 0.8\%. 
Thus, as a general note, changes in the standard error of the mean (SEM) mainly denote higher variability across folds.

\subsubsection{Auxiliary modalities} 
Despite the high accuracy obtained by F for \textit{pleased} and \textit{puzzled}, confusion patterns reveal that \textit{calm} is mostly confused with \textit{puzzled}, which does not occur for \textit{pleased}. This implies that F provides information that is particularly discriminative and potentially correlated to the audio modality specifically for the latter class. 
By further analyzing the F predictions and their correlation to facial expression categories, we observe that 97\% of the \textit{pleased} predictions correspond to audio segments where the majority facial expression is \textit{happy}, despite only 30\% of the audio segments matching to a \textit{happy} expression having the \textit{pleased} annotation. Thus, the same features that correspond to the \textit{happy} facial expression are related to the facial features corresponding to a \textit{pleased} speech. On the contrary, G's results are slightly over random performance, indicating that G alone is not informative enough to recognize the classes considered. 
All unimodal (A, F, G) and unimodal vs bimodal (F vs A+F, G vs A+G) pairwise comparisons are significantly different (p$<$.0001). Trends are overall maintained country-wise, except that, for SP, \textit{calm} benefits more from F and \textit{puzzled} from G. Nevertheless, G and F results are generally less stable than A.

\subsubsection{Multimodality}
\label{sssec:multimodal_whole}
 Overall, incorporating F improves performance over A alone. By contrast, incorporating G seems detrimental on average. The best multimodal approach for WH, A+F, achieves a 2.5\% relative performance increase over A. Class-wise, we observe that adding G increases performance and stability for \textit{calm}, while adding F is beneficial for \textit{pleased} and \textit{puzzled}, despite the stability of the latter decreases. Statistical tests further confirm that A+F vs A+G (p=.038) and A+G vs A+F+G (p=.024) differ significantly. SP and NO follow similar trends class-wise, although for them, G is also beneficial for \textit{puzzled} but to a lesser extent than A. For FR we observe an inverse trend, which we comment below.

\begin{figure}[t!]
\centering
 \includegraphics[width=0.45\textwidth]{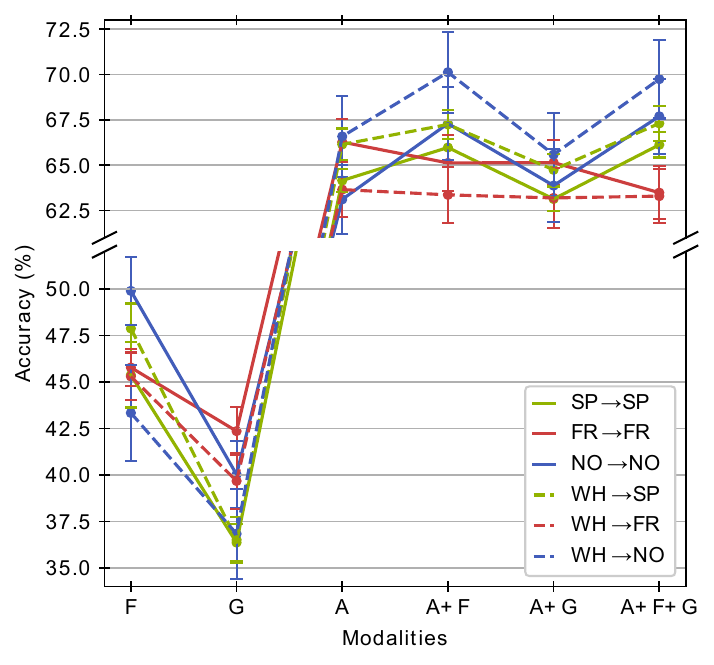}
\caption{\textbf{Per-country audio-based results,} training on either SP, FR, NO, or WH training sets, and evaluating on SP, FR and NO test sets. Reported as unweighted avg. accuracy $\pm$ SEM over 10 folds and 3 runs per fold. }
\label{fig:plot_audio_countries}
\end{figure}

\subsubsection{Comparison across countries}
Fig.~\ref{fig:plot_audio_countries} depicts per-country average accuracy results. With respect to multimodality, for SP and NO, we observe a similar trend to that of Table~\ref{tab:audio_whole_results}, with A+F and A+F+G outperforming A alone. For FR, however, accuracy decreases as the number of features increases, obtaining the highest average accuracy overall with A (66.28\%), which evidences variations among countries. First, FR obtained the lowest inter-agreement score (Sec.~\ref{sec:audio_annotations}). Analyzing the FR dataset distribution, we find that the average SD of the FR audio features is slightly larger than that of the other countries (0.68 for FR, 0.65 for SP, and 0.63 for NO), indicating that the data is more dispersed in the feature space. Furthermore, the best models for FR are the ones with the lowest number of parameters, as opposed to NO, which requires the largest number of parameters despite having a similar sample size (Sec. B.C of the supplementary). These results indicate that adding more features to FR increases the risk of overfitting, thus decreasing performance.
Class-wise, in contrast to WH results, both G and F aid in \textit{puzzled} recognition for SP, while both aid in \textit{calm} recognition for FR. Additionally, FR obtains the highest accuracy for \textit{pleased} with A alone, while for the rest of the classes and countries, multimodal models outperform A. With respect to the auxiliary modalities, NO and FR appear to leverage F and G better than SP, respectively. Per-class accuracies are redistributed with respect to WH.

\subsubsection{Expanding training data including other countries}
Fig.~\ref{fig:plot_audio_countries} also depicts the effect of training with the WH dataset instead of each country separately. As can be seen, adding more data from different countries consistently improves accuracy on average for SP and NO with A-based models, although stability decreases. NO trained on WH obtains the highest accuracy overall (70.13\% with A+F). For FR, however, we find the opposite effect again, with a decrease in accuracy of up to 4.1\% for A, where \textit{pleased} is the most affected class. By contrast, the behavior of the auxiliary modalities is the opposite, with F obtaining the highest accuracy for this class, which additionally increases with respect to FR\textrightarrow FR. Considering previous findings, we hypothesize that the FR audio feature distribution of the \textit{pleased} class is significantly different than that of SP and NO; thus, adding more data is detrimental. Another difference comes from the arousal distribution of FR, being the country with the highest number of annotated \textit{excited} instances (61.75\% compared to 12-18\% for the other countries). Continuing with class-wise results, SP obtains performance increases for all classes in a similar proportion, benefiting from the increased data variability and sample size. By contrast, \textit{calm} performance decreases for NO, which may be caused by the significant increase in the number of training instances for the minority classes (around 279\% and 4369\% increase, respectively). With respect to auxiliary modalities, the general trend shows that adding more data hurts performance. By further analyzing confusion patterns, we observe that these are mostly reversed when training with WH, especially affecting discrimination between \textit{calm} and the other classes for G and \textit{calm} and \textit{puzzled} for F. For the latter modality, though, \textit{pleased} is still recognized accurately.

\subsection{Video-based emotion expression recognition under speech}

\begingroup
\setlength{\tabcolsep}{5pt} 
\begin{table}[t!]
\centering
\caption{\textbf{Video-based results under speech on WHOLE}, reported as unweighted avg. accuracy $\pm$ SEM over 10 folds and 3 runs per fold. Bold: best accuracy per group. Underlined: best accuracy per training type. Italics: best accuracy overall.}
\begin{tabular}{@{}lcccc@{}}
\toprule
\multicolumn{1}{c}{\textbf{Modality}} & \multicolumn{1}{c}{\textbf{\begin{tabular}[c]{@{}c@{}}Neutral\\ Accuracy\end{tabular}}} & \multicolumn{1}{c}{\textbf{\begin{tabular}[c]{@{}c@{}}Happy\\ Accuracy\end{tabular}}} & \multicolumn{1}{c}{\textbf{\begin{tabular}[c]{@{}c@{}}Pensive\\ Accuracy\end{tabular}}} & \multicolumn{1}{c}{\textbf{\begin{tabular}[c]{@{}c@{}}Average\\ Accuracy\end{tabular}}} \\ \midrule 
\multicolumn{5}{l}{\textit{Training on speech data:}}  \\ \midrule
F & 73.98 $\pm$ 1.0 &  72.2 $\pm$ 2.4 & 57.87 $\pm$ 2.1 & 68.02 $\pm$ 1.1  \\ \midrule 
A &  39.07 $\pm$ 1.1 & \textbf{63.57 $\pm$ 2.1} & 66.67 $\pm$ 1.5 & 56.44 $\pm$ 0.6 \\
G &  \textbf{51.13} $\pm$ 1.8 & 46.33 $\pm$ 2.4 & \underline{\textbf{74.47 $\pm$ 2.2}} & \textbf{57.31 $\pm$ 0.5} \\  \midrule 
F+A & 74.5 $\pm$ 0.9 & \underline{\textbf{73.39 $\pm$ 2.4}} & 60.95 $\pm$ 1.9 & 69.61 $\pm$ 1.1 \\
F+G & \textit{\underline{\textbf{76.45 $\pm$ 1.1}}} & 71.58 $\pm$ 2.4 & 66.86 $\pm$ 2.2 & 71.63 $\pm$ 1.1 \\
F+A+G & 76.36 $\pm$ 1.0 & 73.27 $\pm$ 2.4 & \textbf{68.52 $\pm$ 1.9} & \textit{\underline{\textbf{72.72 $\pm$ 1.0}}} \\ 
\midrule
\multicolumn{5}{l}{\textit{Training on all data (speech + silence):}}  \\ \midrule
F & 70.44 $\pm$ 1.1 & \textit{\underline{73.44 $\pm$ 2.2}} & 61.58 $\pm$ 2.1 & 68.49 $\pm$ 1.1 \\ \midrule
G & 42.45 $\pm$ 2.0 & 55.56 $\pm$ 2.1 & \textit{\underline{76.96 $\pm$ 2.0}} & 58.32 $\pm$ 0.5  \\  \midrule 
F+G & \underline{73.55 $\pm$ 1.1} & 73.11 $\pm$ 2.2 & 70.56 $\pm$ 2.1 & \underline{72.41 $\pm$ 1.0} \\ 
\bottomrule

\end{tabular}%
\label{tab:video_speech_results}
\end{table}
\endgroup

Table~\ref{tab:video_speech_results} summarizes results for video-based labels under speech trained and evaluated on WH, using two different training regimes: 1) training on samples where the user is speaking (\textit{speech data}); and 2) training on all samples irrespective of speaking status (\textit{speech+silence}). We report the results below.

\subsubsection{Main modality}
F alone obtains similar accuracy for \textit{neutral} and \textit{happy}, while comparatively struggles with \textit{pensive}, which is highly confused with \textit{neutral}. This behavior is not proportional to the number of instances since \textit{pensive} has more than \textit{happy}. Nonetheless, \textit{happy} performance is slightly less stable than that of \textit{pensive}. Country-wise, however, the performance gap is found between \textit{neutral} and the minority classes instead. The SD across folds is 6.2\%, higher for this scenario than for the audio-based but more consistent. By contrast, the SD across runs is around 0.05\%.
This unveils the large variability across subjects.

\subsubsection{Auxiliary modalities}  A and G obtain accuracy results closer to the main modality than for the audio-based scenario, with G slightly outperforming F on average. Statistical tests show significant differences for F vs A and F vs G (p=.015), and F vs F+G/A (p$<$.001) when training with speech data, and for all comparisons when training with all data (p$<$.001). Remarkably, G achieves the highest accuracy overall for \textit{pensive}. A appears to be informative for \textit{pensive} as well but to a lesser extent, also outperforming F, and is more informative than G for \textit{happy}. Nonetheless, G is less stable than A class-wise, although on average, they are more stable than F.
Trends are maintained across countries class-wise. However, on average, A is more informative than G for them, mostly caused by the extremely low performance of G for \textit{happy}, which gets confused with \textit{neutral}. Analyzing the confusion patterns for all datasets, we confirm that gaze cues are highly discriminative for \textit{pensive} and audio cues are highly discriminative for \textit{happy}. 

\subsubsection{Multimodality} When training on speech data, we observe that adding A or G to F increases accuracy, and the highest is achieved by combining the three modalities, showing a 6.9\% relative improvement over F alone. Class-wise, adding G substantially improves performance for \textit{pensive} followed by \textit{neutral}, while adding F has a more subtle effect. By contrast, \textit{happy} appears to benefit from A and not G, but does so when combining the three modalities. 
We observe similar trends when training on all data. Statistical tests confirm significant differences for the following cases when training on speech data: F vs F+G (p=.031), F vs F+A+G (p=.018), F+G vs F+A (p=.044), and F+A vs F+A+G (p=.015); and when training on all data: F vs F+G (p=.015). Trends are overall maintained across countries, with some differences highlighted below.

\begin{figure*}[t!]
\centering
\begin{subfigure}{0.465\textwidth}
 \includegraphics[width=\textwidth]{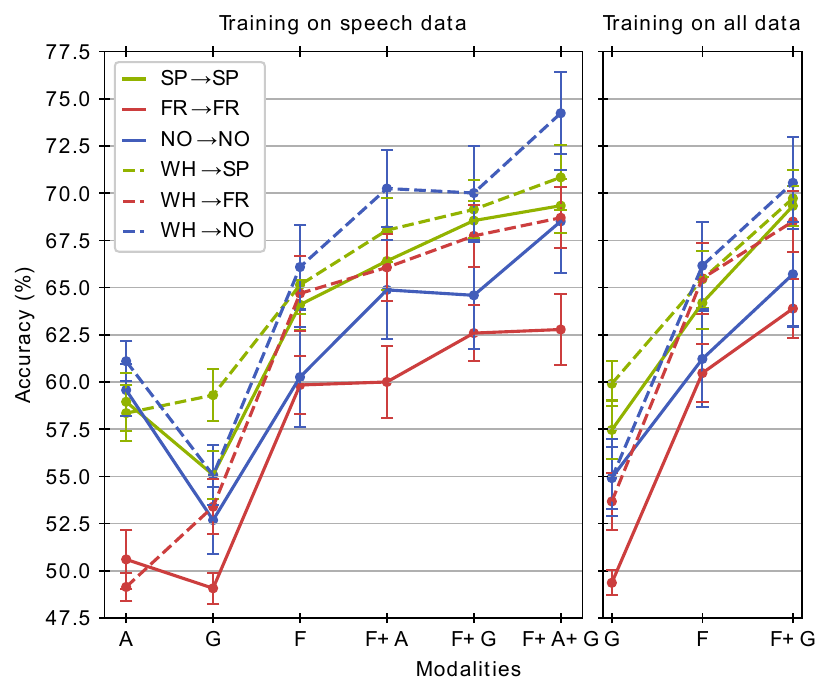}
\vspace{-0.6cm}
\subcaption{Testing on speech data}
\label{fig:plot_video_speech_countries}
\end{subfigure}
\begin{subfigure}{0.465\textwidth}
 \includegraphics[width=\textwidth]{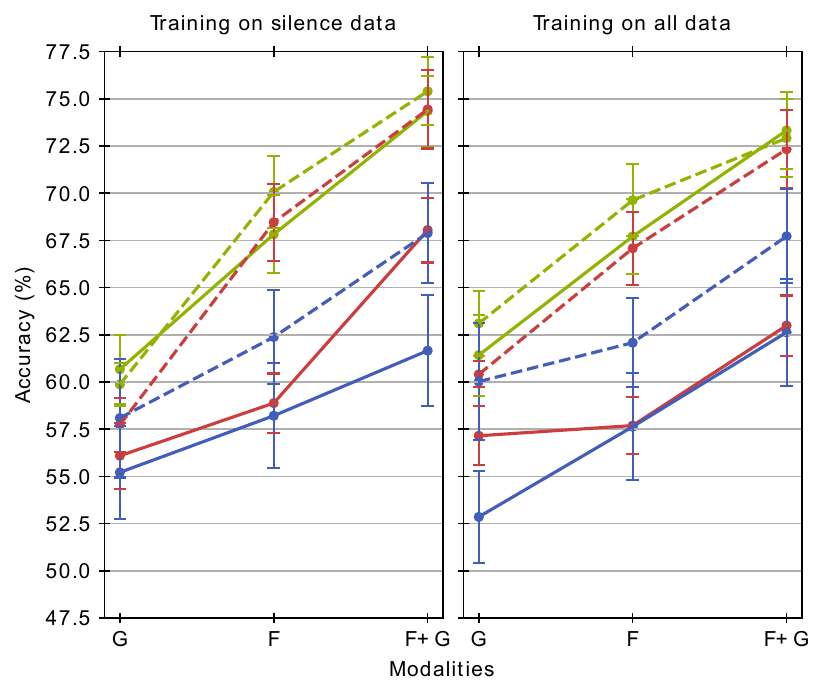}
\vspace{-0.6cm}
\subcaption{Testing on silence data}
\label{fig:plot_video_silence_countries}
\end{subfigure}
\caption{\textbf{Per-country video-based results under (a) speech or (b) silence}, trained on SP, FR, NO, and WH training sets, and evaluated on SP, FR and NO test sets. Reported as unweighted average accuracy $\pm$ SEM over 10 folds and 3 runs per fold. }
\label{fig:plot_video_countries}
\end{figure*}

\subsubsection{Comparison across countries}  Fig.~\ref{fig:plot_video_speech_countries} illustrates per-country average results for all modalities and the two training regimes. In general, SP achieves the highest accuracy for all modality combinations, greatly benefiting from G followed by A when added to F. It obtains the highest accuracy overall with F+A+G (69.34\%) when training with speech, and with F+G (69.33\%) when training with all data. NO achieves a similar accuracy with the trimodal model, although it benefits more from A than from G. By contrast, FR barely benefits from adding A, and repeatedly scores the lowest, despite having slightly more data than NO and almost equal performance on average with F. This might be partly caused by the difference in class proportions across countries, with \textit{happy} being the most variant class (4.7\% of the total data for FR, 0.8\% for SP, and 2.7\% for NO). 
The difference in the distribution of audio features for FR (discussed in Sec.~\ref{ssec:audio_results}) is also noticeable here, with A alone obtaining the lowest accuracy for FR, and with a substantial difference compared to the other countries. Adding A to F hurts \textit{pensive} recognition for FR due to a high confusion between \textit{neutral} and \textit{pensive}, although their performance alone is better, and the highest with A. With respect to G, we observe the highest discriminative power for \textit{pensive} with SP, with more elevated levels of \textit{neutral}-\textit{happy} confusion for the other two countries, thus scoring lower in the comparison.

\subsubsection{Expanding training data including other countries} Fig.~\ref{fig:plot_video_speech_countries} also illustrates the effect of training with WH instead of each country separately for the two training regimes. In general, adding more data increases accuracy on average for all cases except for A, for which accuracy is mostly maintained or slightly reduced. As can be seen, FR obtains the highest performance increase overall despite still scoring the lowest, and NO obtains the highest accuracy results overall, with F+A+G being the top performer (74.24\%). SP and FR also achieve the highest accuracy with F+A+G. Models are slightly less stable when training on WH for F-based models, except for NO. SP obtains the lowest gain, highly likely due to being the country with the most instances in WH for all classes except for \textit{happy}. However, when we investigate class-wise trends, we observe an interesting difference. For NO and FR, the minority classes have their accuracy significantly increased, and the \textit{neutral} accuracy decreased for all modality combinations, proving that the increase in variability and effective training data is beneficial for them to decrease confusion with \textit{neutral}. By contrast, SP sees the \textit{neutral} accuracy increase with all F-based models for all cases and with G only when training with speech data, while F and F+G maintain accuracy for \textit{pensive} and decrease it for \textit{happy}. However, performance increases when adding A for the minority classes, although A alone gets \textit{happy} accuracy reduced, while with G alone they both improve. For silence-based evaluation (Sec.~\ref{ssec:face_silence_results}), \textit{happy} accuracy does increase with F-based models when training on WH. Thus, we believe this difference in \textit{happy} performance is due to the facial deformations caused by speaking being different across countries, greatly increasing variability. Although this is true for all countries, due to the small number of \textit{happy} instances for SP, this distribution might be narrower than that of FR and NO. Therefore, increasing variability could be detrimental. Then, when adding A to F, A helps discriminate better among classes that visually might be more similar. Regarding A alone, we mostly observe a redistribution of accuracies among classes, with an increase in \textit{happy}-\textit{neutral} confusion.

\subsubsection{Expanding training data including silence instances} As can be seen in Table~\ref{tab:video_speech_results} for WH, training on all data marginally but consistently improves performance on average, with the highest improvement obtained with G. Class-wise, confusion patterns reveal that the minority classes are less predicted as \textit{neutral}, with a slight increase in confusion in the other direction in some cases. Nonetheless, the change is positive for \textit{pensive} and \textit{happy}. For G specifically, the \textit{neutral}-\textit{happy} confusion patterns are inverted. The consistent improvement is also observed across countries, as depicted in Fig.~\ref{fig:plot_video_speech_countries}, both when training per country and when training on WH, and the class-wise trends are generally maintained. As a matter of fact, per-class accuracies tend to be more balanced in this setting. This indicates that, by including training instances with no facial deformations caused by speaking, the models can pick up other cues that are consistent regardless of speaking status, which helps detect more actual \textit{happy} and \textit{pensive} instances. Specifically, \textit{pensive} always obtains the highest accuracy overall, with a slight performance increase when training on WH with models including G. The other classes also increase their accuracy when training with all data, although the highest accuracy is obtained from including A, which can only be accomplished when training with speech instances.

\subsection{Video-based emotion expression recognition under silence}
\label{ssec:face_silence_results}
\begingroup
\setlength{\tabcolsep}{5pt} 
\begin{table}[t!]
\centering
\caption{\textbf{Video-based results under silence on WHOLE}, reported as unweighted avg. accuracy $\pm$ SEM over 10 folds and 3 runs per fold.  Underlined: best accuracy per training type. Italics: best accuracy overall.}
\begin{tabular}{@{}lcccc@{}}
\toprule
\multicolumn{1}{c}{\textbf{Modality}} & \multicolumn{1}{c}{\textbf{\begin{tabular}[c]{@{}c@{}}Neutral\\ Accuracy\end{tabular}}} & \multicolumn{1}{c}{\textbf{\begin{tabular}[c]{@{}c@{}}Happy\\ Accuracy\end{tabular}}} & \multicolumn{1}{c}{\textbf{\begin{tabular}[c]{@{}c@{}}Pensive\\ Accuracy\end{tabular}}} & \multicolumn{1}{c}{\textbf{\begin{tabular}[c]{@{}c@{}}Average\\ Accuracy\end{tabular}}} \\ \midrule 
\multicolumn{5}{l}{\textit{Training on silence data:}}  \\ \midrule
F & 73.33 $\pm$ 1.8 & \textit{\underline{74.33 $\pm$ 3.2}} & 57.76 $\pm$ 2.8 & 68.47 $\pm$ 1.5 \\ \midrule
G & 61.3 $\pm$ 1.6 & 48.15 $\pm$ 3.4 & \textit{\underline{74.09 $\pm$ 1.2}} & 61.18 $\pm$ 0.8 \\  \midrule
 
F+G & \underline{79.45 $\pm$ 1.3} & 72.76 $\pm$ 3.2 & 71.74 $\pm$ 1.8 & \textit{\underline{74.65 $\pm$ 1.2}} \\ 

\midrule
\multicolumn{5}{l}{\textit{Training on all data (speech + silence):}}  \\ \midrule
F & 79.56 $\pm$ 1.4 &  \underline{72.3 $\pm$ 3.4} & 51.35 $\pm$ 2.8 & 67.74 $\pm$ 1.5 \\ \midrule
G & 67.34 $\pm$ 1.4 & 43.54 $\pm$ 3.4 &  \underline{72.3 $\pm$ 1.5} & 61.06 $\pm$ 1.1  \\  \midrule
F+G & \textit{\underline{85.23 $\pm$ 1.0}} & 71.35 $\pm$ 3.4 & 61.98 $\pm$ 1.8 & \underline{72.85 $\pm$ 1.1} \\ 
\bottomrule
\end{tabular}%
\label{tab:video_silence_whole_results}
\end{table}
\endgroup

Table~\ref{tab:video_silence_whole_results} summarizes results for video-based labels under silence trained and evaluated on WH, using two different training regimes: 1) training on samples where the user is not speaking (\textit{silence data}); and 2) training on all samples irrespective of speaking status (\textit{silence+speech}).

\subsubsection{Main modality} For WH, results match the video-under-speech scenario on average and per class, with a decrease in stability. The \textit{neutral}-\textit{pensive} confusion also decreases. Country-wise, trends are similar to the video-under-speech scenario but differ from WH. For SP, \textit{pensive} is the top performer (73.3\% accuracy) followed closely by \textit{neutral}, while for FR and NO, the top performer is \textit{neutral} by a great margin (around 78-80\% accuracy). For NO specifically, \textit{pensive} obtains extremely low accuracy (38.3\%). As usual, the latter is mostly misclassified with \textit{neutral}, but also with \textit{happy} to a lesser extent, a behavior we had not observed until now. Compared to the video-under-speech scenario, the proportion of \textit{neutral} instances under silence is higher (around 93-98\% depending on the country), which might explain the accuracy bias. For the other classes, however, accuracy results do not generally match class proportions, with \textit{happy} corresponding only to 1.4\% of the total sample size for WH but scoring very close to \textit{pensive}. The SD across folds is 8.5\%, slightly higher than the speech scenario. The SD across runs has a fairly larger range, with an approximate mean of 0.1\%.

\subsubsection{Auxiliary modality} Following previous trends, G alone achieves the highest accuracy and discriminative power for \textit{pensive}. The average accuracy for G gets the closest to F compared to audio and video-under-speech scenarios. G is more stable than F on average for \textit{neutral} and \textit{pensive}, while for \textit{happy} is not. No significant differences are found between F and G on average. These trends are also found across countries. Interestingly, for NO, \textit{pensive} is misclassified as \textit{happy} at a similar proportion compared to the main modality.

\subsubsection{Multimodality} Adding G to F consistently increases accuracy on average, following previous trends class-wise, i.e., increasing accuracy and stability for \textit{neutral} and \textit{pensive}, while not contributing for \textit{happy}, for which F alone is usually the top performer. As a matter of fact, \textit{pensive} is the most benefited class for WH and across countries, and \textit{neutral} obtains the highest accuracy overall across training regimes. 
No statistically significant differences are found, likely due to the high SEM of F alone, which decreases with multimodality. 

\subsubsection{Comparison across countries} Fig.~\ref{fig:plot_video_silence_countries} depicts per-country average accuracy results. As can be seen, all countries benefit from multimodality, with FR benefiting the most when training on silence data. SP achieves the highest accuracy for all settings by a large margin, obtaining the best accuracy overall with F+G (74.35\%) when training with silence data. Contrary to the video-under-speech evaluation, NO scores the lowest. With respect to the auxiliary modality G, its difference with respect to F is small for FR for both training regimes and for NO only when training when silence, as the performance of G decreases when training with all data. Class-wise, trends vary compared to WH depending on the modality, as already discussed above. Again, one highlight is the low accuracy and stability for \textit{pensive} with either modality for NO compared to the other countries and classes, as well as its confusion with \textit{happy}. In fact, NO is the country with the highest number of \textit{pensive} instances (around 5.4\% of the NO sample size), which might indicate a difference in the annotation procedure. 

\subsubsection{Expanding training data including other countries} Fig.~\ref{fig:plot_video_countries} also shows the effect of training with WH. Similarly to the video-under-speech evaluation scenario, SP barely benefits from adding cross-country data compared to the other countries. Still, the highest accuracy is obtained by SP with F+G (75.4\%) when training with silence. For FR, the highest performance increase is obtained by F while, for NO, G is the most benefited modality. Class-wise, we again observe differences between SP and the other two countries. For NO and FR, F-based models increase performance for \textit{happy} and \textit{pensive} and reduce it for \textit{neutral}, while for SP, \textit{neutral} and \textit{happy} performance increases while \textit{pensive} performance decreases. We conclude that \textit{happy} is easier to recognize for F-based models in all countries when no facial deformations caused by speaking occur. However, for SP, we observe an increase in confusion between \textit{neutral} and \textit{pensive} when training with WH, which might be explained by a difference in user behavior or annotation procedure between SP and the other countries when users do not speak. This, in turn, might explain their difference in \textit{neutral}-\textit{pensive} ratios (61:1 for SP, 33:1 for FR, and 17:1 for NO), compared to their similarity when users speak (around 5:1 for all countries). By contrast, NO shows a different trend compared to the other countries for \textit{pensive} with G when training with WH: for SP and FR, \textit{pensive} accuracy decreases when training with silence and it is maintained when training with speech and silence, while for NO, its accuracy increases for both cases. However, we observe again that this accuracy increase also comes with a slight increase in confusion with \textit{happy}. The performance decrease for \textit{pensive} for FR and SP is opposed to the general increase observed for this class on video-under-speech. 

\subsubsection{Expanding training data including speech instances} As can be seen in Table~\ref{tab:video_silence_whole_results}, this training regime slightly reduces the performance of all models on average when applied on WH except for G, for which accuracy is maintained but stability decreases.  
For F-based models, we observe an increase in confusion between \textit{neutral} and the other two classes since the facial deformations added during training decrease discriminative power, while for G, confusion with \textit{neutral} mostly increases for \textit{happy}. Fig.~\ref{fig:plot_video_silence_countries} allows us to compare cross-country performance.
We observe similar trends as for WH on average and class-wise, except for SP\textrightarrow SP with \textit{happy} and NO\textrightarrow NO.
The difference in overall behavior for NO is unclear. We observe a general performance decrease for \textit{pensive}, which could be caused by the high difference in the number of instances between speech and silence sets (808\% increase for WH when including speech instances, and similarly high per country), causing the models to learn patterns more tailored to \textit{pensive} episodes while speaking. This issue may also be causing part of the performance deterioration for \textit{happy}, and it is the opposite effect observed for the video-under-speech scenario when training on all data. In general, and in contrast to video-under-speech, per-class accuracies tend to be more balanced when trained on silence only for F-based models, but also when trained on WH. For G, it is harder to categorize due to \textit{pensive} being the main discriminated class. Nonetheless, this auxiliary modality tends to perform best on average when training on all data, indicating that it benefits from the added variability from different countries and speaking status, regardless of the evaluation scenario. 

\subsubsection{Effect of speaking status on video-based evaluation}

For WH, the users' emotional expressions can be better recognized when they are not speaking with G and F-based models on average. Indeed, evaluating on silence instances with no facial deformations that add noise should increase the discriminative power for \textit{happy} and \textit{pensive}. However, G should not be directly affected by facial deformations, so this leads us to believe that gaze and head patterns may be correlated to speaking status beyond facial deformations, and that such patterns are more discriminative for silence instances. Country-wise, G also tends to work better when the user is silent. By contrast, trends differ for F-based models depending on the country and the training regime. 
Class-wise, \textit{neutral} always obtains the highest accuracy for silence samples trained on silence and speech per country, along with a higher number of false positives.
\textit{Happy} is easier to recognize during silence instances for SP and FR with F when training on WH with silent instances. For NO, \textit{happy} is easier to recognize during speech instances with models that combine F and A, also when training with WH, highlighting again the importance of A for this country. For WH, both scenarios are equally performant. Finally, \textit{pensive} is easier to discriminate when evaluating on silence instances for FR, SP, and WH, although it is difficult to determine which training regime is best due to fluctuating confusion patterns. By contrast, for NO, this class is easier to categorize when the user is speaking with F+G when training on WH with speech instances. Note that, with respect to sample size, both speech and silence subsets are given in around 1:1 ratio for \textit{neutral} and 2:1 for \textit{happy}, while for \textit{pensive}, ratios range from 14:1 for SP to 4:1 for NO.

%% file: Sections/7_discussion.tex
Here, we summarize the findings for each research question (Sec.~\ref{ssec:questions}), and discuss limitations and potential future work.

\subsubsection{Discriminative power of the main modality} We observe varying levels of confusion between the minority classes and the majority class for both label types, with the highest confusion being observed with \textit{pensive}. We strongly believe that this is mostly caused by the FER model considering only spatial information, since this class is mostly characterized by specific dynamics. Consequently, it is the most benefited when F is combined with G, which does provide salient dynamics.

Nevertheless, the minority classes tend to be well discriminated against each other. Despite employing a balancing strategy, per-class accuracies seem to be associated with their corresponding sample sizes. It is important to note, however, that the majority class in both label types (\textit{neutral} and \textit{calm}) is characterized by a relatively lower intensity or absence of emotional expression and can be encountered in transitional phases between other emotions. This poses a challenge in establishing clear category boundaries. All things considered, average accuracies are around 60-70\%, which is a decent rate for a three-class unimodal classifier given the nature of our scenario. This could be further improved by leveraging temporal dependencies across time steps~\cite{tzirakis2017end, Filali2023}.

\subsubsection{Performance of auxiliary modalities compared to the main modality} This depends on the label type. For the audio-based scenario, G struggles to achieve sufficient accuracy. By contrast, F does a better job, usually recognizing \textit{pleased} with similar or higher accuracy and discrimination power than A alone, particularly due to the similarity between \textit{happy} and \textit{pleased} facial features. Conversely, for the video-based scenario, both A and G achieve accuracies closer to the main modality, especially when evaluating on silence instances. In particular, G alone is able to recognize \textit{pensive} with extremely high accuracy, always better than F alone. In addition, in some cases, A is capable of obtaining higher accuracies than F for \textit{pensive} and \textit{happy}, being especially informative for the latter.

Achieving such recognition rates using only the auxiliary modalities is advantageous in scenarios where network or sensor failures may produce potential asynchronies between audio and video streams or even data loss. Having channel-specific labels already allows for individual processing. Therefore, if one stream is affected or even deactivated to reduce processing times, the system can still maintain functionality by utilizing the auxiliary modalities. 
In addition, the video modality may be intentionally altered or disabled for privacy purposes. In the case of alteration, G could still be extracted; however, in the event of video deactivation, only A would remain useful, from which the important video-based events could still be recognized. Nonetheless, as a prospective direction, it is worth considering crossmodal training techniques~\cite{abdou2022gaze}, which learn from multiple modalities at training time to improve single-modality recognition during inference.

\subsubsection{Multimodality} It is beneficial, provided that the modality that is added to the main modality provides discriminative information for a given class. This is the case when adding F to A for the audio-based scenario for the minority classes, and when adding A or G to F for the video-based scenario for all classes, except \textit{happy} for G. The result also depends on the distribution of features: the number of network parameters usually increases with the number of input features, which may lead to overfitting if not accounted for. 

The increase in accuracy when combining A and F is in line with the large audiovisual emotion recognition literature~\cite{poria2017review} and with the few works addressing our target population~\cite{ma2019elderreact}. It is difficult to contextualize the A/F+G results within the literature, as the conclusions depend on the scenario and features used, and most speech-video-gaze/head works employ the VAD model instead~\cite{o2018affective}. 
As an example of discrete emotion recognition, our results resemble the findings of the aforementioned crossmodal work~\cite{abdou2022gaze}, for which including gaze features also improved performance for video-test, while for audio-test only one of their gaze alternatives outperforms the no-gaze option. Class-wise, they observed a similar gaze spatial distribution between \textit{neutral} and \textit{happy}.

Ideally, however, a multimodal system should effectively disregard irrelevant information from individual modalities, thereby preventing any detrimental impact on performance. In that sense, simple feature concatenation may not be a solution to this problem, as it may fail to capture the interactions between features adequately. By contrast, attention-based methods are known to adaptively balance the contributions of different modalities~\cite{guo2019deep}. Nonetheless, our preliminary experiments found no differences among fusion types. Although it is possible that both approaches do yield similar results in our scenario, 1) the stage-wise training, and 2) the fixed-modality synchronization applied are factors that may have influenced the outcome. For the former, end-to-end training would allow modality features to jointly evolve from early network layers~\cite{tzirakis2017end}. For the latter, a flexible temporal synchronization able to capture long-term cross-modal dependencies would compensate for the unaligned nature of communication~\cite{tsai2019multimodal}.

\subsubsection{Cross-country differences}
There are numerous differences that have been previously discussed and will not be reiterated here, which have exposed disparities in how emotions are expressed and perceived across countries. Such differences in behavior and annotation are also reflected in the divergence of the proportions of per-class sample sizes and distribution in the feature space. Nonetheless, results are subject to the imbalance in the number of samples among countries. This causes NO, the country with the lowest sample size, to obtain lower performances in general.

\subsubsection{Multi-country training}
In general, main modalities and multimodal combinations benefit from this training regime. With respect to auxiliary modalities, however, only G benefits from it for the video-based scenario, indicating that the additional features are not as transferable cross-country as the main modalities. Performance gains mainly come from the increased variability, which especially benefits minority classes. However, the extent of such gains does not correlate with the sample size increase, since such increase is proportionally similar for the audio-based scenario than for the video-based scenario, but gains are higher for the latter. This can be attributed to: 1) the audio-based scenario having fewer samples than video; 2) A features already being more generic since they come from a large-scale model, thus more variability may not affect results; 3) some acoustic features of emotional expressions being more language- or culture-specific and hence less transferable across countries than facial expressions, which is in line with previous studies~\cite{riviello2012cross}. Note that no sampling strategy has been applied to balance the number of data samples across countries, hence having a bias towards SP, the country with the highest number of samples. Nonetheless, as found in our experiments, the effect of multi-country training will always depend on the feature distribution and the cultural similarities across countries.

\subsubsection{Training with spoken and silent instances} Training on all data consistently but marginally improves performance when evaluating on spoken instances, increasing discriminative power for the minority classes. This is due to the fact that the added variability, without the noise produced by speaking, helps the models learn discriminative features that are less influenced by the speaking effect. Interestingly though, the performance obtained with this training regime is on par with that obtained when combining F and A trained on spoken instances only. By contrast, when evaluating on silent instances, training with all data produces the opposite effect, as the variability added makes the recognition task harder. Note that there is no sampling strategy to balance the number of spoken/silent data, hence having a slight bias towards spoken samples, primarily for the minority classes. Nonetheless, these findings indicate that speaking status significantly influences the learning process and should be taken into account when devising similar solutions to ours.

\subsubsection{Speech vs silence performance} 
This is highly country-specific. In general, and related to RQ$_{6}$, both speaking statuses achieve similar accuracy on average, but the minority classes tend to be better discriminated against when the user is not speaking, provided the model was trained on silent instances only (note that the original F features were trained on both instance types, so differences in performance will come exclusively from the final models). This conclusion is intuitive for F-based models since facial deformations affect discriminative power. However, since this difference in performance is also observed for G, this indicates that gaze and head patterns are correlated with speaking status beyond facial deformations, and such patterns are more discriminative for silence instances. Nonetheless, if A is combined with F for spoken instances, both spoken and silent instances can obtain comparable results. 

\subsubsection{Audio vs video performance} Audio-based results are constrained by the limited amount of data. While this effect is balanced for A by the use of a pre-trained large-scale model, which may also aid in generalization ability, auxiliary modalities are indeed impacted. By contrast, the video-based evaluation can leverage sample sizes two orders of magnitude higher, but with higher redundancy among samples. Overall, we observe a higher performance for video-based evaluation than for audio. This can be attributed to the differences in the number of samples. Another reason could be that WavLM was trained on English speech; therefore, even if the extracted features are generic enough due to the large pre-training, they are not specialized to the languages considered in this work and hence important subtleties might have been lost.

%% file: Sections/8_conclusion.tex
In this paper, we presented a first comprehensive study on non-verbal emotion expression recognition in interactions between older adults and a simulated VC within the context of the EMPATHIC project. We also described the rationale for data collection and annotation procedure aimed at developing a computational approach that could leverage cues from audio and video channels separately. By analyzing the influence of different modalities, training approaches, and communication modes, this research aimed to shed light on some of the factors that affect the effectiveness of emotion recognition in this scenario. Our findings demonstrate that facial, speech, head, and gaze cues can contribute to the accurate recognition of the channel-specific emotional expressions considered with varying levels of discriminative power. As the evaluation was conducted in a subject-independent manner, these cues would prove even more valuable for a personalized online setup, in which the model could continuously learn from the user's behavior during the interaction. Furthermore, we determined that multi-country training can generally compensate for limited data of a particular country, thereby enhancing overall performance despite country-specific differences.

The insights gained are expected to contribute to the development of more accurate emotion recognition systems, and pave the way for improved VC experiences and personalized technologies catering to the well-being of this age group.

%% file: main.bbl
\begin{thebibliography}{100}
\providecommand{\url}[1]{#1}
\csname url@samestyle\endcsname
\providecommand{\newblock}{\relax}
\providecommand{\bibinfo}[2]{#2}
\providecommand{\BIBentrySTDinterwordspacing}{\spaceskip=0pt\relax}
\providecommand{\BIBentryALTinterwordstretchfactor}{4}
\providecommand{\BIBentryALTinterwordspacing}{\spaceskip=\fontdimen2\font plus
\BIBentryALTinterwordstretchfactor\fontdimen3\font minus
  \fontdimen4\font\relax}
\providecommand{\BIBforeignlanguage}[2]{{%
\expandafter\ifx\csname l@#1\endcsname\relax
\typeout{** WARNING: IEEEtran.bst: No hyphenation pattern has been}%
\typeout{** loaded for the language `#1'. Using the pattern for}%
\typeout{** the default language instead.}%
\else
\language=\csname l@#1\endcsname
\fi
#2}}
\providecommand{\BIBdecl}{\relax}
\BIBdecl

\bibitem{jaimes2007multimodal}
A.~Jaimes and N.~Sebe, ``Multimodal human--computer interaction: A survey,''
  \emph{Computer vision and image understanding}, vol. 108, no. 1-2, pp.
  116--134, 2007.

\bibitem{mckeown2011semaine}
G.~McKeown, M.~Valstar, R.~Cowie, M.~Pantic, and M.~Schroder, ``The semaine
  database: Annotated multimodal records of emotionally colored conversations
  between a person and a limited agent,'' \emph{IEEE transactions on affective
  computing}, vol.~3, no.~1, pp. 5--17, 2011.

\bibitem{EMPATHIC_ANALYSIS}
R.~Justo, L.~Ben~Letaifa, C.~Palmero, E.~Gonzalez-Fraile, A.~Torp~Johansen,
  A.~V{\'a}zquez, G.~Cordasco, S.~Schl{\"o}gl, B.~Fern{\'a}ndez-Ruanova,
  M.~Silva, S.~Escalera, M.~deVelasco, J.~Tenorio-Laranga, A.~Esposito,
  M.~Korsnes, and M.~I. Torres, ``Analysis of the interaction between elderly
  people and a simulated virtual coach,'' \emph{Journal of Ambient Intelligence
  and Humanized Computing}, vol.~11, pp. 6125--6140, 2020.

\bibitem{Vazquez2023}
A.~Vázquez, A.~López~Zorrilla, J.~M. Olaso, and M.~I. Torres, ``Dialogue
  management and language generation for a robust conversational virtual coach:
  Validation and user study,'' \emph{Sensors}, vol.~23, no.~3, 2023.

\bibitem{poria2017review}
S.~Poria, E.~Cambria, R.~Bajpai, and A.~Hussain, ``A review of affective
  computing: From unimodal analysis to multimodal fusion,'' \emph{Information
  fusion}, vol.~37, pp. 98--125, 2017.

\bibitem{rouast2019deep}
P.~V. Rouast, M.~T. Adam, and R.~Chiong, ``Deep learning for human affect
  recognition: Insights and new developments,'' \emph{IEEE Transactions on
  Affective Computing}, vol.~12, no.~2, pp. 524--543, 2019.

\bibitem{d2015review}
S.~K. D'mello and J.~Kory, ``A review and meta-analysis of multimodal affect
  detection systems,'' \emph{ACM computing surveys}, vol.~47, no.~3, pp. 1--36,
  2015.

\bibitem{schuller2019affective}
B.~Schuller, F.~Weninger, Y.~Zhang, F.~Ringeval, A.~Batliner, S.~Steidl,
  F.~Eyben, E.~Marchi, A.~Vinciarelli, K.~Scherer, M.~Chetouani, and
  M.~Mortillaro, ``Affective and behavioural computing: Lessons learnt from the
  first computational paralinguistics challenge,'' \emph{Computer Speech \&
  Language}, vol.~53, pp. 156--180, 2019.

\bibitem{chakraborty2017}
R.~Chakraborty, M.~Pandharipande, and S.~K. Kopparapu, \emph{Analyzing emotion
  in spontaneous speech}.\hskip 1em plus 0.5em minus 0.4em\relax Springer,
  2017.

\bibitem{deVelasco2022a}
M.~deVelasco, R.~Justo, and M.~I. Torres, ``Automatic identification of
  emotional information in spanish tv debates and human-machine interactions,''
  \emph{Applied Sciences}, vol.~12, no.~4, 2022.

\bibitem{mariooryad2015facial}
S.~Mariooryad and C.~Busso, ``Facial expression recognition in the presence of
  speech using blind lexical compensation,'' \emph{IEEE Transactions on
  Affective Computing}, vol.~7, no.~4, pp. 346--359, 2015.

\bibitem{zeng2007survey}
Z.~Zeng, M.~Pantic, G.~Roisman, and T.~Huang, ``A survey of affect recognition
  methods: Audio, visual, and spontaneous expressions,'' \emph{IEEE
  Transactions on Pattern Analysis and Machine Intelligence}, vol.~1, no.~31,
  pp. 39--58, 2009.

\bibitem{CITA_GO_ON_PETRA}
J.~M. Olaso, M.~García-Sebastián, A.~López~Zorrilla, M.~Tainta,
  M.~Ecay-Torres, M.~Torres, and P.~Martínez-Lage, ``The cita go-on dialogue
  system: Mid-term achievements,'' in \emph{Proc. ACM Int. Conf. on PErvasive
  Technologies Related to Assistive Environments}, 2023.

\bibitem{McTear_VITA2023}
M.~McTear, K.~Jokinen, M.~M. Alam, Q.~Saleem, G.~Napolitano, F.~Szczepaniak,
  M.~Hariz, G.~Chollet, C.~Lohr, J.~Boudy, Z.~Azimi, S.~D. Roelen, and
  R.~Wieching, ``Interaction with a virtual coach for active and healthy
  ageing,'' \emph{Sensors}, vol.~23, no.~5, 2023.

\bibitem{demaeght2022multimodal}
A.~Demaeght, C.~Miclau, J.~Hartmann, J.~Markwardt, and O.~Korn, ``Multimodal
  emotion analysis of robotic assistance in elderly care,'' in
  \emph{Proceedings of the 15th International Conference on PErvasive
  Technologies Related to Assistive Environments}, 2022, pp. 230--236.

\bibitem{magai2006emotion}
C.~Magai, N.~S. Consedine, Y.~S. Krivoshekova, E.~Kudadjie-Gyamfi, and
  R.~McPherson, ``Emotion experience and expression across the adult life span:
  insights from a multimodal assessment study.'' \emph{Psychology and aging},
  vol.~21, no.~2, p. 303, 2006.

\bibitem{Folster2014agevary}
M.~Fölster, U.~Hess, and K.~Werheid, ``Facial age affects emotional expression
  decoding,'' \emph{Frontiers in Psychology}, vol.~5, 2014.

\bibitem{levenson1991emotion}
R.~W. Levenson, L.~L. Carstensen, W.~V. Friesen, and P.~Ekman, ``Emotion,
  physiology, and expression in old age.'' \emph{Psychology and aging}, vol.~6,
  no.~1, p.~28, 1991.

\bibitem{ma2019elderreact}
K.~Ma, X.~Wang, X.~Yang, M.~Zhang, J.~M. Girard, and L.-P. Morency,
  ``Elderreact: a multimodal dataset for recognizing emotional response in
  aging adults,'' in \emph{International Conference on Multimodal Interaction},
  2019, pp. 349--357.

\bibitem{wang2015speech}
K.~Wang, N.~An, B.~N. Li, Y.~Zhang, and L.~Li, ``Speech emotion recognition
  using fourier parameters,'' \emph{IEEE Transactions on affective computing},
  vol.~6, no.~1, pp. 69--75, 2015.

\bibitem{lopes2018facial}
N.~Lopes, A.~Silva, S.~R. Khanal, A.~Reis, J.~Barroso, V.~Filipe, and
  J.~Sampaio, ``Facial emotion recognition in the elderly using a svm
  classifier,'' in \emph{2nd International Conference on Technology and
  Innovation in Sports, Health and Wellbeing}, 2018, pp. 1--5.

\bibitem{schuller2020interspeech}
B.~W. Schuller, A.~Batliner, C.~Bergler, E.-M. Messner, A.~Hamilton,
  S.~Amiriparian, A.~Baird, G.~Rizos, M.~Schmitt, L.~Stappen, H.~Baumeister,
  A.~, Deighton~MacIntyre, and S.~Hantke, ``The interspeech 2020 computational
  paralinguistics challenge: Elderly emotion, breathing \& masks,'' 2020.

\bibitem{kossaifi2019sewa}
J.~Kossaifi, R.~Walecki, Y.~Panagakis, J.~Shen, M.~Schmitt, F.~Ringeval,
  J.~Han, V.~Pandit, A.~Toisoul, B.~Schuller, K.~Star, E.~Hajiyev, and
  M.~Pantic, ``Sewa db: A rich database for audio-visual emotion and sentiment
  research in the wild,'' \emph{IEEE transactions on pattern analysis and
  machine intelligence}, vol.~43, no.~3, pp. 1022--1040, 2019.

\bibitem{Empathic_Midterm}
M.~I. Torres, J.~M. Olaso, C.~Montenegro, R.~Santana, A.~V\'{a}zquez, R.~Justo,
  J.~A. Lozano, S.~Schl\"{o}gl, G.~Chollet, N.~Dugan, M.~Irvine, N.~Glackin,
  C.~Pickard, A.~Esposito, G.~Cordasco, A.~Troncone, D.~Petrovska-Delacretaz,
  A.~Mtibaa, M.~A. Hmani, M.~S. Korsnes, L.~J. Martinussen, S.~Escalera, C.~P.
  Cantari\~{n}o, O.~Deroo, O.~Gordeeva, J.~Tenorio-Laranga, E.~Gonzalez-Fraile,
  B.~Fernandez-Ruanova, and A.~Gonzalez-Pinto, ``The empathic project: Mid-term
  achievements,'' in \emph{Proceedings of the 12th ACM International Conference
  on PErvasive Technologies Related to Assistive Environments}, 2019, p.
  629–638.

\bibitem{Empathic_Demo}
J.~M. Olaso, A.~V\'{a}zquez, L.~Ben~Letaifa, M.~de~Velasco, A.~Mtibaa, M.~A.
  Hmani, D.~Petrovska-Delacr\'{e}taz, G.~Chollet, C.~Montenegro,
  A.~L\'{o}pez-Zorrilla, R.~Justo, R.~Santana, J.~Tenorio-Laranga,
  E.~Gonz\'{a}lez-Fraile, B.~n. Fern\'{a}ndez-Ruanova, G.~Cordasco,
  A.~Esposito, K.~B. Gjellesvik, A.~T. Johansen, M.~S. Kornes, C.~Pickard,
  C.~Glackin, G.~Cahalane, P.~Buch, C.~Palmero, S.~Escalera, O.~Gordeeva,
  O.~Deroo, A.~Fern\'{a}ndez, D.~Kyslitska, J.~A. Lozano, M.~I. Torres, and
  S.~Schl\"{o}gl, ``The empathic virtual coach: A demo,'' in
  \emph{International Conference on Multimodal Interaction}, 2021, p.
  848–851.

\bibitem{ekman1999basic}
P.~Ekman, ``Basic emotions,'' \emph{Handbook of cognition and emotion},
  vol.~98, no. 45-60, p.~16, 1999.

\bibitem{gunes2010automatic}
H.~Gunes and M.~Pantic, ``Automatic, dimensional and continuous emotion
  recognition,'' \emph{International Journal of Synthetic Emotions}, vol.~1,
  no.~1, pp. 68--99, 2010.

\bibitem{calvo2010affect}
R.~A. Calvo and S.~D'Mello, ``Affect detection: An interdisciplinary review of
  models, methods, and their applications,'' \emph{IEEE Transactions on
  affective computing}, vol.~1, no.~1, pp. 18--37, 2010.

\bibitem{schuller2011recognising}
B.~Schuller, A.~Batliner, S.~Steidl, and D.~Seppi, ``Recognising realistic
  emotions and affect in speech: State of the art and lessons learnt from the
  first challenge,'' \emph{Speech communication}, vol.~53, no. 9-10, pp.
  1062--1087, 2011.

\bibitem{russell1980circumplex}
J.~A. Russell, ``A circumplex model of affect.'' \emph{Journal of personality
  and social psychology}, vol.~39, no.~6, p. 1161, 1980.

\bibitem{Valstar:2014}
M.~Valstar, B.~Schuller, K.~Smith, T.~Almaev, F.~Eyben, J.~Krajewski, R.~Cowie,
  and M.~Pantic, ``Avec 2014: 3d dimensional affect and depression recognition
  challenge,'' in \emph{Proceedings of the 4th International Workshop on
  Audio/Visual Emotion Challenge}.\hskip 1em plus 0.5em minus 0.4em\relax ACM,
  2014, pp. 3--10.

\bibitem{huang2019speech}
K.-Y. Huang, C.-H. Wu, Q.-B. Hong, M.-H. Su, and Y.-H. Chen, ``Speech emotion
  recognition using deep neural network considering verbal and nonverbal speech
  sounds,'' in \emph{IEEE International Conference on Acoustics, Speech and
  Signal Processing}, 2019, pp. 5866--5870.

\bibitem{deVelasco_tesis}
M.~deVelasco, ``{Analysis and Automatic Identification of Spontaneous Emotions
  in Speech from Human-Human and Human-Machine Communication},'' Ph.D.
  dissertation, Departamento de Electricidad y Electrónica. Universidad del
  País Vasco UPV/EHU, 2023.

\bibitem{PandaAudioFeatures}
R.~Panda, R.~M. Malheiro, and R.~P. Paiva, ``Audio features for music emotion
  recognition: a survey,'' \emph{IEEE Transactions on Affective Computing}, pp.
  1--1, 2020.

\bibitem{schuller2013interspeech}
B.~Schuller, S.~Steidl, A.~Batliner, A.~Vinciarelli, K.~Scherer, F.~Ringeval,
  M.~Chetouani, F.~Weninger, F.~Eyben, E.~Marchi, M.~Mortillaro, H.~Salamin,
  A.~Polychroniou, F.~Valente, and S.~Kim, ``The interspeech 2013 computational
  paralinguistics challenge: Social signals, conflict, emotion, autism,'' in
  \emph{Proceedings INTERSPEECH 2013, 14th Annual Conference of the
  International Speech Communication Association, Lyon, France}, 2013.

\bibitem{eyben2015}
F.~Eyben, K.~R. Scherer, B.~W. Schuller, J.~Sundberg, E.~Andr{\'e}, C.~Busso,
  L.~Y. Devillers, J.~Epps, P.~Laukka, S.~S. Narayanan, and K.~P. Truong, ``The
  geneva minimalistic acoustic parameter set (gemaps) for voice research and
  affective computing,'' \emph{IEEE transactions on affective computing},
  vol.~7, no.~2, pp. 190--202, 2015.

\bibitem{Baevski2020}
A.~Baevski, Y.~Zhou, A.~Mohamed, and M.~Auli, ``wav2vec 2.0: A framework for
  self-supervised learning of speech representations,'' \emph{Advances in
  neural information processing systems}, vol.~33, pp. 12\,449--12\,460, 2020.

\bibitem{Luna-Jimenez2021}
C.~Luna-Jiménez, R.~Kleinlein, D.~Griol, Z.~Callejas, J.~Montero, and
  F.~Fernández-Martínez, ``A proposal for multimodal emotion recognition
  using aural transformers and action units on ravdess dataset,'' \emph{Applied
  Sciences}, vol.~12, 2022.

\bibitem{Hsu2021}
W.-N. Hsu, B.~Bolte, Y.-H.~H. Tsai, K.~Lakhotia, R.~Salakhutdinov, and
  A.~Mohamed, ``Hubert: Self-supervised speech representation learning by
  masked prediction of hidden units,'' \emph{IEEE/ACM Trans. on Audio, Speech,
  and Language Processing}, vol.~29, pp. 3451--3460, 2021.

\bibitem{wang2021unispeech}
C.~Wang, Y.~Wu, Y.~Qian, K.~Kumatani, S.~Liu, F.~Wei, M.~Zeng, and X.~Huang,
  ``Unispeech: Unified speech representation learning with labeled and
  unlabeled data,'' in \emph{International Conference on Machine
  Learning}.\hskip 1em plus 0.5em minus 0.4em\relax PMLR, 2021, pp.
  10\,937--10\,947.

\bibitem{chen2022wavlm}
S.~Chen, C.~Wang, Z.~Chen, Y.~Wu, S.~Liu, Z.~Chen, J.~Li, N.~Kanda,
  T.~Yoshioka, X.~Xiao, J.~Wu, L.~Zhou, S.~Ren, Y.~Qian, Y.~Qian, J.~Wu,
  M.~Zeng, X.~Yu, and F.~Wei, ``Wavlm: Large-scale self-supervised pre-training
  for full stack speech processing,'' \emph{IEEE Journal of Selected Topics in
  Signal Processing}, vol.~16, no.~6, pp. 1505--1518, 07 2022.

\bibitem{singh2022}
Y.~B. Singh and S.~Goel, ``A systematic literature review of speech emotion
  recognition approaches,'' \emph{Neurocomputing}, vol. 492, pp. 245--263,
  2022.

\bibitem{deLope2023review}
J.~{de Lope} and M.~Graña, ``An ongoing review of speech emotion
  recognition,'' \emph{Neurocomputing}, vol. 528, pp. 1--11, 2023.

\bibitem{Wang2020}
J.~Wang, M.~Xue, R.~Culhane, E.~Diao, J.~Ding, and V.~Tarokh, ``Speech emotion
  recognition with dual-sequence lstm architecture,'' in \emph{IEEE
  International Conference on Acoustics, Speech and Signal Processing}, 2020,
  pp. 6474--6478.

\bibitem{Morais2022}
E.~Morais, R.~Hoory, W.~Zhu, I.~Gat, M.~Damasceno, and H.~Aronowitz, ``Speech
  emotion recognition using self-supervised features,'' in \emph{IEEE
  International Conference on Acoustics, Speech and Signal Processing}, 2022,
  pp. 6922--6926.

\bibitem{wagner2023dawn}
J.~Wagner, A.~Triantafyllopoulos, H.~Wierstorf, M.~Schmitt, F.~Burkhardt,
  F.~Eyben, and B.~W. Schuller, ``Dawn of the transformer era in speech emotion
  recognition: closing the valence gap,'' \emph{IEEE Transactions on Pattern
  Analysis and Machine Intelligence}, 2023.

\bibitem{atmaja2022survey}
B.~T. Atmaja, A.~Sasou, and M.~Akagi, ``Survey on bimodal speech emotion
  recognition from acoustic and linguistic information fusion,'' \emph{Speech
  Communication}, vol. 140, pp. 11--28, 2022.

\bibitem{Li_Deng2018}
S.~Li and W.~Deng, ``Deep facial expression recognition: A survey,'' \emph{IEEE
  transactions on affective computing}, vol.~13, no.~3, pp. 1195--1215, 2020.

\bibitem{Mohammadi2014}
M.~Mohammadi, E.~Fatemizadeh, and M.~Mahoor, ``{PCA}-based dictionary building
  for accurate facial expression recognition via sparse representation,''
  \emph{Journal of Visual Communication and Image Representation}, vol.~25,
  no.~5, pp. 1082--1092, Jul. 2014.

\bibitem{ChengjunLiu2002}
C.~Liu and H.~Wechsler, ``Gabor feature based classification using the enhanced
  fisher linear discriminant model for face recognition,'' \emph{{IEEE}
  Transactions on Image Processing}, vol.~11, no.~4, pp. 467--476, 2002.

\bibitem{Shan2009}
C.~Shan, S.~Gong, and P.~W. McOwan, ``Facial expression recognition based on
  local binary patterns: A comprehensive study,'' \emph{Image and Vision
  Computing}, vol.~27, no.~6, pp. 803--816, May 2009.

\bibitem{Mavadati2013}
S.~M. Mavadati, M.~H. Mahoor, K.~Bartlett, P.~Trinh, and J.~F. Cohn, ``{DISFA}:
  A spontaneous facial action intensity database,'' \emph{{IEEE} Transactions
  on Affective Computing}, vol.~4, no.~2, pp. 151--160, 2013.

\bibitem{corneanu2016survey}
C.~A. Corneanu, M.~O. Sim{\'o}n, J.~F. Cohn, and S.~E. Guerrero, ``Survey on
  rgb, 3d, thermal, and multimodal approaches for facial expression
  recognition: History, trends, and affect-related applications,'' \emph{IEEE
  transactions on pattern analysis and machine intelligence}, vol.~38, no.~8,
  pp. 1548--1568, 2016.

\bibitem{ko2018brief}
B.~C. Ko, ``A brief review of facial emotion recognition based on visual
  information,'' \emph{sensors}, vol.~18, no.~2, p. 401, 2018.

\bibitem{mollahosseini2016going}
A.~Mollahosseini, D.~Chan, and M.~H. Mahoor, ``Going deeper in facial
  expression recognition using deep neural networks,'' in \emph{IEEE Winter
  conference on applications of computer vision}, 2016, pp. 1--10.

\bibitem{graham2012neurocognitive}
R.~Graham and K.~S. LaBar, ``Neurocognitive mechanisms of gaze-expression
  interactions in face processing and social attention,''
  \emph{Neuropsychologia}, vol.~50, no.~5, pp. 553--566, 2012.

\bibitem{lim2020emotion}
J.~Z. Lim, J.~Mountstephens, and J.~Teo, ``Emotion recognition using
  eye-tracking: taxonomy, review and current challenges,'' \emph{Sensors},
  vol.~20, no.~8, p. 2384, 2020.

\bibitem{o2018affective}
J.~O'Dwyer, N.~Murray, and R.~Flynn, ``Affective computing using speech and eye
  gaze: a review and bimodal system proposal for continuous affect
  prediction,'' \emph{arXiv preprint arXiv:1805.06652}, 2018.

\bibitem{zhang2019evaluation}
X.~Zhang, Y.~Sugano, and A.~Bulling, ``Evaluation of appearance-based methods
  and implications for gaze-based applications,'' in \emph{Proc. CHI conference
  on human factors in computing systems}, 2019, pp. 1--13.

\bibitem{ghosh2021automatic}
S.~Ghosh, A.~Dhall, M.~Hayat, J.~Knibbe, and Q.~Ji, ``Automatic gaze analysis:
  A survey of deep learning based approaches,'' \emph{arXiv preprint
  arXiv:2108.05479}, 2021.

\bibitem{alghowinem2016multimodal}
S.~Alghowinem, R.~Goecke, M.~Wagner, J.~Epps, M.~Hyett, G.~Parker, and
  M.~Breakspear, ``Multimodal depression detection: fusion analysis of
  paralinguistic, head pose and eye gaze behaviors,'' \emph{IEEE Transactions
  on Affective Computing}, vol.~9, no.~4, pp. 478--490, 2016.

\bibitem{abdou2022gaze}
A.~Abdou, E.~Sood, P.~M{\"u}ller, and A.~Bulling, ``Gaze-enhanced crossmodal
  embeddings for emotion recognition,'' \emph{Proc. ACM on Human-Computer
  Interaction}, vol.~6, no. ETRA, pp. 1--18, 2022.

\bibitem{cortacero2019rt}
K.~Cortacero, T.~Fischer, and Y.~Demiris, ``Rt-bene: A dataset and baselines
  for real-time blink estimation in natural environments,'' in \emph{Proc.
  International Conference on Computer Vision Workshops}, 2019.

\bibitem{guitton1987gaze}
D.~Guitton and M.~Volle, ``Gaze control in humans: eye-head coordination during
  orienting movements to targets within and beyond the oculomotor range,''
  \emph{Journal of neurophysiology}, vol.~58, no.~3, pp. 427--459, 1987.

\bibitem{el2005real}
R.~El~Kaliouby and P.~Robinson, ``Real-time inference of complex mental states
  from facial expressions and head gestures,'' \emph{Real-time vision for
  human-computer interaction}, pp. 181--200, 2005.

\bibitem{hess2007looking}
U.~Hess, R.~B. Adams, and R.~E. Kleck, ``Looking at you or looking elsewhere:
  The influence of head orientation on the signal value of emotional facial
  expressions,'' \emph{Motivation and Emotion}, vol.~31, pp. 137--144, 2007.

\bibitem{busso2007rigid}
C.~Busso, Z.~Deng, M.~Grimm, U.~Neumann, and S.~Narayanan, ``Rigid head motion
  in expressive speech animation: Analysis and synthesis,'' \emph{IEEE
  transactions on audio, speech, and language processing}, vol.~15, no.~3, pp.
  1075--1086, 2007.

\bibitem{karg2013body}
M.~Karg, A.-A. Samadani, R.~Gorbet, K.~K{\"u}hnlenz, J.~Hoey, and D.~Kuli{\'c},
  ``Body movements for affective expression: A survey of automatic recognition
  and generation,'' \emph{IEEE Transactions on Affective Computing}, vol.~4,
  no.~4, pp. 341--359, 2013.

\bibitem{gunes2010dimensional}
H.~Gunes and M.~Pantic, ``Dimensional emotion prediction from spontaneous head
  gestures for interaction with sensitive artificial listeners,'' in
  \emph{Intelligent Virtual Agents: 10th International Conference, IVA 2010.
  Proceedings 10}.\hskip 1em plus 0.5em minus 0.4em\relax Springer, 2010, pp.
  371--377.

\bibitem{adams2015decoupling}
A.~Adams, M.~Mahmoud, T.~Baltru{\v{s}}aitis, and P.~Robinson, ``Decoupling
  facial expressions and head motions in complex emotions,'' in \emph{2015
  International conference on affective computing and intelligent
  interaction}.\hskip 1em plus 0.5em minus 0.4em\relax IEEE, 2015, pp.
  274--280.

\bibitem{samanta2017role}
A.~Samanta and T.~Guha, ``On the role of head motion in affective expression,''
  in \emph{IEEE International Conference on Acoustics, Speech and Signal
  Processing}.\hskip 1em plus 0.5em minus 0.4em\relax IEEE, 2017, pp.
  2886--2890.

\bibitem{ding2018low}
Y.~Ding, L.~Shi, and Z.~Deng, ``Low-level characterization of expressive head
  motion through frequency domain analysis,'' \emph{IEEE Transactions on
  Affective Computing}, vol.~11, no.~3, pp. 405--418, 2018.

\bibitem{samanta2020emotion}
A.~Samanta and T.~Guha, ``Emotion sensing from head motion capture,''
  \emph{IEEE Sensors Journal}, vol.~21, no.~4, pp. 5035--5043, 2020.

\bibitem{khan2021head}
K.~Khan, R.~U. Khan, R.~Leonardi, P.~Migliorati, and S.~Benini, ``Head pose
  estimation: A survey of the last ten years,'' \emph{Signal Processing: Image
  Communication}, vol.~99, p. 116479, 2021.

\bibitem{xue2021investigating}
T.~Xue, A.~E. Ali, G.~Ding, and P.~Cesar, ``Investigating the relationship
  between momentary emotion self-reports and head and eye movements in
  hmd-based 360 vr video watching,'' in \emph{Extended abstracts of the 2021
  conference on human factors in computing systems}, 2021, pp. 1--8.

\bibitem{o2019eye}
J.~O'Dwyer, N.~Murray, and R.~Flynn, ``Eye-based continuous affect
  prediction,'' in \emph{2019 8th International Conference on Affective
  Computing and Intelligent Interaction}.\hskip 1em plus 0.5em minus
  0.4em\relax IEEE, 2019, pp. 137--143.

\bibitem{ramachandram2017deep}
D.~Ramachandram and G.~W. Taylor, ``Deep multimodal learning: A survey on
  recent advances and trends,'' \emph{IEEE signal processing magazine},
  vol.~34, no.~6, pp. 96--108, 2017.

\bibitem{baltruvsaitis2018multimodal}
T.~Baltru{\v{s}}aitis, C.~Ahuja, and L.-P. Morency, ``Multimodal machine
  learning: A survey and taxonomy,'' \emph{IEEE transactions on pattern
  analysis and machine intelligence}, vol.~41, no.~2, pp. 423--443, 2018.

\bibitem{guo2019deep}
W.~Guo, J.~Wang, and S.~Wang, ``Deep multimodal representation learning: A
  survey,'' \emph{IEEE Access}, vol.~7, pp. 63\,373--63\,394, 2019.

\bibitem{wu2014survey}
C.-H. Wu, J.-C. Lin, and W.-L. Wei, ``Survey on audiovisual emotion
  recognition: databases, features, and data fusion strategies,'' \emph{APSIPA
  transactions on signal and information processing}, vol.~3, p. e12, 2014.

\bibitem{cohn2004multimodal}
J.~F. Cohn, L.~I. Reed, T.~Moriyama, J.~Xiao, K.~Schmidt, and Z.~Ambadar,
  ``Multimodal coordination of facial action, head rotation, and eye motion
  during spontaneous smiles,'' in \emph{IEEE Int. Conf. on Automatic Face and
  Gesture Recognition}, 2004, pp. 129--135.

\bibitem{wu2019continuous}
S.~Wu, Z.~Du, W.~Li, D.~Huang, and Y.~Wang, ``Continuous emotion recognition in
  videos by fusing facial expression, head pose and eye gaze,'' in \emph{Int.
  Conf. on Multimodal Interaction}, 2019, pp. 40--48.

\bibitem{alhargan2017multimodal}
A.~Alhargan, N.~Cooke, and T.~Binjammaz, ``Multimodal affect recognition in an
  interactive gaming environment using eye tracking and speech signals,'' in
  \emph{Proceedings of the 19th ACM international conference on multimodal
  interaction}, 2017, pp. 479--486.

\bibitem{o2019speech}
J.~O'Dwyer, ``Speech, head, and eye-based cues for continuous affect
  prediction,'' in \emph{8th International Conference on Affective Computing
  and Intelligent Interaction Workshops and Demos}, 2019, pp. 16--20.

\bibitem{Schogl2015}
S.~Schlögl, G.~Doherty, and S.~Luz, ``Wizard of oz experimentation for
  language technology applications: challenges and tools.'' \emph{Interacting
  with Computers}, vol.~27, no.~6, pp. 592--615, Nov 2015.

\bibitem{cowen2017self}
A.~S. Cowen and D.~Keltner, ``Self-report captures 27 distinct categories of
  emotion bridged by continuous gradients,'' \emph{Proc. of the national
  academy of sciences}, vol. 114, no.~38, pp. E7900--E7909, 2017.

\bibitem{esposito2009}
A.~Esposito, ``The perceptual and cognitive role of visual and auditory
  channels in conveying emotional information,'' \emph{Cognitive Computation},
  vol.~1, pp. 268--278, 2009.

\bibitem{russell2003facial}
J.~A. Russell, J.-A. Bachorowski, and J.-M. Fern{\'a}ndez-Dols, ``Facial and
  vocal expressions of emotion,'' \emph{Annual review of psychology}, vol.~54,
  no.~1, pp. 329--349, 2003.

\bibitem{riviello2012cross}
M.~T. Riviello and A.~Esposito, ``A cross-cultural study on the effectiveness
  of visual and vocal channels in transmitting dynamic emotional information,''
  \emph{Acta Polytechnica Hungarica}, vol.~9, no.~1, pp. 157--170, 2012.

\bibitem{mchugh2012interrater}
M.~L. McHugh, ``Interrater reliability: the kappa statistic,'' \emph{Biochemia
  medica}, vol.~22, no.~3, pp. 276--282, 2012.

\bibitem{steininger2002development}
S.~Steininger, F.~Schiel, O.~Dioubina, and S.~Raubold, ``Development of
  user-state conventions for the multimodal corpus in smartkom,'' in
  \emph{Proc. Workshop on Multimodal Resources and Multimodal Systems
  Evaluation}.\hskip 1em plus 0.5em minus 0.4em\relax Citeseer, 2002, pp.
  33--37.

\bibitem{greco2021emotional}
C.~Greco, C.~Buono, P.~Buch-Cardona, G.~Cordasco, S.~Escalera, A.~Esposito,
  A.~Fernandez, D.~Kyslitska, M.~S. Kornes, C.~Palmero, J.~Tenorio-Laranga,
  A.~Torp~Johansen, and M.~I. Torres, ``Emotional features of interactions with
  empathic agents,'' in \emph{Proc. IEEE/CVF International Conference on
  Computer Vision}, 2021, pp. 2168--2176.

\bibitem{zhang2017faceboxes}
S.~Zhang, X.~Zhu, Z.~Lei, H.~Shi, X.~Wang, and S.~Z. Li, ``Faceboxes: A cpu
  real-time face detector with high accuracy,'' in \emph{IJCB}, 2017.

\bibitem{guo2020towards}
J.~Guo, X.~Zhu, Y.~Yang, F.~Yang, Z.~Lei, and S.~Z. Li, ``Towards fast,
  accurate and stable 3d dense face alignment,'' in \emph{European Conference
  on Computer Vision}.\hskip 1em plus 0.5em minus 0.4em\relax Springer, 2020,
  pp. 152--168.

\bibitem{Chollet2016}
F.~Chollet, ``Xception: Deep learning with depthwise separable convolutions,''
  in \emph{Proc. IEEE conference on computer vision and pattern recognition},
  2017, pp. 1251--1258.

\bibitem{Deng2009}
J.~Deng, W.~Dong, R.~Socher, L.-J. Li, K.~Li, and L.~Fei-Fei, ``{ImageNet}: A
  large-scale hierarchical image database,'' in \emph{{IEEE} Conference on
  Computer Vision and Pattern Recognition}.\hskip 1em plus 0.5em minus
  0.4em\relax {IEEE}, Jun. 2009.

\bibitem{huber2016multiresolution}
P.~Huber, G.~Hu, R.~Tena, P.~Mortazavian, P.~Koppen, W.~J. Christmas,
  M.~Ratsch, and J.~Kittler, ``A multiresolution 3d morphable face model and
  fitting framework,'' in \emph{Proceedings of the 11th international joint
  conference on computer vision, imaging and computer graphics theory and
  applications}, 2016.

\bibitem{Lepetit:160138}
V.~Lepetit, F.~Moreno-Noguer, and P.~Fua, ``Epnp: An accurate o(n) solution to
  the pnp problem,'' \emph{International Journal Of Computer Vision}, vol.~81,
  pp. 155--166, 2009.

\bibitem{zhang2020eth}
X.~Zhang, S.~Park, T.~Beeler, D.~Bradley, S.~Tang, and O.~Hilliges,
  ``Eth-xgaze: A large scale dataset for gaze estimation under extreme head
  pose and gaze variation,'' in \emph{European Conference on Computer
  Vision}.\hskip 1em plus 0.5em minus 0.4em\relax Springer, 2020, pp. 365--381.

\bibitem{bulling2010eye}
A.~Bulling, J.~A. Ward, H.~Gellersen, and G.~Tr{\"o}ster, ``Eye movement
  analysis for activity recognition using electrooculography,'' \emph{IEEE
  transactions on pattern analysis and machine intelligence}, vol.~33, no.~4,
  pp. 741--753, 2010.

\bibitem{amorese2022using}
T.~Amorese, C.~Greco, M.~Cuciniello, C.~Buono, C.~Palmero, P.~Buch-Cardona,
  S.~Escalera, M.~I. Torres, G.~Cordasco, and A.~Esposito, ``Using eye tracking
  to investigate interaction between humans and virtual agents,'' in \emph{IEEE
  Conference on Cognitive and Computational Aspects of Situation Management
  (CogSIMA)}, 2022, pp. 125--132.

\bibitem{rajan2022cross}
V.~Rajan, A.~Brutti, and A.~Cavallaro, ``Is cross-attention preferable to
  self-attention for multi-modal emotion recognition?'' in \emph{IEEE
  International Conference on Acoustics, Speech and Signal Processing}, 2022,
  pp. 4693--4697.

\bibitem{bouckaert2004evaluating}
R.~R. Bouckaert and E.~Frank, ``Evaluating the replicability of significance
  tests for comparing learning algorithms,'' in \emph{Pacific-Asia Conference
  on Knowledge Discovery and Data Mining}, vol. 3056.\hskip 1em plus 0.5em
  minus 0.4em\relax Springer, 2004, pp. 3--12.

\bibitem{benjamini2006adaptive}
Y.~Benjamini, A.~M. Krieger, and D.~Yekutieli, ``Adaptive linear step-up
  procedures that control the false discovery rate,'' \emph{Biometrika},
  vol.~93, no.~3, pp. 491--507, 2006.

\bibitem{tzirakis2017end}
P.~Tzirakis, G.~Trigeorgis, M.~A. Nicolaou, B.~W. Schuller, and S.~Zafeiriou,
  ``End-to-end multimodal emotion recognition using deep neural networks,''
  \emph{IEEE Journal of selected topics in signal processing}, vol.~11, no.~8,
  pp. 1301--1309, 2017.

\bibitem{Filali2023}
A.~Filali~Razzouki, L.~Jeancolas, G.~Mangone, S.~Sambin, A.~Chalançon,
  M.~Gomes, S.~Lehéricy, J.-C. Corvol, M.~Vidailhet, I.~Arnulf, M.~A.
  El-Yacoubi, and D.~Petrovska-Delacrétaz, ``Early-stage parkinson's disease
  detection based on action unit derivatives,'' in \emph{"Dispositifs
  biomédicaux et technologies numériques en santé ; des besoins aux usages"
  (Colloque JETSAN )}, 2023, pp. 1--8.

\bibitem{tsai2019multimodal}
Y.-H.~H. Tsai, S.~Bai, P.~P. Liang, J.~Z. Kolter, L.-P. Morency, and
  R.~Salakhutdinov, ``Multimodal transformer for unaligned multimodal language
  sequences,'' in \emph{Proceedings of the conference. Association for
  Computational Linguistics. Meeting}, vol. 2019, 2019, p. 6558.

\bibitem{shin2016normal}
Y.~Shin, H.~Lim, M.~Kang, M.~Seong, H.~Cho, and J.~Kim, ``Normal range of eye
  movement and its relationship to age,'' \emph{Acta Ophthalmologica}, vol.~94,
  2016.

\bibitem{bahill1975main}
A.~T. Bahill, M.~R. Clark, and L.~Stark, ``The main sequence, a tool for
  studying human eye movements,'' \emph{Mathematical biosciences}, vol.~24, no.
  3-4, pp. 191--204, 1975.

\bibitem{grossman1988frequency}
G.~E. Grossman, R.~J. Leigh, L.~A. Abel, D.~J. Lanska, and S.~Thurston,
  ``Frequency and velocity of rotational head perturbations during
  locomotion,'' \emph{Experimental brain research}, vol.~70, no.~3, pp.
  470--476, 1988.

\bibitem{comaniciu2002mean}
D.~Comaniciu and P.~Meer, ``Mean shift: A robust approach toward feature space
  analysis,'' \emph{IEEE Transactions on pattern analysis and machine
  intelligence}, vol.~24, no.~5, pp. 603--619, 2002.

\end{thebibliography}
